\documentclass[11pt]{article}
\usepackage[T1]{fontenc}
\usepackage{lmodern}
\ifdefined\XeTeXversion
  \usepackage{fontspec}
\fi
\usepackage[letterpaper,margin=1in,headheight=14pt]{geometry}
\usepackage{microtype}
\usepackage[numbers,sort&compress]{natbib}
\setcitestyle{numbers,square,citesep={,}}
\usepackage{fancyhdr}
\usepackage[font=small,labelfont=bf]{caption}
\ifdefined\pdfsuppressptexinfo\pdfsuppressptexinfo=-1\fi

\usepackage{xcolor}
\usepackage{hyperref}
\usepackage{xurl}      
\usepackage{url}

\usepackage[utf8]{inputenc} 
\ifdefined\DeclareUnicodeCharacter
  \DeclareUnicodeCharacter{22C5}{\ensuremath{\cdot}}
\fi
\usepackage{longtable}
\usepackage{array}
\usepackage{booktabs}       
\usepackage{amsfonts}       
\usepackage{nicefrac}       

\usepackage{bm}
\usepackage{amsmath}
\usepackage{float}
\usepackage{wasysym}
\usepackage{multirow}
\usepackage{graphicx}
\graphicspath{{figures/}{arXiv/figures/}}
\usepackage{adjustbox}

\usepackage[normalem]{ulem}

\title{NEMORA: Neural Equivariant Multipole Operators for Long-Range Atomistic Learning}

\hypersetup{colorlinks=true,linkcolor=blue!55!black,citecolor=blue!55!black,urlcolor=blue!55!black}
\renewcommand{\lhead}[1]{\fancyhead[L]{\small\itshape #1}}

\makeatletter
\renewcommand{\@maketitle}{%
  \newpage
  \null
  \vskip 1em%
  \begin{center}%
    {\LARGE\bfseries \@title \par}%
    \vskip 1.5em%
    {\small\lineskip .5em%
      \ignorespaces\@author\par}%
  \end{center}%
  \par
  \vskip 1.5em}
\makeatother

\author{\textbf{Jay L. Kaplan\textsuperscript{1}, Samuel Varner\textsuperscript{2}, Rebecca Willett\textsuperscript{3,4,5},} \textbf{Juan J. de Pablo\textsuperscript{1,2,6,*}}\\[0.5em]
\textsuperscript{\textbf{1}}Courant Institute School of Mathematics, Computing, and Data Science,\\
New York University, New York, NY 10012, USA\\[0.3em]
\textsuperscript{\textbf{2}}Department of Chemical and Biomolecular Engineering, Tandon School of Engineering, \\New York University, Brooklyn, NY 11201, USA\\[0.3em]
\textsuperscript{\textbf{3}}Department of Statistics, The University of Chicago, Chicago, IL 60637, USA\\[0.3em]
\textsuperscript{\textbf{4}}Department of Computer Science, The University of Chicago, Chicago, IL 60637, USA\\[0.3em]
\textsuperscript{\textbf{5}}NSF-Simons National Institute for Theory and Mathematics in Biology, Chicago, IL 60611, USA \\[0.3em]
\textsuperscript{\textbf{6}}Department of Physics, New York University, New York, NY 10012, USA\\[0.5em]
\textsuperscript{*}Corresponding author: Juan J. de Pablo, \protect\href{mailto:jjd8110@nyu.edu}{jjd8110@nyu.edu}
}

\begin{document}

\maketitle
\lhead{Preprint}

\begin{abstract}
Equivariant graph neural networks have emerged as foundational architectures for machine-learned interatomic potentials, approaching quantum-chemical accuracy at a fraction of the computational cost. These models describe local atomic environments accurately, but finite spatial cutoffs truncate long-range information flow, and stacking message-passing layers can lead to over-smoothing and over-squashing. Existing long-range extensions either prescribe a fixed analytical propagation kernel, restrict long-range communication to scalars or degree-preserving channels, are only approximately equivariant, or incur super-linear computational cost. Combining learnable long-range equivariant transport with multiscale many-body expressivity and efficient scaling for larger systems remains a central challenge. We introduce Neural Equivariant Multipole Operators (NEMORA), a neural equivariant extension of the Fast Multipole Method (FMM) for learning long-range tensorial representations. NEMORA generalizes the FMM's analytical multipole expansion and translation operators to learned equivariant counterparts on an adaptive spatial hierarchy. Its operators couple angular degrees and form many-body interactions across length scales, retaining the FMM's hierarchical organization and analytical radial factors as physical inductive biases while learning data-dependent long-range couplings. NEMORA evaluates in linear time and memory complexity, allowing it to treat larger systems than other long-range methods reaching hundreds of thousands of atoms, and it augments both symmetry-constrained and unconstrained short-range backbones. On non-local benchmarks, it reduces force and energy errors relative to the short-range backbones by over an order of magnitude and up to three orders of magnitude, respectively, which is better than or competitive with existing long-range extensions in accuracy.
\end{abstract}

\section{Introduction}

Modeling atomic interactions underpins central tasks in computational science, from finding low-energy conformations to computing thermodynamic and kinetic properties by molecular simulation. Classical force fields enable large simulations at modest cost, but their prescribed functional forms limit the chemistry they can represent. Machine-learned interatomic potentials (MLIPs) learn potential energy surfaces from quantum-chemical data, approaching this same accuracy at a fraction of the cost \citep{behlerGeneralizedNeuralNetworkRepresentation2007,bartokGaussianApproximationPotentials2010,smithANI1ExtensibleNeural2017,unkeMachineLearningForce2021}. Equivariant architectures are especially accurate and data-efficient because they build the symmetries of atomic environments into their representations \citep{Schutt2021EquivariantSpectra,batznerE3equivariantGraphNeural2022,batatiaMACEHigherOrder2023,Musaelian2023Allegro,liaoEquiformerEquivariantGraph2023,Frank2024Euclidean,Wang2024VisNet}.

However, these architectures rely on finite spatial cutoffs for efficiency, which limits their treatment of long-range interactions such as electrostatics, charge transfer, dispersion, polarization and other non-local electronic effects. These interactions govern interfaces, charged systems, polar condensed phases and biomolecular assemblies, among others \citep{jacksonRecentAdvancesMachine2019,zhouElectrostaticInteractionsProtein2018,yueWhenShortrangeAtomistic2021,niblettLearningIntermolecularForces2021,parsaeifardFingerprintBasedDetectionNonLocal2021,Wang2024AbInitio,parkerFalseMetallizationShortRanged2026}. Stacking message-passing layers extends the receptive field on connected graphs, but increases cost, causes over-smoothing and over-squashing \citep{oonoGraphNeuralNetworks2020,alonBottleneckGraphNeural2021,giovanni_over-squashing_2023}, and cannot connect fragments that are disconnected in the cutoff graph. These limitations motivate dedicated long-range channels with tractable scaling.

Long-range extensions include latent scalar methods \citep{kosmalaEwaldbasedLongRangeMessage2023,chengLatentEwaldSummation2025}, charge equilibration \citep{koFourthgenerationHighdimensionalNeural2021,fuchsLearningNonlocalMolecular2025,gublerAcceleratingFourthGenerationMachine2024b}, latent tensor fields \citep{rumiantsevLearningLongRangeRepresentations2026,zhangSO3equivariantReciprocalspaceNeural2026}, polarizable multipoles \citep{batatia_mace-polar-1_2026,parkPolarizableAtomicMultipoles2026}, global attention and aggregation \citep{quRecipeScalableAttentionbased2026,frankMachineLearningGlobal2026,carusoExtendingRangeGraph2026}, matrix-function representations \citep{Batatia2024EquivariantNetworks}, and multiscale representations \citep{hanEquivariantGraphHierarchyBased2022,zhdanovErwinTreebasedHierarchical2025,soleMachineLearningMultiscale2026} (Appendix Table~\ref{tab:methods}). These advances extend non-local modeling, but do not jointly learn the long-range propagation kernel with a physics-inductive bias, cross-degree angular coupling, or explicit higher-body-order interactions among distant sources on an adaptive hierarchy with $\mathcal{O}(N)$ time and memory. The Fast Multipole Method (FMM) is a natural foundation for these requirements \citep{greengard_fast_1987,cheng_fast_1999}. It represents well-separated source clusters by compact multipole expansions across a spatial hierarchy with linear complexity. For the Coulomb kernel, spherical harmonics and analytical radial factors give a physically grounded basis for equivariant transport, built from the same ingredients as equivariant message passing \citep{cheng_fast_1999, batatia_design_2025}.

\begin{figure}[!t]
\centering
\includegraphics[width=\linewidth]{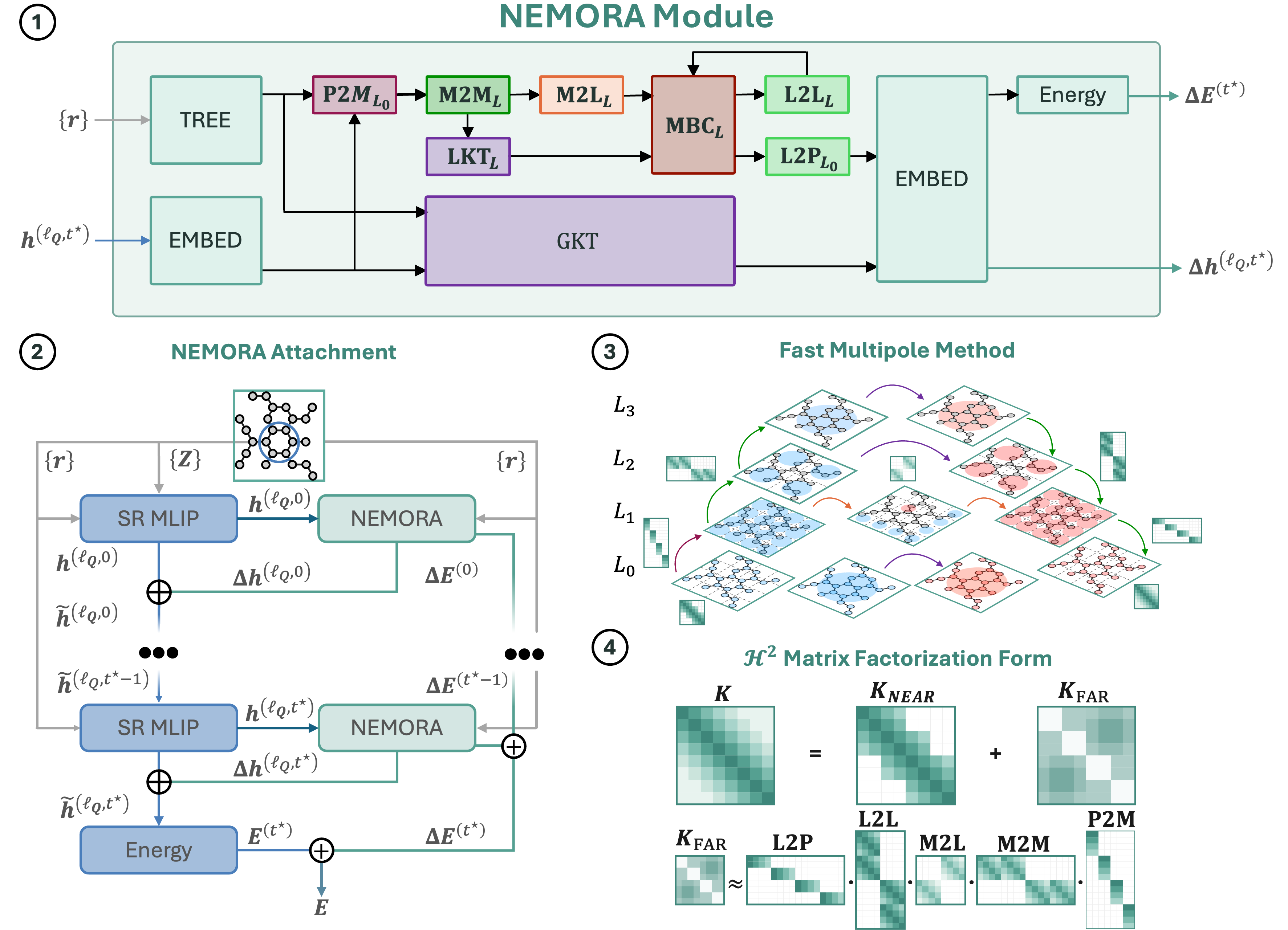}
\caption{\textbf{NEMORA: a learnable Fast Multipole Method for atomistic ML.}
\textbf{(1) NEMORA module (top).} Given backbone features $\bm{h}_i^{(\ell_Q, t^\star)}$ and positions $\{\bm{r}_i\}$, NEMORA returns equivariant corrections $\Delta \bm{h}_i^{(\ell_Q, t^\star)}$ and a scalar contribution $\Delta E^{(t^\star)}$. An upward sweep (P2M, M2M) builds cluster moments; at each downward level, the M2L far field is fused with a local Kelvin self-correction (LKT) and a many-body contraction (MBC) before L2L. Leaf L2P outputs are summed with a global Kelvin reaction-field branch (GKT).
\textbf{(2) NEMORA--MLIP integration (bottom left).} NEMORA updates the backbone equivariants at iteration $t^\star$, and per-iteration readouts add $\Delta E^{(t^\star)}$ to the atomic energy.
\textbf{(3, 4) Classical FMM (right).} Translation of multipole (blue) and local (red) expansions, and the corresponding $\mathcal{H}^2$ factorization~\citep{hackbuschSparseMatrixArithmetic1999} $K = K_{\text{NEAR}} + K_{\text{FAR}}$ with $K_{\text{FAR}} \approx \text{L2P}\cdot\text{L2L}\cdot\text{M2L}\cdot\text{M2M}\cdot\text{P2M}$.}
\label{fig:1overview}
\end{figure}

We introduce \textit{Neural Equivariant Multipole Operators (NEMORA)}, a neural equivariant extension of the FMM for learning long-range many-body interactions (Figure~\ref{fig:1overview}). NEMORA replaces scalar source charges with learned equivariant tensor charges, and the analytical expansion and translation operators with learnable $E(3)$-equivariant counterparts on an adaptive spatial hierarchy. Its operators learn weights over Clebsch--Gordan paths, coupling angular degrees in the long-range channel. Learnable non-linear radial modulations and invariant gates adapt the analytical Green's-function kernels to the data. Symmetric tensor contractions \citep{drautz_atomic_2019,batatiaMACEHigherOrder2023} on the aggregated fields form explicit higher-body-order interactions among distant collections of particles. The long-range messages then update the short-range backbone features and are passed through a readout layer as a residual energy term, with forces obtained via automatic differentiation of the energy (Section~\ref{sec:hierarchy}). Evaluation is $\mathcal{O}(N)$ in time and memory for both periodic and non-periodic systems. Our contributions are:

\begin{itemize}\setlength{\itemsep}{0pt}
\item \textbf{Learnable equivariant FMM.} Learned angular (degree-coupling) transport and symmetric tensor contractions build multiscale long-range many-body representations with $\mathcal{O}(N)$ time and memory cost over an adaptive spatial hierarchy that preserves equivariance up to eigenframe degeneracy.

\item \textbf{Global reaction-field channel.} Inspired by continuum electrostatics \citep{bornVolumenUndHydratationswaerme1920,onsagerElectricMomentsMolecules1936,kirkwoodDielectricPolarizationPolar1939}, a learnable Kelvin transform \citep{amaral_kelvin_2017} inverts globally pooled multipoles into a same-center local expansion, an equivariant continuum boundary response broadcast to atoms that complements hierarchical transport (Appendix~\ref{si:kelvin}).

\item \textbf{Accuracy, portability, and scaling.} NEMORA reduces force and energy errors by over an order of magnitude and up to three orders of magnitude, respectively, relative to short-range models on non-local benchmarks, which is better than or competitive with other long-range methods. Additionally, improvements on four short-range backbones demonstrate NEMORA's portability to both symmetry-constrained and -unconstrained architectures, where the lift for scalar-only hosts recovers part of the native-equivariant gain. We show that NEMORA is linear in evaluation runtime and memory up to approximately 340,000 atoms on a single GPU.
\end{itemize}

\section{Background}
\subsection{Equivariant Machine-Learned Interatomic Potentials}
\label{sec:mlips}
\paragraph{Machine-Learned Interatomic Potentials} An MLIP maps atomic positions $\{\bm r_i\}_{i=1}^N\subset\mathbb R^3$ and species $\{Z_i\}_{i=1}^N$ to an invariant potential energy $E=\sum_i E_i$, with forces $\bm F_i=-\nabla_{\bm r_i}E$ \citep{behlerGeneralizedNeuralNetworkRepresentation2007,batznerE3equivariantGraphNeural2022}. Species embeddings initialize scalar node features $\bm h_i^{(0)}$, and a cutoff graph defines $\mathcal N(i)=\{j\ne i:\|\bm r_j-\bm r_i\|<r_{\mathrm{cut}}\}$. With $\bm r_{ij}=\bm r_j-\bm r_i$, message-passing layer $t\in\{1,\ldots,T\}$ updates these features as \citep{gilmer_neural_2017}
\begin{equation}
\label{eq:mpnn}
\bm h_i^{(t)}=U_{t}\!\Bigl(\bm h_i^{(t-1)},\;\bigoplus_{j\in\mathcal N(i)}M_{t}\!\bigl(\bm h_i^{(t-1)},\bm h_j^{(t-1)},\bm r_{ij}\bigr)\Bigr),
\end{equation}
where $M_{t}$ and $U_{t}$ are learned message and update functions, and $\bigoplus$ is permutation-invariant aggregation. Invariant readouts give $E_i=\sum_{t=1}^{T}\mathcal R_{t}(\bm h_i^{(t)})$.

\paragraph{Equivariant Graph Neural Networks} Equivariant features are organized into blocks $h_{i,km}^{(\ell)}$ with channel $k$, angular degree $\ell$, and magnetic component $m\in\{-\ell,\ldots,\ell\}$. Under a rotation $R\in\mathrm{SO}(3)$, $h_{i,km}^{(\ell)}\mapsto\sum_{m'}D_{mm'}^{(\ell)}(R)h_{i,km'}^{(\ell)}$, where $D^{(\ell)}$ is a Wigner D-matrix. The Clebsch--Gordan (CG) tensor product couples feature blocks~\citep{kondor_n-body_2018}:
\begin{equation}
\label{eq:cg-product}
\bigl(\bm f^{(\ell_1)}\otimes\bm g^{(\ell_2)}\bigr)^{(\ell)}_m
=\sum_{m_1,m_2}C^{\ell m}_{\ell_1m_1,\ell_2m_2}\,f^{(\ell_1)}_{m_1}g^{(\ell_2)}_{m_2},
\qquad |\ell_1-\ell_2|\le\ell\le\ell_1+\ell_2.
\end{equation}
Learned weights, which may depend on invariants such as pair distances, combine channels along the allowed coupling paths \citep{thomasTensorFieldNetworks2018,geigerE3nnEuclideanNeural2022}. Pooling these tensor-product messages over neighbors gives the one-particle features $\bm a_i$. Atomic cluster expansions build higher-body-order features from $\bm a_i$ by symmetric tensor contractions, where correlation order $\nu$ gives terms of up to $\nu+1$ atoms \citep{drautz_atomic_2019,dusson_atomic_2022,batatiaMACEHigherOrder2023}.

\subsection{Fast Multipole Method}
\label{sec:fmm}
For scalar charges $q_s$, the Coulomb potential $V(\bm r_i)=(4\pi\varepsilon_0)^{-1}\sum_{s\ne i}q_s/|\bm r_i-\bm r_s|$ costs $\mathcal O(N^2)$ to evaluate directly. The FMM reduces this to $\mathcal O(N)$ by grouping particles in a spatial hierarchy and summarizing each distant group by a truncated multipole expansion \citep{greengard_fast_1987,cheng_fast_1999}. It rests on the Laplace expansion of $1/|\bm r_>-\bm r_<|$, where $\bm r_<=\bm r_s-\bm c_s$ is a particle's offset from its source center and $\bm r_>=\bm c_t-\bm c_s$ is the separation of source and target centers. This expansion splits each interaction into a regular factor $|\bm r_<|^{\ell}Y^{\star}_{\ell m}(\hat{\bm r}_<)$ and an irregular factor $Y_{\ell m}(\hat{\bm r}_>)/|\bm r_>|^{\ell+1}$. The regular factor builds and re-centers expansions. Particle-to-multipole (P2M) and multipole-to-multipole (M2M) aggregate source moments up the tree, and local-to-local (L2L) and local-to-particle (L2P) carry local expansions down and evaluate them. The irregular factor performs the single long-range step, multipole-to-local (M2L), which converts the multipoles of well-separated clusters into a local expansion. For the source list $\mathcal N_\tau(t)$ at stage $\tau\in\{\mathrm{P2M,M2M,M2L,L2L,L2P}\}$, each operator has the form:
\begin{equation}
\label{eq:fmm-translation}
\begin{aligned}
\Phi_t^{(\ell_t^\tau)}=\sum_{s\in\mathcal N_\tau(t)}\Bigl[\Phi_s^{(\ell_s^\tau)}\otimes_{c^\tau}\bigoplus_{\ell=0}^{p_\tau}R_\tau^{(\ell)}(r)\,\bm Y^{(\ell)}(\hat{\bm r})\Bigr]^{(\ell_t^\tau)}.
\end{aligned}
\end{equation}
Here $\Phi_s^{(\ell_s^\tau)}$ is the source state at stage $\tau$: the particle charge $q_s$ at P2M and the multipole or local coefficients thereafter. The target state $\Phi_t^{(\ell_t^\tau)}$ contains multipole coefficients at P2M and M2M, local coefficients at M2L and L2L, and the scalar far-field output after L2P. Adding the directly evaluated near field and applying the electrostatic prefactor gives the total Coulomb potential. The superscripts $\ell_s^\tau,\ell_t^\tau$ label degree-block collections rather than single degrees: $\ell_s^{\mathrm{P2M}}=\{0\}$ and $\ell_t^{\mathrm{P2M}}=\{0,\ldots,p\}$, with these reversed at L2P; both collections contain degrees $0,\ldots,p$ at M2M, M2L and L2L. Here $\bm r$ points from the target particle or center to the source particle or center, $r=\|\bm r\|$, and $\hat{\bm r}=\bm r/r$. The stage definitions are $p_\tau=p$ and $R_\tau^{(\ell)}(r)=r^\ell$ for P2M, M2M, L2L and L2P, and $p_\tau=2p$ and $R_\tau^{(\ell)}(r)=r^{-(\ell+1)}$ for M2L. M2L requires filter degrees through $2p$ because its input and output spatial degrees each extend to $p$, while stored expansions still stop at $p$. At L2P, filter degrees through $p$ contract with the local coefficients to yield a scalar. The vector $\bm Y^{(\ell)}$ contains all $2\ell+1$ real spherical-harmonic components. Scalar FMM has scalar particle charges and output potentials, but its intermediate degree-$\ell$ coefficient blocks transform as irreducible tensors. The weighted tensor product performs magnetic-component contractions and compatible-path sums, with fixed addition-theorem and normalization weights $c^\tau$ (Appendix~\ref{si:scalar-stage-weights}). Coulomb and scalar-FMM equivariance are proved in Appendix~\ref{si:coulomb-equivariance}.
The far-field series converges for separated source and target enclosures, $(R_s+R_t)/\|\bm r_>\|<\theta<1$. At fixed $p$, with a linear-size hierarchy, bounded leaf occupancy and bounded interaction lists, evaluation takes $\mathcal O(N)$ time and memory~\citep{cheng_fast_1999} (Appendix~\ref{si:linear-time}).

\section{Related Work}
\label{sec:related_work}

\paragraph{Scalar Long-Range Channels and Charge Equilibration}
Physics-informed corrections predict per-atom scalars and couple them through prescribed Coulomb or damped dispersion potentials \citep{yaoTensorMol01ModelChemistry2018,unkePhysNetNeuralNetwork2019,unke_spookynet_2021,pleForceFieldEnhancedNeuralNetwork2023,kabyldaMolecularSimulationsPretrained2024}. Reciprocal-space approaches communicate the same invariant channels through Ewald~\citep{ewaldBerechnungOptischerUnd1921} or particle-mesh~\citep{hockneyComputerSimulationUsing1988,dardenParticleMeshEwald1993} sums \citep{kosmalaEwaldbasedLongRangeMessage2023,wangNeuralP3MLongRange2024,chengLatentEwaldSummation2025,locheFastFlexibleLongrange2025}. Sum-of-Gaussians networks instead learn the reciprocal-space decay tail at near-linear cost via non-uniform FFTs \citep{jiMachineLearningInteratomicPotentials2025,jiAccurateLearningLongrange2026}, and prolate spheroidal mollification represents prescribed inverse-power channels with fewer Fourier modes \citep{liangProlateSpheroidalWave2026}. Charge equilibration redistributes globally coupled scalar charges \citep{rappeChargeEquilibrationMolecular1991,ghasemiInteratomicPotentialsIonic2015,khajehpashaCENT2ImprovedCharge2022,koFourthgenerationHighdimensionalNeural2021,fuchsLearningNonlocalMolecular2025,marufLearningLongRangeInteractions2025a,pleFoundationModelAccurate2025a}. Dense solves cost $\mathcal{O}(N^3)$ time and $\mathcal{O}(N^2)$ memory \citep{fuchsLearningNonlocalMolecular2025,gublerAcceleratingFourthGenerationMachine2024b}, and particle-mesh, iterative, and variational formulations improve scaling \citep{gublerAcceleratingFourthGenerationMachine2024b,kocerIterativeChargeEquilibration2025a,shaiduIncorporatingLongrangeElectrostatics2024,vondrakPushingChargeEquilibrationbased2025,kimLongrangeElectrostaticsMachine2026}. NEMORA transmits scalar and higher-degree tensor charges through a learned kernel at $\mathcal{O}(N)$ cost, on a real-space hierarchy that requires neither a Fourier grid nor periodicity.

\paragraph{Tensorial Transport and Many-Body Representations}
Density-based descriptors encode non-local environments through atomic density and potential representations \citep{grisafiIncorporatingLongrangePhysics2019,fallerDensityBasedLongRangeElectrostatic2024}. Multipolar corrections \citep{thurlemannAnisotropicMessagePassing2022,thurlemannLearningAtomicMultipoles2022,parkPolarizableAtomicMultipoles2026,batatia_mace-polar-1_2026,guoDifferentiableParticleMeshEwald2026} and multipole embeddings of environmental charges \citep{barrett_incorporating_2026} capture directional electrostatics but keep prescribed interaction kernels. Tensorial long-range extensions use fixed componentwise radial kernels or degree-preserving spectral filters \citep{rumiantsevLearningLongRangeRepresentations2026,zhangSO3equivariantReciprocalspaceNeural2026}. For equivariant linear maps conditioned only on invariants, Schur's lemma restricts learned weights to channel mixing within each degree \citep{geigerE3nnEuclideanNeural2022,rumiantsevLearningLongRangeRepresentations2026}. Global matrix-function approaches give non-local equivariant many-body representations, but at $\mathcal{O}(N^3)$ time and $\mathcal{O}(N^2)$ memory \citep{Batatia2024EquivariantNetworks}. NEMORA's operators condition on directions through $\bm Y^{(\ell)}(\hat{\bm r})$, so learned CG paths couple degrees, and symmetric contractions of the aggregated tensors add many-body terms at each level, at linear cost.

\paragraph{Global and Hierarchical Communication}
Global all-to-all attention learns non-local coupling at $\mathcal{O}(N^2)$ cost \citep{quRecipeScalableAttentionbased2026}. Scalar virtual-node aggregation is linear but transmits only invariants \citep{carusoExtendingRangeGraph2026}. Tensorial global attention is linear at a fixed spherical quadrature, but its rotational equivariance is exact only up to the quadrature degree, which must grow with frequency and spatial extent \citep{frankMachineLearningGlobal2026}. Equivariant atom--fragment attention transmits directional features over an enlarged finite cutoff \citep{wangScalableMachineLearning2026}. Graph hierarchies communicate across scales through coarsening, message passing and pooling, including equivariant constructions \citep{yingHierarchicalGraphRepresentation2019,rampasekHierarchicalGraphNeural2021,ngoMultiresolutionGraphTransformers2023,hanEquivariantGraphHierarchyBased2022,hyMultiresolutionEquivariantGraph2023,xuDualEquiNetDualSpaceHierarchical2025,soleMachineLearningMultiscale2026}. Tree-based architectures combine spatial partitioning with local attention \citep{zhdanovErwinTreebasedHierarchical2025}, fragment-based corrections couple atoms and chemistry-defined fragments \citep{liLongShortRangeMessagePassingPhysicsInformed2024}, and hierarchical particle models canonicalize two-level local and global graphs in PCA frames, with predefined groups aggregated into averaged invariant nodes \citep{linotHierarchicalEquivariantGraph2025}. NEMORA instead builds its hierarchy on the FMM well-separation criterion, so its analytical limit inherits multipole convergence guarantees.

\paragraph{FMM-Inspired Neural Architectures}
Multilevel graph kernels and $\mathcal{H}^2$-matrix~\citep{hackbuschSparseMatrixArithmetic1999} networks learn non-local operators on graphs and point clouds \citep{li_multipole_2020,sushnikovaFMMNetNeuralNetwork2022}. Neural FMMs learn hierarchical translation stages for PDE solution operators, including point-cloud variants \citep{fogniniLearningGreensOperators2025,fogniniCloudNFMM2026}, and multipole-inspired attention cuts global attention cost \citep{colagrandeLinearAttentionGlobal2025,kangFastMultipoleAttention2023}. Differentiable FMM solvers give gradients for inverse problems with prescribed kernels \citep{stuckerJZFMM2026}. For atomistic potentials, NEMORA translates multipole features between well-separated regions of an adaptive hierarchy with equivariant tensor transport, learned angular coupling, modulated analytical radial factors, and many-body contractions.

\section{NEMORA Architecture}
\paragraph{NEMORA Framework}
NEMORA makes the FMM operators (P2M, M2M, M2L, L2L, L2P) learnable, passing equivariant tensor features up to a chosen $\ell_{\max}$ through an adaptive hierarchy. Unless stated otherwise, NEMORA is attached to a MACE backbone~\citep{batatiaMACEHigherOrder2023}.

\subsection{Neural Equivariant Multipole Operators}

\paragraph{Learnable equivariant tensor charges.}
We replace the scalar FMM charges $q_j$ with an equivariant tensor charge $\bm Q_j^{(\ell_Q)}$. In a coupled real spherical-harmonic basis, each stage has the same form as Equation~\ref{eq:fmm-translation}, with tensor-valued source states and the corresponding fixed re-coupling weights. The component-wise Coulomb kernel is unchanged:
\begin{equation}
\label{eq:equivariant-fmm-translation}
\begin{aligned}
\bm\Phi_t^{(\ell_t^\tau)}=\sum_{s\in\mathcal N_\tau(t)}\Bigl[\bm\Phi_s^{(\ell_s^\tau)}\otimes_{c^\tau}\bigoplus_{\ell=0}^{p_\tau}R_\tau^{(\ell)}(r)\,\bm Y^{(\ell)}(\hat{\bm r})\Bigr]^{(\ell_t^\tau)}.
\end{aligned}
\end{equation}
At P2M, the particle input has $\ell_s^{\mathrm{P2M}}=\{\ell_Q\}$. For each spatial order $n=0,\ldots,p$, the target expansion retains every coupled degree $L$ satisfying $|\ell_Q-n|\le L\le\ell_Q+n$. These blocks form both the source and target representations for M2M, M2L and L2L, with repeated degrees arising from different spatial orders retained as distinct multiplicity channels. At L2P, the source contains these expansion blocks and the target returns to $\ell_t^{\mathrm{L2P}}=\{\ell_Q\}$. Thus $\ell_Q$ labels the original charge, not every intermediate source degree. The weighted tensor product performs the magnetic-component contractions and sums over compatible input blocks and coupling paths; $\bigoplus_\ell$ retains the harmonic blocks, rather than summing them to a scalar. Kernel and expansion equivariance are proved in Appendix~\ref{si:tensor-equivariance}, with fixed stage weights and phase conventions in Appendix~\ref{si:equivariant-fmm}. This is an analytical subset of NEMORA: long-range operators that approximate an analytical long-range kernel to transmit learned charges, which are scalar monopoles at $\ell_Q=0$ and tensor-valued sources at higher degree, with physical atomic multipoles requiring kernel derivatives.
These tensor-valued charges, which can be read as learned atomic multipoles, add angular degrees of freedom that distributed scalar charges lack. The source degree $\ell_Q$ sets expressivity and the expansion order $p$ sets accuracy. Latent-Ewald-based methods~\citep{chengLatentEwaldSummation2025,rumiantsevLearningLongRangeRepresentations2026} evaluate learned charges with Ewald sums, whereas analytical NEMORA uses the adaptive, $\mathcal O(N)$ FMM algorithm.

\subsection{MLIP Messages and Multipole Expansions}
Centered on atom $t$ with backbone features as sources, the P2M coupling underlying Equation~\ref{eq:equivariant-fmm-translation} takes the form of an equivariant MLIP message:
\begin{equation}
\label{eq:local-multipole-analogy}
\bm a_t^{(\tilde\ell_t)}
=\sum_{s\in\mathcal N_t}\Bigl[
\bm h_s^{(\tilde\ell_s)}
\otimes_\theta
\bigoplus_{\ell=0}^{\ell_{\max}}
\bm R_\theta^{(\ell)}(r)\,
\bm Y^{(\ell)}(\hat{\bm r})
\Bigr]^{(\tilde\ell_t)}.
\end{equation}
Here $\bm r=\bm r_s-\bm r_t$, $r=\|\bm r\|$, $\hat{\bm r}=\bm r/r$, and $\mathcal N_t$ is the local neighborhood of $t$. The tildes in $\tilde\ell_s,\tilde\ell_t$ denote architecture-configured source and target degree-block collections and multiplicities, rather than analytically prescribed ones. This is the pooled tensor-product message of Section~\ref{sec:mlips}, read as a learnable local multipole expansion: the source feature $\bm h_s$ plays the role of the tensor charge $\bm Q_s$, the channel-wise radial weights $\bm R_\theta^{(\ell)}(r)$ replace the regular radial factor $r^\ell$, and the learned path weights replace the fixed analytical coupling weights. Setting $h_s^{(0)}=q_s$ and $\bm R_\theta^{(\ell)}(r)=r^\ell\bm 1$, with matching harmonic normalization, path weights and retained degrees, recovers the classical moments about atom $t$. As in Equation~\ref{eq:equivariant-fmm-translation}, the concatenated harmonic blocks retain their angular degrees, and the tensor product combines compatible coupling paths into the configured target blocks. We therefore read the pooled features of the one-particle basis~\citep{batatia_design_2025} as learnable local multipole expansions, whose symmetric contractions build body order, and let the FMM carry them across length scales. This correspondence is NEMORA's design principle: local message passing already makes P2M learnable, and NEMORA does the same for every other FMM stage.

\subsection{Learnable Equivariant Multipole Operators}
Equations~\ref{eq:fmm-translation}--\ref{eq:equivariant-fmm-translation} describe individual analytical FMM stages, and Equation~\ref{eq:local-multipole-analogy} gives their analogy with local message-passing. Following this correspondence, we replace each analytical operator with a learnable equivariant one:
\begin{equation}
\label{eq:nemora-translation}
\bm\Psi_t^{(\tilde\ell_t^\tau)}
=\sum_{s\in\mathcal N_\tau(t)}
\alpha_{\theta,st}^{(\tilde\ell_t^\tau)}
\Bigl[
\bm W^V\bm\Psi_s^{(\tilde\ell_s^\tau)}
\otimes_{\theta^\tau}
\bigoplus_{\ell=0}^{p}
R_\tau^{(\ell)}(r)\,
\bm R_\theta^{(\ell)}(r)\,
\bm Y^{(\ell)}(\hat{\bm r})
\Bigr]^{(\tilde\ell_t^\tau)}.
\end{equation}
Here $\bm\Psi_s$ contains the tensor charges at P2M and the learned multipole or local features thereafter. The target $\bm\Psi_t$ contains the corresponding updated features, with L2P returning particle-level equivariant features for energy fusion and readout, rather than a prescribed Coulomb potential. The tilded source and target degree-block collections are architecture-configured as in Equation~\ref{eq:local-multipole-analogy}, with $\tau$ identifying the FMM operator stage. The displacement $\bm r$, radius $r$ and scalar radial factor $R_\tau^{(\ell)}(r)$ follow the stage definitions in Section~\ref{sec:fmm}; numerical implementations normalize and regularize the radial coordinates.

The map $\bm W^V$ mixes channels within each degree, while $\otimes_{\theta^\tau}$ learns weights over compatible CG coupling paths, incorporating the magnetic-component sums and generalizing the fixed analytical stage coefficients. Harmonic-normalization and recoupling factors can be absorbed into these weights on the represented paths. The channelwise function $\bm R_\theta^{(\ell)}(r)$ provides a bounded multiplicative modulation of the analytical radial factor, evaluated using a sinc--Bessel basis~\citep{gasteiger_dimenet_2020}. As in the MLIP equation, this notation includes the radial-basis evaluation and does not imply parameter sharing between models or stages. A gate $\alpha_{\theta,st}^{(\ell_M)}=\bm\phi_\theta(\mathcal G_{st}\oplus\mathcal M_{st}\oplus\mathcal A_{st})^{(\ell_M)}$ weights each source using geometric invariants $\mathcal G_{st}$, normalized features $\mathcal M_{st}$ and alignment invariants $\mathcal A_{st}$, acting blockwise for each target degree $\ell_M$ in $\tilde\ell_t^\tau$. It serves as a learned analogue of damping and polarization factors~\citep{tangImprovedSimpleModel1984,ponderCurrentStatusAMOEBA2010,stoneTheory2013} (Appendix~\ref{si:learnable-multipole-gate}).

Setting $\bm W^V=\bm I$, $\bm R_\theta^{(\ell)}=\bm 1$, $\alpha_\theta=1$ and the analytical path weights formally recovers the analytical operator when the required intermediate representations, filter degrees and normalizations are retained. At each downward level, symmetric contractions of the pooled M2L field add explicit many-body features (MBC; Appendices~\ref{si:symcontract} and~\ref{si:symcontract-proof}), and a learnable Kelvin transform adds a continuum boundary response (LKT/GKT; Appendices~\ref{si:kelvin} and~\ref{si:kelvin-local}). Under periodic boundary conditions, a finite-shell, degree-diagonal correction at the root accounts for periodic images (Equation~\ref{eq:pbc-tinfoil}).

\subsection{Energy Fusion and Readout}
NEMORA acts at chosen layers $t^\star$ of the backbone (Section~\ref{sec:mlips}) and takes its tensor charges from these features. An input projection mixes channels within each degree, $\bm Q_i^{(\ell_Q)}=\bm W_{\mathrm{in}}^{(\ell_Q)}\bm h_i^{(\ell_Q,t^\star)}$, and the operators couple degrees along the hierarchy. A degree-preserving projection $\bm W_{\mathrm{out}}^{(\ell_Q)}$ and an invariant gate return the far-field update $\Delta\bm h_i^{(\ell_Q,t^\star)}$. The corrected features replace $\bm h_i^{(\ell_Q,t^\star)}$ in the layer-$t^\star$ readout and all later layers, and a scalar readout of the update adds $\Delta E_i^{(t^\star)}$ to the energy:
\begin{equation}
\label{eq:tensor-charge-routing}
\tilde{\bm h}_i^{(\ell_Q,t^\star)}=\bm h_i^{(\ell_Q,t^\star)}+\Delta\bm h_i^{(\ell_Q,t^\star)},\qquad
E_i=\sum_{t=1}^T\mathcal R_{t}(\bm h_i^{(t)})+\underbrace{\mathrm{MLP}\bigl(\Delta\bm h_i^{(0,t^\star)}\bigr)}_{\Delta E_i^{(t^\star)}}.
\end{equation}
The equivariance and invariant-readout arguments are given in Appendices~\ref{si:learned-equivariance} and~\ref{si:readout-proof}.

\subsection{Adaptive $E(3)$-Equivariant Hierarchy}
\label{sec:hierarchy}
The hierarchy is built by top-down adaptive subdivision~\citep{barnesHierarchicalLogForcecalculation1986,warrenAstrophysicalNbodySimulations1992,yingKernelindependentAdaptiveFast2004}. Each cluster splits along the PCA eigenframe of its centered atomic offsets, or the Gram--Schmidt lattice frame under PBC. An axis is active when its eigenvalue ratio $\lambda_k/\lambda_{\max}$ exceeds a threshold, and the fanout $2^{n_{\mathrm{active}}}$ gives binary, quadtree, or octree splits for chain, slab, and bulk geometries. A cluster becomes a leaf when its predicted child diameter falls below the backbone receptive field $n_{\mathrm{SR}}r_{\mathrm{cut}}$ or its atom count drops below a threshold. Equivariance requires frame selection and cluster assignment to transform with the geometry, including periodic cells, with symmetry-consistent handling of degenerate eigenframes and sign or partition ties (Appendix~\ref{si:coulomb-equivariance}).
At level $L$, the near set is $\mathrm{Near}_L(B)=\{A:\|\bm c_A-\bm c_B\|_{\mathrm{MIC}}\le\rho_{\mathrm{near}}(R_A+R_B)\}$ with $\rho_{\mathrm{near}}\ge1$, so every far pair has separation ratio $(R_A+R_B)/\|\bm c_A-\bm c_B\|_{\mathrm{MIC}}<1/\rho_{\mathrm{near}}\le1$, ensuring convergence of the analytical expansion. The M2L list $\mathcal{I}_L(B)$ holds the children of the parent's near clusters that are not near $B$. Near leaf pairs also use M2L, weighted by an envelope of their gap distance $d_{\min}=\max(\|\bm c_A-\bm c_B\|_{\mathrm{MIC}}-R_A-R_B,0)$ so that the short- and long-range channels hand off smoothly \citep{gasteiger_dimenet_2020}. Evaluation is linear in $N$ when the hierarchy has $\mathcal O(N)$ nodes and bounded interaction lists (Appendix~\ref{si:linear-time}).

\section{Results}
\label{sec:results}

We evaluate NEMORA on molecular and periodic systems that probe different forms of nonlocal interaction. Unless otherwise stated, tables report means over three independent training runs, with energy in meV/atom and force in meV/\AA. Training details are in Appendix~\ref{si:model-settings}.

\subsection{Cumulene}
\label{sec:cumulene-results}

Two cumulene experiments probe long-range electronic delocalization and its dependence on geometry~\citep{Batatia2024EquivariantNetworks}. The first trains on $n_C=3$--$10,13,14$ and tests on these lengths (in-distribution, ID), on the unseen lengths 11 and 12 (semi-out-of-distribution, SOD), and on the longer chains 15 and 16 (out-of-distribution, OOD). The second covers the end-group torsional angle energy profile at a fixed chain length as well as the chain-length dependence of the energy at fixed torsional angle on a separate 3--64 carbon set. Both of these benchmarks test the ability of models to effectively transmit angular information over long range and learn the effect that local angular structure has on the long-range interactions.

\begin{table}[!ht]
\centering\small
\setlength{\tabcolsep}{3.5pt}
{\def\twoline#1#2{\begin{tabular}[b]{@{}c@{}}#1\\#2\end{tabular}}%
\caption{Cumulene chain-length, nonlocal charge-transfer, and MD22 benchmark errors. Each entry is E RMSE (meV/atom)\,/\,F RMSE (meV/\AA). Cumulene splits are defined in Section~\ref{sec:cumulene-results}. Nonlocal charge-transfer datasets and training protocols are detailed in Appendices~\ref{si:datasets} and~\ref{si:system-settings}. MD22 entries are means over three seeds; protocol exceptions are given in Appendix~\ref{si:md22}. In each row and metric, lowest means are bold and second-lowest underlined (ties retained).}
\label{tab:cumulene_splits}
\label{tab:cumulene-forces}
\label{tab:ko}
\label{tab:ko-forces}
\label{tab:ko-literature-energy}
\label{tab:md22}
\label{tab:md22-energy}
\label{tab:md22-forces}
\begin{adjustbox}{max width=\linewidth}
\begin{tabular}{@{}l*{6}{r}@{\hspace{8pt}}r@{}}
\toprule
System & MACE & MACE--LES & MACE--RANGE & LOREM & EFA & MACE--MFN & \textbf{NEMORA (Ours)} \\
 & {\scriptsize E RMSE\,/\,F RMSE} & {\scriptsize E RMSE\,/\,F RMSE} & {\scriptsize E RMSE\,/\,F RMSE} & {\scriptsize E RMSE\,/\,F RMSE} & {\scriptsize E RMSE\,/\,F RMSE} & {\scriptsize E RMSE\,/\,F RMSE} & {\scriptsize E RMSE\,/\,F RMSE} \\
\midrule
\multicolumn{8}{@{}l}{\textit{Cumulene}} \\
\quad ID & $16.898$\,/\,$85.07$ & $17.235$\,/\,$84.82$ & $19.505$\,/\,$95.60$ & $6.756$\,/\,$29.36$ & $5.918$\,/\,$30.28$ & $\underline{3.231}$\,/\,$\mathbf{13.83}$ & $\mathbf{3.203}$\,/\,$\underline{26.85}$ \\
\quad SOD & $16.709$\,/\,$102.16$ & $16.560$\,/\,$102.17$ & $18.796$\,/\,$112.98$ & $27.251$\,/\,$166.70$ & $27.884$\,/\,$158.17$ & $\mathbf{0.643}$\,/\,$\mathbf{14.25}$ & $\underline{0.744}$\,/\,$\underline{47.87}$ \\
\quad OOD & $11.994$\,/\,$72.26$ & $11.180$\,/\,$72.36$ & $16.411$\,/\,$94.90$ & $25.008$\,/\,$133.06$ & $20.810$\,/\,$93.92$ & $\mathbf{2.812}$\,/\,$\mathbf{15.28}$ & $\underline{3.770}$\,/\,$\underline{57.01}$ \\
\addlinespace[3pt]
\multicolumn{8}{@{}l}{\textit{Nonlocal charge transfer}} \\
\quad Carbon chain & $0.425$\,/\,$15.00$ & $0.524$\,/\,$20.15$ & $0.932$\,/\,$12.38$ & $0.640$\,/\,$4.62$ & $0.502$\,/\,$5.64$ & $\mathbf{0.156}$\,/\,$\underline{1.90}$ & $\underline{0.175}$\,/\,$\mathbf{1.88}$ \\
\quad NaCl & $1.600$\,/\,$43.89$ & $0.380$\,/\,$7.21$ & $0.622$\,/\,$2.93$ & $\mathbf{0.243}$\,/\,$\underline{1.76}$ & $0.861$\,/\,$18.13$ & $\underline{0.361}$\,/\,$2.75$ & $0.412$\,/\,$\mathbf{1.75}$ \\
\quad Au$_2$/MgO & $2.385$\,/\,$110.07$ & $1.696$\,/\,$21.85$ & $\underline{0.207}$\,/\,$12.18$ & $0.235$\,/\,$\mathbf{4.64}$ & $1.890$\,/\,$13.03$ & --- & $\mathbf{0.197}$\,/\,$\underline{5.26}$ \\
\addlinespace[3pt]
\multicolumn{8}{@{}l}{\textit{MD22}} \\
\quad Ac-Ala3-NHMe & $0.405$\,/\,$7.44$ & $\underline{0.348}$\,/\,$\underline{7.08}$ & $0.461$\,/\,$12.45$ & $1.367$\,/\,$13.96$ & $1.146$\,/\,$53.63$ & --- & $\mathbf{0.323}$\,/\,$\mathbf{6.93}$ \\
\quad DHA & $\underline{0.575}$\,/\,$\mathbf{5.64}$ & $0.586$\,/\,$5.87$ & $0.747$\,/\,$9.08$ & $1.273$\,/\,$9.74$ & $0.735$\,/\,$41.51$ & --- & $\mathbf{0.455}$\,/\,$\underline{5.70}$ \\
\quad Stachyose & $\underline{0.375}$\,/\,$6.64$ & $0.463$\,/\,$\mathbf{6.24}$ & $\mathbf{0.309}$\,/\,$10.15$ & $1.271$\,/\,$12.36$ & $0.604$\,/\,$53.19$ & --- & $0.401$\,/\,$\underline{6.45}$ \\
\quad AT-AT & $1.168$\,/\,$10.78$ & $1.005$\,/\,$\underline{8.04}$ & $\underline{0.739}$\,/\,$15.88$ & $1.475$\,/\,$14.90$ & $1.043$\,/\,$63.83$ & --- & $\mathbf{0.340}$\,/\,$\mathbf{7.85}$ \\
\quad AT-AT-CG-CG & $0.766$\,/\,$10.26$ & $0.927$\,/\,$\mathbf{8.12}$ & $0.647$\,/\,$16.73$ & $1.241$\,/\,$18.10$ & $\underline{0.606}$\,/\,$63.28$ & --- & $\mathbf{0.341}$\,/\,$\underline{8.60}$ \\
\quad buckyball-catcher & $1.099$\,/\,$8.22$ & $1.087$\,/\,$\underline{7.96}$ & $1.200$\,/\,$17.91$ & $1.350$\,/\,$10.42$ & $\mathbf{0.731}$\,/\,$102.00$ & --- & $\underline{1.081}$\,/\,$\mathbf{7.54}$ \\
\quad Nanotube & $1.406$\,/\,$\underline{24.75}$ & $1.806$\,/\,$28.91$ & $5.551$\,/\,$45.57$ & $\underline{1.361}$\,/\,$28.07$ & $\mathbf{1.086}$\,/\,$174.25$ & --- & $1.362$\,/\,$\mathbf{22.32}$ \\
\bottomrule
\end{tabular}
\end{adjustbox}}
\end{table}

The first section of Table~\ref{tab:cumulene_splits} shows that NEMORA substantially improves energy and force predictions over short-range MACE and scalar long-range models, as well as equivariant long-range models that are sub-cubic in complexity. NEMORA has the lowest ID energy error and is second only to MACE--MFN, which scales as $\mathcal{O}{N^3}$, on the remaining splits and metrics. NEMORA has lower energy and similar force ID errors to LOREM and EFA, the other tensor long-range models, but on unseen chain lengths LOREM and EFA degrade to higher errors than short-range MACE, whereas NEMORA remains well below the MACE error. This indicates that NEMORA is learning a more effective representation of the features that determine the energy in the long-range cumulene system. Degradation of performance upon extrapolation in chain-length can indicate overfitting to training data, rather than learning correct representations. MACE--MFN remains the most accurate on these splits except for ID energy, at the cost of cubic time and quadratic memory scaling (Figure~\ref{fig:complexity}). Together with the additional scalar NEMORA ablations provided in Appendix~\ref{si:ablation-protocol}, these results indicate that transmitting angular information is vital to performance in cumulene.

\begin{figure}[!htbp]
\centering
\includegraphics[width=\linewidth]{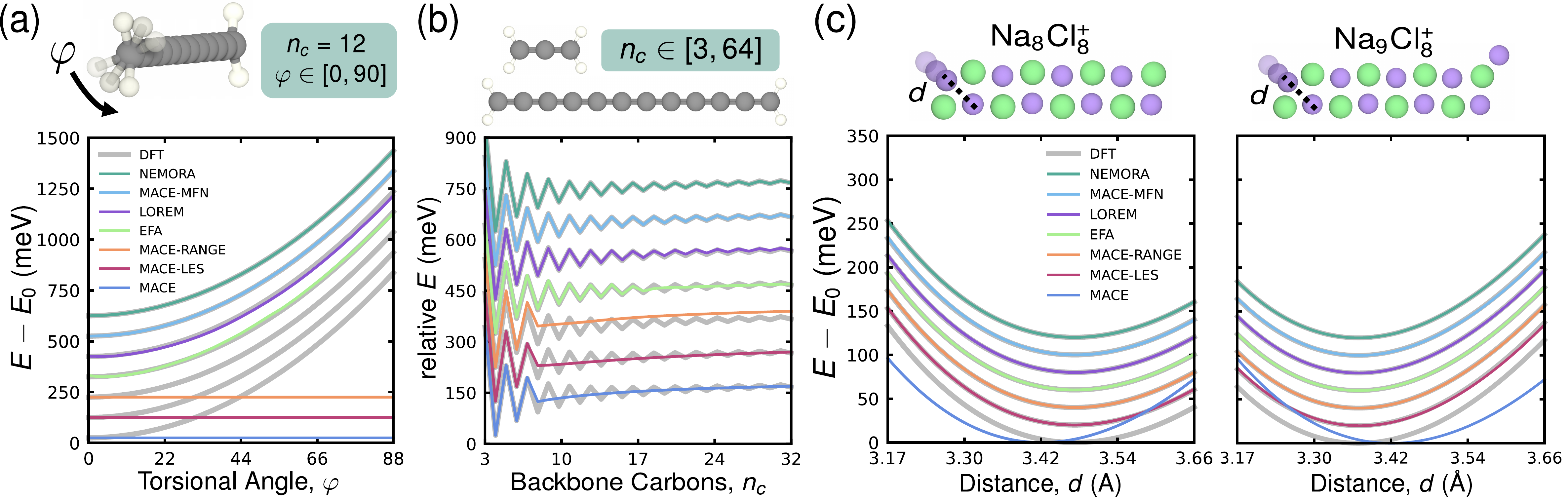}
\vspace{-0.5cm}
\caption{Cumulene angular and chain-length benchmarks~\citep{Batatia2024EquivariantNetworks} and a charged sodium chloride cluster benchmark~\citep{koFourthgenerationHighdimensionalNeural2021}. (a) Torsional energy profile for a 12-carbon chain. (b) Energy as a function of chain length at $\varphi=5^\circ$. (c) Energy versus Na displacement in Na$_8$Cl$_8^+$ and Na$_9$Cl$_8^+$, where an extra neutral Na atom $\sim$21~\AA\ away changes the binding profile. Gray curves are DFT references, and model curves are shifted vertically for visual clarity.}
\label{fig:cumulene}
\end{figure}

Figure~\ref{fig:cumulene}(a,b) presents the separate angular and chain-length experiments. Rotating the terminal groups changes the energy significantly, even when the terminal groups lie beyond the short-range receptive field. Short-range MACE and the scalar long-range models, MACE--LES and MACE--RANGE, predict flat torsional profiles, missing this angular dependence. The higher-order long-range models are able to follow the DFT angular profile, with NEMORA and MACE--MFN being closest. Similar results are obtained for the even--odd (zig-zag) energy variation with chain length, with short-range models and scalar long-range models failing to capture the energy of the different angular conformations beyond their receptive field. Long-range models with higher-order angular information, however, are able to capture the zig-zag profile to differing degrees, with NEMORA and MACE--MFN matching the DFT profile best over the full range of chain lengths while the torsional profile of LOREM and EFA degrade, partially and completely, respectively, after seventeen carbons. The corresponding energy and force errors are provided in Table~\ref{tab:cumulene-additional} in the Appendix.

\subsection{Nonlocal charge transfer}
\label{sec:ko-results}

The nonlocal charge transfer benchmark datasets introduced by \citet{koFourthgenerationHighdimensionalNeural2021} test the effect of nonlocal charge transfer in different contexts. In the C$_{10}$H$_2$/C$_{10}$H$_3^+$ system, protonation changes the charge distribution and electronic structure along a conjugated carbon chain. In the Na$_8$Cl$_8^+$/Na$_9$Cl$_8^+$ (NaCl) cluster, an additional neutral sodium atom affects the binding of another sodium atom on the opposite side of the cluster. The Au$_2$/MgO dataset tests the effect of substrate doping on an adsorbed gold dimer under periodic boundary conditions.

As shown in the second section of Table~\ref{tab:ko}, NEMORA substantially improves upon MACE across all three datasets and, among the long-range models, ranks first or second on five out of six metrics, with the lowest force errors on the carbon chain and NaCl and the lowest energy error on Au$_2$/MgO. While some other long-range models do compete for accuracy on these benchmarks, we stress the ability of NEMORA to achieve this state-of-the-art accuracy while retaining $\mathcal{O}(N)$ time and memory scaling in both periodic and non-periodic systems alike. Figure~\ref{fig:cumulene}(c) illustrates the energy profile along the sodium removal pathway for the NaCl cluster, with and without the additional neutral sodium on the opposing side. Unsurprisingly, short-range models are unable to accurately capture the shape and minimum of the energy profile in both cases, while all models with long-range information generally capture these effects. These benchmarks combined with the cumulene benchmarks indicate that NEMORA is capable of capturing different types of long-range interactions ranging from highly angular (cumulene) to scalar charge redistribution (nonlocal charge transfer).

Table~\ref{tab:ko} excludes charge-equilibration and explicitly electrostatic models (CELLI, 4G--HDNNP and SpookyNet)~\citep{fuchsLearningNonlocalMolecular2025,koFourthgenerationHighdimensionalNeural2021,unke_spookynet_2021}, which add reference-charge terms to the loss or use supervision through total charge and spin inputs. Using these extra inputs and loss terms, CELLI and SpookyNet reach lower NaCl and Au$_2$/MgO energy errors but perform worse than short-range benchmarks on cumulene (Table~\ref{tab:charge-equilibration}, Appendix~\ref{si:additional-results}).

\subsection{MD22}
\label{sec:md22}

Table~\ref{tab:md22} also includes MACE--RANGE, LOREM and EFA evaluated on the same MD22 test splits, with three seeds per model and system. Their checkpoint-selection and numerical-evaluation protocols are given in Appendix~\ref{si:md22}. MACE--MFN is omitted on MD22 because the native training campaign was stopped before completion owing to computational cost; early-epoch timings on L40S/H200 indicated multi-day runs. Training also developed exploding losses on the buckyball catcher system.

To test whether NEMORA degrades accuracy when a short-range model already performs well owing to relatively weak long-range interactions, we train and evaluate it on all seven MD22 systems~\citep{chmielaAccurateGlobalMachine2023}, comparing MACE, MACE--LES, MACE--RANGE, LOREM, EFA, and NEMORA (Appendix~\ref{si:md22}). As shown in Table~\ref{tab:md22}, NEMORA lowers energy errors on six systems and force errors on six systems relative to MACE, and achieves the lowest energy and force error among MACE, MACE--LES and NEMORA on four systems. The largest gains occur in the DNA base-pair systems, where energy errors fall from 1.168 to 0.340~meV/atom (AT-AT) and from 0.766 to 0.341~meV/atom (AT-AT-CG-CG). For the nanotube system, NEMORA reduces the force error from 24.75 to 22.32~meV/\AA, whereas MACE--LES increases it to 28.91~meV/\AA. MACE--RANGE, LOREM, and EFA also have higher force errors on all seven systems relative to the short-range baseline and higher energy errors on four to seven systems. The two NEMORA performance regressions relative to MACE are minor: an increase of 0.06~meV/\AA\ in force on DHA and 0.026~meV/atom in energy on Stachyose. NEMORA therefore generally preserves or improves the accuracy of the short-range backbone on these systems. These results are consistent with a largely non-disruptive extension, contributing when long-range interactions become non-negligible rather than interfering with the underlying short-range backbone's learned representations.

\subsection{Portability across backbones and the role of angular features}
\label{sec:portability}

\begin{table}[!htbp]
\centering\small
\setlength{\tabcolsep}{3pt}
\caption{Backbone and source-degree ablations, mean validation E\,/\,F RMSE (meV/atom\,/\,meV/\AA). SR is the short-range backbone, and $\ell_Q=0$ and $\ell_Q=2$ are scalar and degree-two NEMORA sources, lifted for scalar and untyped hosts. Bold and underline mark the lowest and second-lowest of the three settings per host, dataset and metric (Appendix~\ref{si:backbone-protocol}).}
\label{tab:backbone-nacl}
\label{tab:backbone-cumulene}
\begin{adjustbox}{max width=\linewidth}
\begin{tabular}{llrrr@{\hspace{10pt}}rrr}
\toprule
 & & \multicolumn{3}{c}{NaCl} & \multicolumn{3}{c}{Cumulene} \\\cmidrule(lr){3-5}\cmidrule(lr){6-8}
Backbone & Host & SR & $\ell_Q=0$ & $\ell_Q=2$ & SR & $\ell_Q=0$ & $\ell_Q=2$ \\
\midrule
MACE \citep{batatiaMACEHigherOrder2023} & Native & $1.597\,/\,43.92$ & $\underline{0.457}\,/\,\underline{3.45}$ & $\mathbf{0.406}\,/\,\mathbf{1.80}$ & $\underline{13.672}\,/\,109.82$ & $13.767\,/\,\underline{105.93}$ & $\mathbf{3.776}\,/\,\mathbf{65.93}$ \\
 & Scalar & $1.600\,/\,44.51$ & $\underline{0.475}\,/\,\underline{7.56}$ & $\mathbf{0.437}\,/\,\mathbf{1.88}$ & $\underline{52.466}\,/\,196.28$ & $53.357\,/\,\underline{195.93}$ & $\mathbf{4.134}\,/\,\mathbf{77.32}$ \\
NequIP \citep{batznerE3equivariantGraphNeural2022} & Native & $1.565\,/\,43.89$ & $\underline{0.664}\,/\,\underline{6.39}$ & $\mathbf{0.258}\,/\,\mathbf{1.85}$ & $\underline{54.699}\,/\,\underline{203.77}$ & $55.097\,/\,206.56$ & $\mathbf{3.916}\,/\,\mathbf{46.06}$ \\
 & Scalar & $1.774\,/\,62.70$ & $\underline{0.561}\,/\,\underline{11.40}$ & $\mathbf{0.242}\,/\,\mathbf{2.23}$ & $\underline{55.939}\,/\,239.69$ & $56.752\,/\,\underline{239.29}$ & $\mathbf{5.567}\,/\,\mathbf{81.59}$ \\
ViSNet \citep{Wang2024VisNet} & Native & $1.533\,/\,42.02$ & $\underline{0.230}\,/\,\underline{4.04}$ & $\mathbf{0.145}\,/\,\mathbf{1.67}$ & $\underline{14.630}\,/\,\underline{120.86}$ & $25.020\,/\,150.24$ & $\mathbf{3.723}\,/\,\mathbf{46.35}$ \\
 & Scalar & $1.711\,/\,59.46$ & $\underline{0.876}\,/\,\underline{8.12}$ & $\mathbf{0.302}\,/\,\mathbf{2.30}$ & $\underline{56.132}\,/\,240.12$ & $56.871\,/\,\underline{239.85}$ & $\mathbf{4.150}\,/\,\mathbf{70.53}$ \\
PET \citep{pozdnyakov_smooth_2023} & Untyped & $1.524\,/\,38.91$ & $\mathbf{0.260}\,/\,\underline{5.64}$ & $\underline{0.485}\,/\,\mathbf{4.55}$ & $54.814\,/\,241.43$ & $\underline{45.183}\,/\,\underline{221.56}$ & $\mathbf{9.363}\,/\,\mathbf{112.11}$ \\
\bottomrule
\end{tabular}
\end{adjustbox}
\end{table}

Although we attach NEMORA to MACE throughout this work for consistency, it can be attached to a variety of architecturally distinct short-range architectures. To demonstrate this ability, we test MACE, NequIP, ViSNet, and PET~\citep{batatiaMACEHigherOrder2023,batznerE3equivariantGraphNeural2022,Wang2024VisNet,pozdnyakov_smooth_2023}, while varying the degree of the short-range features and the NEMORA expansions (Table~\ref{tab:backbone-nacl}). For MACE, NequIP, and ViSNet, we compare the native equivariant backbone with a scalar-only variant. Each is trained without NEMORA, with scalar NEMORA ($\ell_Q=0$), and with degree-two NEMORA ($\ell_Q=2$). When the backbone supplies only scalar features, a geometry-based lift constructs the tensorial sources by performing a tensor product of the source features with the concatenated spherical harmonics in the P2M operator. The same interface allows attachment to the rotationally unconstrained PET architecture.

On the charged NaCl cluster (Na$_8$Cl$_8^+$/Na$_9$Cl$_8^+$), where the nonlocal interaction is predominantly electrostatic, scalar NEMORA ($\ell_Q=0$) already captures most of the improvement over the short-range models. Degree-two sources ($\ell_Q=2$) further reduce force errors for every backbone, and energy errors for all but PET. On cumulene, scalar long-range communication gives little or no improvement, while degree-two sources reduce errors substantially for every backbone. Surprisingly, the geometry-based lift recovers much of this improvement when the short-range backbone contains only scalar features. These results show that the NEMORA channel can supply angular information missing from the backbone, across all four backbones tested. Interestingly, PET is not equivariant by construction, but the higher degree NEMORA attachment is able to significantly improve its performance by incorporating an equivariant long-range correction.

\subsection{Periodic ions and water}
\label{sec:periodic-results}

\begin{table}[!htbp]
\centering\small
\setlength{\tabcolsep}{3pt}
\caption{Periodic-ion and water test errors, E RMSE (meV/atom)\,/\,F RMSE (meV/\AA). Ion models are trained on $N=256$--$1024$. Water models are trained on 128 and 256 molecules. The analytical NEMORA with scalar ($\ell_Q=0$) or scalar-plus-dipole ($\ell_Q=1$) sources is used (Appendix~\ref{si:latent-fmm}).}
\label{tab:periodic-ions}
\label{tab:water-nemora}
\begin{adjustbox}{max width=\linewidth}
\begin{tabular}{@{}lrrrr@{}}
\toprule
System & MACE & MACE--LES & NEMORA ($\ell_Q=0$) & NEMORA ($\ell_Q=1$) \\
\midrule
\multicolumn{5}{@{}l}{\textit{Periodic ions}} \\
\quad $N=256$ & $43.43$\,/\,$168.6$ & $0.308$\,/\,$9.80$ & $\underline{0.284}$\,/\,$\underline{9.64}$ & $\mathbf{0.237}$\,/\,$\mathbf{8.41}$ \\
\quad $N=512$ & $24.59$\,/\,$195.8$ & $\underline{0.203}$\,/\,$10.01$ & $0.211$\,/\,$\underline{9.83}$ & $\mathbf{0.173}$\,/\,$\mathbf{8.58}$ \\
\quad $N=1024$ & $65.90$\,/\,$265.2$ & $0.155$\,/\,$9.88$ & $\underline{0.133}$\,/\,$\underline{9.69}$ & $\mathbf{0.123}$\,/\,$\mathbf{8.46}$ \\
\quad $N=2048$ (OOD) & $97.24$\,/\,$329.8$ & $0.142$\,/\,$9.95$ & $\underline{0.109}$\,/\,$\underline{9.78}$ & $\mathbf{0.080}$\,/\,$\mathbf{8.53}$ \\
\quad $N=4096$ (OOD) & $110.63$\,/\,$363.5$ & $0.120$\,/\,$9.95$ & $\underline{0.090}$\,/\,$\underline{9.79}$ & $\mathbf{0.067}$\,/\,$\mathbf{8.54}$ \\
\addlinespace[3pt]
Water & $0.415$\,/\,$20.83$ & $\underline{0.400}$\,/\,$\underline{14.35}$ & $0.405$\,/\,$14.44$ & $\mathbf{0.305}$\,/\,$\mathbf{11.86}$ \\
\bottomrule
\end{tabular}
\end{adjustbox}
\end{table}

We next consider a synthetic periodic binary ionic melt benchmark with Lennard-Jones interactions and fixed charges of $\pm0.6e$, using Ewald/PPPM~\citep{ewaldBerechnungOptischerUnd1921,hockneyComputerSimulationUsing1988} for reference energies and forces. For this pairwise system, we use analytical NEMORA and compare it directly to MACE and MACE--LES. As shown in Table~\ref{tab:periodic-ions}, both MACE--LES and NEMORA reduce the errors of short-range MACE by two to three orders of magnitude in energy, and by more than an order of magnitude in force. Furthermore, both models extrapolate to larger, unseen system sizes without losing accuracy, with NEMORA remaining slightly more accurate at nearly every size. Additionally, both models recover the $\pm0.6e$ charge within their latent scalar features (Table~\ref{tab:latent-fmm-ionic}). Thus, both successfully learn the true potential energy and forces arising from long-range Coulomb interactions, alongside the short-range interactions, without explicit supervision.

We trained MACE, MACE--LES, and NEMORA on a periodic system of water labeled at the revPBE-D3 level, as described in Appendices~\ref{si:analytical-water} and \ref{si:water}. In this case, dipolar NEMORA ($\ell_Q=1$) has the lowest energy and force errors (Table~\ref{tab:water-nemora}) among the tested models. With scalar sources, NEMORA matches MACE--LES (Table~\ref{tab:latent-fmm-water}), indicating that the accuracy gain comes from the dipoles.

\begin{figure}[hbt!]
    \centering
    \includegraphics[width=\linewidth]{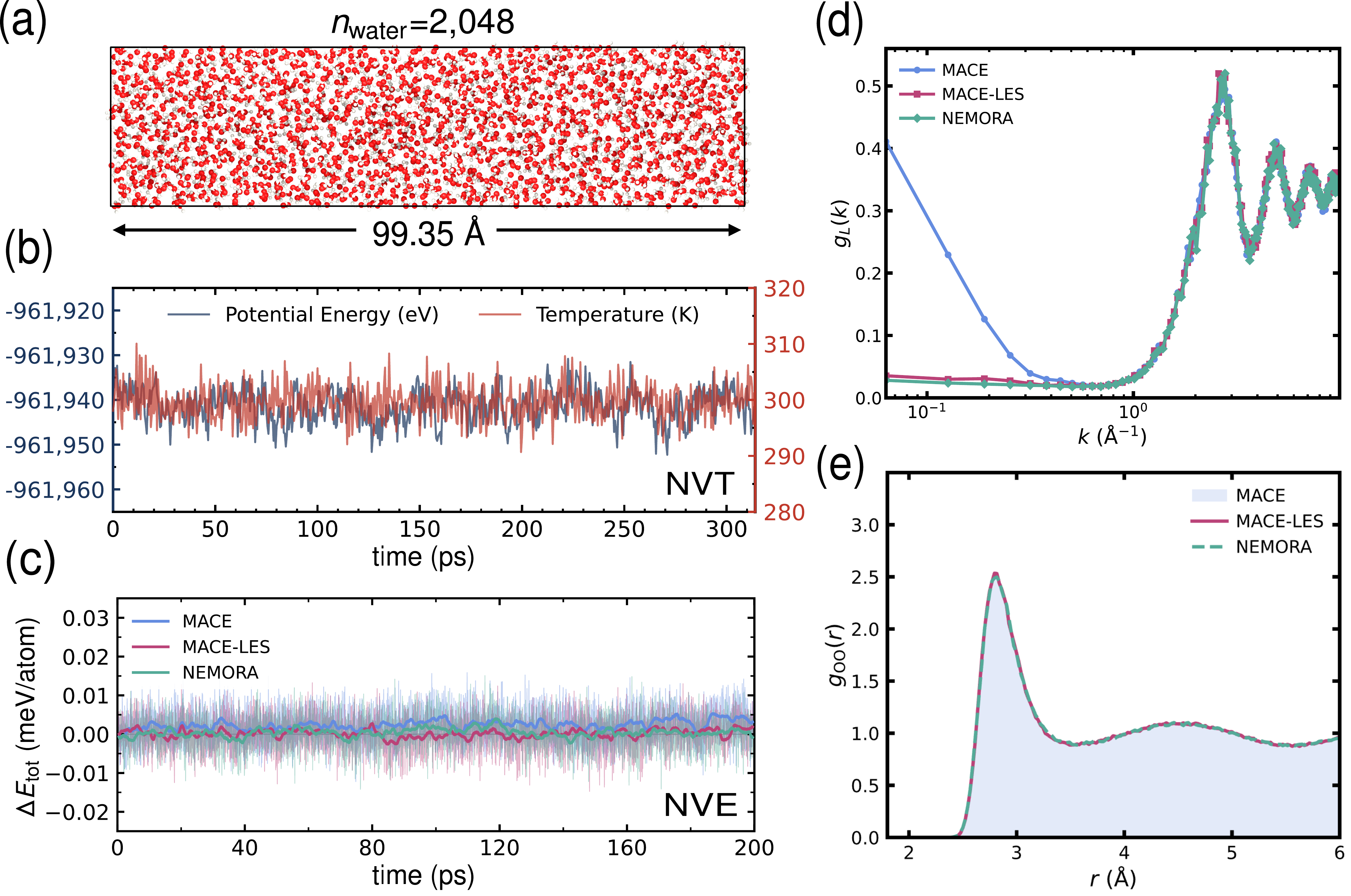}
    \vspace{-0.5cm}
    \caption{Water molecular dynamics. (a) 2048 water molecules in a $24.84\times24.84\times99.35$~\AA\ box at 300~K. (b) NEMORA potential energy and temperature along the NVT trajectory. (c) Total energy drift during NVE simulation of 512 water molecules in a cubic box. (d) Longitudinal dipole correlation $g_L(k)$. (e) O--O radial distribution $g_{\mathrm{OO}}(r)$. Models are those of Table~\ref{tab:periodic-ions}.}
    \label{fig:water}
\end{figure}

To test NEMORA in molecular dynamics, we simulate water in the NVE and NVT ensembles with the scalar model ($\ell_Q=0$). NVE trajectories test the numerical energy conservation of the learned potential. Over 200~ps of velocity-Verlet integration with a 0.5~fs timestep, deuterium masses and a temperature of about 300~K, a cubic box of 512 water molecules has an overall drift of $6\cdot10^{-4}$~meV/atom (Figure~\ref{fig:water}). In the timestep scan from 0.5 to 2~fs, the root-mean-square fluctuation of the total energy grows approximately as $\Delta t^2$, as expected for the velocity-Verlet integrator (Figure~\ref{fig:nve-dt}, Appendix~\ref{si:water-nve}).

Figure~\ref{fig:water}(a,b,d,e) shows NVT molecular dynamics of 2048 water molecules, eight times the largest training system, in an elongated periodic box whose long axis gives access to small wavevectors (Appendix~\ref{si:water-elongated}). Importantly, we find that the potential energy and temperature remain stable along the trajectory (b). Both NEMORA and MACE--LES reproduce the long-wavelength behavior of the polarization density correlation function, $g_L(k)$ (d), whereas short-range MACE rises sharply as $k\rightarrow0$. On the other hand, all three models give nearly identical O--O radial distributions, indicating that standard structural observables cannot distinguish them, and highlighting the need for dedicated long-range benchmarks.

\subsection{Time and memory scaling}
\label{sec:scaling-results}

\begin{figure}[!hbt]
    \centering
    \includegraphics[width=\linewidth]{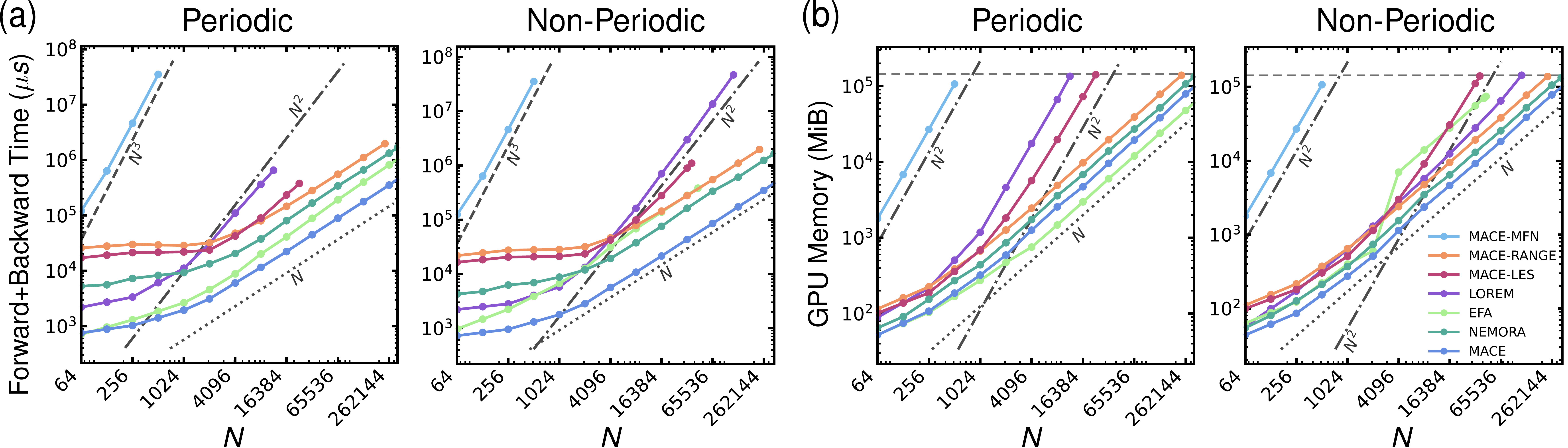}
    \vspace{-0.7cm}
    \caption{(a) Time and (b) peak GPU memory of one energy and force evaluation on fixed-density periodic and non-periodic particle configurations, run from $N=64$ up to about $340{,}000$ atoms on one H200. Curves ending before the upper $N$ limit exhausted the GPU memory. Raw data is summarized in Tables~\ref{tab:scaling-1}--\ref{tab:scaling-4}. All models share the MACE short-range backbone except EFA and LOREM, which use their own short-range architectures.}
    \label{fig:complexity}
\end{figure}

We next test the evaluation time and GPU memory scaling over a broad range of system sizes using randomized configurations of particles in non-periodic clusters and periodic cubic boxes. As shown in Figure~\ref{fig:complexity}, NEMORA scales linearly in both time and memory, in non-periodic and periodic systems alike, and we ran it to about $340{,}000$ atoms on a single H200, where one energy and force evaluation takes about 1.2--1.3~s at $262{,}144$ atoms. The fitted exponents of the long-range component are 0.89--1.00; values slightly below one reflect fixed per-call overhead rather than a sublinear algorithm.

The long-range models closest to NEMORA in accuracy on the benchmarks above do not scale in this way. Dense MFN has quadratic memory scaling and exhausts the memory of the H200 beyond roughly 512 atoms, where computation times are already intractable with cubic scaling. LOREM and LES evaluate the long-range energy with an Ewald sum in periodic systems and a direct pairwise sum in non-periodic ones, and both scale as $\mathcal{O}(N^2)$ in the tested backends; an optimally tuned periodic Ewald sum would scale as $\mathcal{O}(N^{3/2})$ and particle-mesh variants as $\mathcal{O}(N\log N)$, as demonstrated for LOREM by its authors~\citep{rumiantsevLearningLongRangeRepresentations2026}, although that backend is not used here. LOREM exhausts the memory of an H200 at about $11{,}500$ atoms in periodic boxes, and $100{,}000$ atoms in non-periodic clusters.

RANGE and EFA also scale linearly in time and memory (Tables~\ref{tab:scaling-1}--\ref{tab:scaling-4}), but they are less accurate on the long-range benchmarks as demonstrated in previous sections, and RANGE uses more memory than NEMORA at fixed backbone capacity. In large non-periodic systems EFA uses more memory and more time per step than NEMORA, owing to its Lebedev-grid attention; in periodic systems the Lebedev grid is not used and EFA needs less memory. EFA and LOREM use different backbones and feature representations from NEMORA's correlation-order-three MACE backbone, so these costs are not an accuracy-matched comparison. For non-periodic EFA, our fixed 194-point grid and $b_{\max}=4\pi$ imply $\omega_{\max}=4\pi/r_{\max}$, which decreases as the cluster diameter $r_{\max}$ grows; maintaining a fixed physical frequency range and quadrature accuracy would require increasing $b_{\max}$ and the grid size, with additional time and memory costs~\citep{frankMachineLearningGlobal2026}, whereas our NEMORA benchmark grows its hierarchy at fixed expansion order.

\section{Discussion and Conclusion}
NEMORA extends short-range interatomic potentials with an equivariant multipole hierarchy that learns to transmit and combine angular information over long distances. On cumulene, scalar long-range models miss the response to terminal-group rotation, while NEMORA captures the torsional profile and, on unseen chain lengths, is second only to MACE--MFN at linear rather than cubic cost. The backbone and component ablations support this across architectures. On ionic clusters, where scalar charges describe much of the interaction, scalar NEMORA already offers a significant improvement.

Fixing the transport operators gives analytical NEMORA with learned atomic charges and dipoles. The scalar model evaluated on the synthetic ionic melt benchmark and water is comparable to MACE-LES, but higher-degree atomic dipoles improve water predictions. The learned NEMORA operators have linear runtime and memory. NEMORA generally preserves or improves short-range accuracy on MD22, augments a variety of symmetry-constrained and -unconstrained short-range backbones, and supports stable periodic-water dynamics, at twice and eight times the largest training size on NVE and NVT ensembles, respectively. Its hierarchy is equivariant only up to eigenframe degeneracy, and learned operators forgo the convergence guarantees of the analytical FMM. Next steps are broader chemical tests on more complex systems.

\subsection*{Reproducibility, data, and code availability}
The reported experiments use both PyTorch and JAX implementations. Benchmark run counts and exclusions are recorded in Appendix~\ref{si:additional-results}; differences in training schedules and the backbone, optimizer and NEMORA settings for each benchmark are given in Tables~\ref{tab:sr-mace-params} and~\ref{tab:optimizer-settings} and Appendix~\ref{si:system-settings}. Additional experiment protocols accompany their results. Across-run uncertainties are SEMs unless explicitly identified as block estimates for molecular dynamics. Single-run measurements and selected-seed profiles are identified separately.

Datasets originate from the cited primary sources, except the synthetic ionic datasets and periodic water dataset generated in this work. Code and both datasets will be released upon publication at {\hypersetup{urlcolor=blue}\url{https://github.com/JayLKap/nemora}}. Trained NEMORA models will be made publicly available in a future release. Training and simulation scripts will be made available by the corresponding author upon reasonable request.

Generative AI tools assisted with code implementation, experiment execution, figure preparation, literature review, and manuscript editing under author supervision. The authors reviewed the AI-assisted work and take full responsibility for the final content.

\IfFileExists{arxiv_references.bib}{%
  \bibliography{arxiv_references}
  \bibliographystyle{arxiv}
}{%
  \bibliography{arXiv/arxiv_references}
  \bibliographystyle{arXiv/arxiv}
}


\appendix

\clearpage
\section{Scalar FMM}
\label{si:proofs}
\label{si:scalar-fmm}

\subsection{Kernel and expansion equivariance}
\label{si:coulomb-equivariance}
Let $R\in\mathrm{SO}(3)$ be a rotation and $\bm r_i'=R\bm r_i+\bm b$. For $s\ne t$,
\begin{equation}
\label{eq:coulomb-invariance}
G(\bm r_t',\bm r_s')
=\frac{1}{\|R(\bm r_t-\bm r_s)\|}
=G(\bm r_t,\bm r_s),\qquad G(\bm r_t,\bm r_s)=\|\bm r_t-\bm r_s\|^{-1}.
\end{equation}
Thus $V'(\bm r_t')=V(\bm r_t)$, and reindexing the sources leaves the sum unchanged. The harmonic and radial factors obey
\[
\bm Y^{(\ell)}(R\hat{\bm r})=D^{(\ell)}(R)\bm Y^{(\ell)}(\hat{\bm r}),
\qquad R_\tau^{(\ell)}(\|R\bm r\|)=R_\tau^{(\ell)}(r).
\]
In particular, with covariant centers, the scalar-source multipoles transform as
\[
\bm\Phi_s'{}^{(\ell)}
=\sum_{i\in s}q_i\tilde{\bm R}^{(\ell)}\!\left(R(\bm r_i-\bm c_s)\right)
=D^{(\ell)}(R)\bm\Phi_s^{(\ell)}.
\]
Replacing $\tilde{\bm R}^{(\ell)}$ by $\tilde{\bm I}^{(\ell)}$ gives the same law for local coefficients. The CG identity~\citep{geigerE3nnEuclideanNeural2022} then gives stage covariance and complete-block truncation invariance~\citep{greengard_fast_1987,cheng_fast_1999}:
\begin{equation}
\label{eq:fmm-stage-equivariance}
T^\tau(R\bm r)D_s^\tau(R)=D_t^\tau(R)T^\tau(\bm r),
\qquad P_pD(R)=D(R)P_p.
\end{equation}
Here $D_s^\tau,D_t^\tau$ act on $\ell_s^\tau,\ell_t^\tau$, and $P_p$ retains complete spatial degrees $0,\ldots,p$. Composition through L2P gives $V_p'(\bm r_t')=V_p(\bm r_t)$, including the direct near field. Since $\bm Y^{(\ell)}(-\hat{\bm r})=(-1)^\ell\bm Y^{(\ell)}(\hat{\bm r})$ and every analytical path obeys $(-1)^L=(-1)^{j+\ell}$, the proof extends to $R\in O(3)$. Throughout, centers, memberships, interaction lists and tie handling must transform consistently with geometry and particle reindexing. Under this condition, relative displacements ensure translation invariance and source sums ensure permutation equivariance.

\subsection{Stage operators and phase conventions}
\label{si:scalar-stage-weights}
We use orthonormal complex harmonics with the Condon--Shortley phase and Racah-normalized solid harmonics~\citep{greengard_fast_1987,cheng_fast_1999}:
\begin{equation}
\label{eq:fmm-expansions}
\tilde{\bm R}^{(\ell)}(\bm x)=\sqrt{\frac{4\pi}{2\ell+1}}\,\|\bm x\|^{\ell}\,\bm Y^{(\ell)}(\hat{\bm x}),
\qquad
\tilde{\bm I}^{(\ell)}(\bm x)=\sqrt{\frac{4\pi}{2\ell+1}}\,\|\bm x\|^{-(\ell+1)}\,\bm Y^{(\ell)}(\hat{\bm x}).
\end{equation}
The source moments and target local coefficients of Section~\ref{sec:fmm} satisfy
\[
\begin{aligned}
\bm\Phi_s^{(\ell)}&=\sum_{i\in s}q_i\tilde{\bm R}^{(\ell)}(\bm r_i-\bm c_s),
&\bm\Phi_t^{(\ell)}&=\sum_{i\in\mathrm{far}(t)}q_i\tilde{\bm I}^{(\ell)}(\bm r_i-\bm c_t),\\
V_{\mathrm{far}}(\bm c_t+\bm x)&\approx\frac{1}{4\pi\varepsilon_0}\sum_{\ell=0}^p\bm\Phi_t^{(\ell)}\cdot\tilde{\bm R}^{(\ell)}(\bm x),
&\bm a\cdot\bm b&=\sum_m a_mb_m^\star.
\end{aligned}
\]
Here $\mathrm{far}(t)$ contains the transferred source particles. At the origin, $\tilde{\bm R}^{(0)}=1$ and $\tilde{\bm R}^{(\ell>0)}=\bm0$. Transforming harmonics, coefficients and CG tensors consistently gives the real basis of the main text, where conjugation drops.

The addition theorems in this normalization are
\begin{equation}
\label{eq:fmm-component-kernel}
\begin{aligned}
\tilde{\bm R}^{(\ell)}(\bm a+\bm b)
&=\sum_{n=0}^{\ell}\binom{2\ell}{2n}^{1/2}
\bigl[\tilde{\bm R}^{(\ell-n)}(\bm a)\otimes\tilde{\bm R}^{(n)}(\bm b)\bigr]^{(\ell)},\\
\tilde{\bm I}^{(\ell)}(\bm a-\bm b)
&=\sum_{\ell'=0}^{\infty}(-1)^{\ell+\ell'}\left[\frac{2(\ell+\ell')+1}{2\ell+1}\binom{2(\ell+\ell')}{2\ell'}\right]^{1/2}
\bigl[\tilde{\bm R}^{(\ell')}(\bm a)\otimes\tilde{\bm I}^{(\ell+\ell')}(\bm b)\bigr]^{(\ell)}.
\end{aligned}
\end{equation}
The regular identity is exact; the irregular identity requires $\|\bm a\|<\|\bm b\|$. Setting $\bm a=\bm r_i-\bm c_s$ and $\bm b=\bm c_t-\bm c_s=-\bm r$ gives the Laplace expansion at $\ell=0$ and M2L at $\ell>0$, truncated to input/output orders $\le p$. Regular shifts give M2M and, by transposition, L2L; L2P evaluates the local field. Enclosures obeying $(R_s+R_t)/\|\bm c_t-\bm c_s\|<1$ ensure convergence throughout the target region.

For input/output degrees $j\in\ell_s^\tau$, $L\in\ell_t^\tau$ and channels $b,a$, the weighted CG product is
\begin{equation}
\label{eq:weighted-stage-product}
\Bigl[\bm x\otimes_w\bigoplus_\ell\bm f^{(\ell)}\Bigr]_{aM}^{(L)}
=\sum_{b,j,\ell}\sum_{m_j,m}w_{aL,bj,\ell}\,C^{LM}_{jm_j,\ell m}\,x_{b m_j}^{(j)}f_m^{(\ell)}.
\end{equation}
The path sum also includes filter channels when present. Scalar-source FMM has one channel per spatial degree. With $a_\ell=\sqrt{4\pi/(2\ell+1)}$, the addition theorems give
\begin{equation}
\label{eq:fixed-stage-weights}
\frac{c^\tau_{Lj\ell}}{a_\ell}=
\begin{cases}
1, & \tau=\mathrm{P2M},\quad j=0,\ \ell=L,\\
\sqrt{\binom{2L}{2j}}, & \tau=\mathrm{M2M},\quad L=j+\ell,\\
\sqrt{\dfrac{2\ell+1}{2L+1}\binom{2\ell}{2j}}, & \tau=\mathrm{M2L},\quad \ell=j+L,\\
\sqrt{\dfrac{2j+1}{2L+1}\binom{2j}{2L}}, & \tau=\mathrm{L2L},\quad j=L+\ell,\\
\sqrt{2j+1}, & \tau=\mathrm{L2P},\quad L=0,\ \ell=j.
\end{cases}
\end{equation}
Other paths have zero weight. These coefficients assume Racah-normalized states, orthonormal harmonic filters and source-minus-target displacements, including center minus particle at L2P; the direction and CG coefficients carry the signs.

\clearpage
\section{Tensor FMM}
\label{si:tensor-fmm}
\subsection{Kernel and expansion equivariance}
\label{si:tensor-equivariance}
Let $\bm Q_s'{}^{(\ell_Q)}=D^{(\ell_Q)}(R)\bm Q_s^{(\ell_Q)}$ and $\bm r_s'=R\bm r_s+\bm b$. Invariance of pair distances gives componentwise Coulomb equivariance directly~\citep{rumiantsevLearningLongRangeRepresentations2026}:
\begin{equation}
\label{eq:equiv-coulomb}
\bm V^{(\ell_Q)}(\bm r_t)
=\frac{1}{4\pi\varepsilon_0}\sum_{s\ne t}
\frac{\bm Q_s^{(\ell_Q)}}{\|\bm r_t-\bm r_s\|},
\qquad \bm V'{}^{(\ell_Q)}(\bm r_t')=D^{(\ell_Q)}(R)\bm V^{(\ell_Q)}(\bm r_t).
\end{equation}
To prove the same result for the expansions, first keep the charge and spatial indices separate. At a center $\bm c_u$, write
\begin{equation}
\label{eq:tensor-expansion-equivariance}
\begin{aligned}
\widehat{\bm\Phi}_u^{(n)}
&=\sum_{i\in\mathcal S_u}\bm Q_i^{(\ell_Q)}\otimes\bm H^{(n)}(\bm r_i-\bm c_u),\\
\widehat{\bm\Phi}_u'{}^{(n)}
&=\bigl[D^{(\ell_Q)}(R)\otimes D^{(n)}(R)\bigr]\widehat{\bm\Phi}_u^{(n)}.
\end{aligned}
\end{equation}
Here $\bm H^{(n)}=\tilde{\bm R}^{(n)}$ for multipoles and $\tilde{\bm I}^{(n)}$ for local coefficients; $\mathcal S_u$ contains the corresponding source particles. The second line follows by transforming each factor and using $\bm c_u'=R\bm c_u+\bm b$. In this equation $\otimes$ is the ordinary tensor product, before CG coupling. For the CG map $C_{\ell_Q n}^{L}$~\citep{geigerE3nnEuclideanNeural2022},
\begin{equation}
\label{eq:tensor-coupled-equivariance}
\bm\Phi_{u,n}^{(L)}=C_{\ell_Q n}^{L}\widehat{\bm\Phi}_u^{(n)},
\qquad
\bm\Phi_{u,n}'{}^{(L)}
=C_{\ell_Q n}^{L}(D^{(\ell_Q)}\otimes D^{(n)})\widehat{\bm\Phi}_u^{(n)}
=D^{(L)}\bm\Phi_{u,n}^{(L)},
\end{equation}
where all $D$ matrices are evaluated at $R$ and $|\ell_Q-n|\le L\le\ell_Q+n$. Thus every coupled expansion block is equivariant, with spatial order $n$ retained as a multiplicity label.

For a scalar stage $T^\tau$ from Appendix~\ref{si:scalar-stage-weights}, the uncoupled tensor stage is $\widehat T^\tau=I_{\ell_Q}\otimes T^\tau$, with $I_{\ell_Q}$ the identity on charge components. Equation~\ref{eq:fmm-stage-equivariance} gives
\begin{equation}
\label{eq:tensor-stage-equivariance}
\widehat T^\tau(R\bm r)\bigl[D^{(\ell_Q)}(R)\otimes D_s^\tau(R)\bigr]
=\bigl[D^{(\ell_Q)}(R)\otimes D_t^\tau(R)\bigr]\widehat T^\tau(\bm r).
\end{equation}
Unitary CG changes of basis preserve this identity. So does truncation by $I_{\ell_Q}\otimes P_p$, provided every coupling of $\ell_Q$ with $n=0,\ldots,p$ and its multiplicity is retained. L2P contracts the spatial part to degree zero, leaving the charge degree $\ell_Q$:
\[
\bm V_p'{}^{(\ell_Q)}(\bm r_t')=D^{(\ell_Q)}(R)\bm V_p^{(\ell_Q)}(\bm r_t).
\]
For charge parity $\eta_Q$, a spatial-order-$n$ block has parity $\eta_Q(-1)^n$; preserving these parities extends the proof to $O(3)$. The hierarchy conditions of Appendix~\ref{si:coulomb-equivariance} give translation and permutation equivariance. This is tensor transport through a scalar kernel, distinct from physical multipole interactions, which require kernel derivatives~\citep{stoneTheory2013}.

\subsection{Stage operators and phase conventions}
\label{si:equivariant-fmm}
The coupled input and output degrees satisfy $|\ell_Q-n|\le j\le\ell_Q+n$ and $|\ell_Q-n'|\le L\le\ell_Q+n'$. In the convention of Appendix~\ref{si:scalar-stage-weights}, changing coupling order from $((\ell_Q\otimes n)_j\otimes\ell)_L$ to $(\ell_Q\otimes(n\otimes\ell)_{n'})_L$ gives
\begin{equation}
\label{eq:tensor-stage-weights}
c^\tau_{(n',L),(n,j),\ell}
=c^\tau_{n'n\ell}\,(-1)^{\ell_Q+n+\ell+L}\sqrt{(2j+1)(2n'+1)}
\begin{Bmatrix}\ell_Q&n&j\\\ell&L&n'\end{Bmatrix}.
\end{equation}
The braces denote a Wigner $6j$ symbol and $c^\tau_{n'n\ell}$ is the scalar weight in Equation~\ref{eq:fixed-stage-weights}. Spatial orders $n,n'$ remain multiplicity labels; $\ell_Q=0$ recovers scalar FMM.

The paired Coulomb kernel follows from two CG identities:
\begin{equation}
\label{eq:fmm-pairing}
\sum_{m=-\ell}^{\ell}Y^{\star}_{\ell m}(\hat{\bm a})\,Y_{\ell m}(\hat{\bm b})=(-1)^{\ell}\sqrt{2\ell+1}\,\bigl[\bm Y^{(\ell)}(\hat{\bm a})\otimes\bm Y^{(\ell)}(\hat{\bm b})\bigr]^{(0)},
\end{equation}
and, with $c_{\ell_1}=(-1)^{\ell_Q+\ell_1}\sqrt{(2\ell_1+1)/(2\ell_Q+1)}$,
\begin{equation}
\label{eq:fmm-re-coupling}
\sum_{\ell_1}c_{\ell_1}\Bigl[\bigl[\bm Q^{(\ell_Q)}\otimes\bm Y^{(\ell)}(\hat{\bm a})\bigr]^{(\ell_1)}\otimes\bm Y^{(\ell)}(\hat{\bm b})\Bigr]^{(\ell_Q)}=\bm Q^{(\ell_Q)}\sum_{m=-\ell}^{\ell}Y^{\star}_{\ell m}(\hat{\bm a})\,Y_{\ell m}(\hat{\bm b}).
\end{equation}
These follow from the scalar CG coefficient and $6j$ recoupling of $\ell_Q,\ell,\ell$. The pairing weights $c_{\ell_1}$ are distinct from the stage weights $c^\tau$. As for scalar FMM, a change to real harmonics must transform the CG tensors and coefficient vectors consistently.

\section{Learnable NEMORA: Proofs}
\label{si:nemora-proofs}
\subsection{Learned operator equivariance}
\label{si:learned-equivariance}
For $|\ell_1-\ell_2|\le\ell\le\ell_1+\ell_2$, the CG product of Equation~\ref{eq:cg-product} satisfies~\citep{geigerE3nnEuclideanNeural2022}
\begin{equation}
\label{eq:cg-intertwiner-proof}
\bigl[D^{(\ell_1)}(R)\bm f^{(\ell_1)}\otimes D^{(\ell_2)}(R)\bm g^{(\ell_2)}\bigr]^{(\ell)}
=D^{(\ell)}(R)\bigl[\bm f^{(\ell_1)}\otimes\bm g^{(\ell_2)}\bigr]^{(\ell)}.
\end{equation}
For $O(3)$, output parity equals the product of input parities. In Equation~\ref{eq:nemora-translation}, invariant radial weights and channel mixing within the same degree and parity give $\bm T_{st}'{}^{(\ell_M)}=D^{(\ell_M)}(R)\bm T_{st}^{(\ell_M)}$ for each ungated block, $\ell_M\in\tilde\ell_t^\tau$. Summation over a covariantly selected interaction list preserves this identity. The hierarchy conditions are those of Appendix~\ref{si:coulomb-equivariance}. Analytical recovery sets $\alpha_\theta=1$, $\bm R_\theta^{(\ell)}=\bm1$, $\bm W^V=\bm I$ and the weights of Appendix~\ref{si:equivariant-fmm}, retaining all required blocks and M2L filters through $2p$.

\subsection{Invariant gates}
\label{si:learnable-multipole-gate}
The gated sum is
\begin{equation}
\label{eq:gated-equivariant-translation}
\bm\Psi_t^{(\ell_M)}=\sum_{s\in\mathcal N_\tau(t)}\alpha_{\theta,st}^{(\ell_M)}\bm T_{st}^{(\ell_M)},
\qquad
\alpha_{\theta,st}^{(\ell_M)}=\bm\phi_\theta(\mathcal G_{st}\oplus\mathcal M_{st}\oplus\mathcal A_{st})^{(\ell_M)}.
\end{equation}
The inputs are geometric invariants, block norms, and same-irrep, same-parity alignments:
\begin{equation}
\label{eq:alignment-features}
\mathcal A_{st}=\bigoplus_\ell\langle\bm q_t^{(\ell)},\bm k_s^{(\ell)}\rangle,
\qquad \bm q_t^{(\ell)}=\bm W_Q\bm\Psi_t^{(\ell)},
\qquad \bm k_s^{(\ell)}=\bm W_K\bm\Psi_s^{(\ell)}.
\end{equation}
Orthogonality of $D^{(\ell)}(R)$ gives $\langle D\bm q,D\bm k\rangle=\langle\bm q,\bm k\rangle$, hence $\alpha'_{\theta,st}=\alpha_{\theta,st}$ and
\[
\bm\Psi_t'{}^{(\ell_M)}
=\sum_s\alpha_{\theta,st}^{(\ell_M)}D^{(\ell_M)}(R)\bm T_{st}^{(\ell_M)}
=D^{(\ell_M)}(R)\bm\Psi_t^{(\ell_M)}.
\]
Gates are analogous to damping and polarization factors~\citep{tangImprovedSimpleModel1984,ponderCurrentStatusAMOEBA2010,stoneTheory2013}, not self-consistent polarization equations.

\subsection{Invariant readout and equivariant forces}
\label{si:readout-proof}
Equivariant backbone, lift, residuals, contractions, Kelvin maps and periodic correction, followed by even-scalar atomic readouts, imply
\[
E(\{R\bm r_i+\bm b\},R\bm H)=E(\{\bm r_i\},\bm H),
\qquad \bm F_i'= -R\nabla_{\bm r_i}E=R\bm F_i.
\]
The force identity holds wherever $E$ is differentiable; omit $\bm H$ for nonperiodic systems. Atomic summation preserves permutation invariance.

\section{Symmetric Contraction: Proofs}
\label{si:symcontract-proof}
\subsection{Kernel factorization and distributivity}
On an admissible domain, assume a finite factorization and separable pair expansions:
\[
K^{(\nu+1)}(\bm r_t,\bm r_{s_1},\ldots,\bm r_{s_\nu})=\sum_a\prod_{k=1}^\nu K_{ak}(\bm r_t,\bm r_{s_k}),
\qquad K_{ak}(\bm r_t,\bm r_s)=\sum_\beta u_\beta(\bm r_t)v_{ak,\beta}(\bm r_s).
\]
Here $a$ indexes factorization rank and $\beta$ basis/channel components. Define accumulated coefficients $A_{ak,\beta}=\sum_s v_{ak,\beta}(\bm r_s)$ and fields $F_{ak}(\bm r_t)=\sum_\beta u_\beta(\bm r_t)A_{ak,\beta}$. For FMM, $u_\beta$ is a regular solid-harmonic component about the target center and $A_{ak,\beta}$ its local coefficient. Distributivity gives
\begin{equation}
\label{eq:factorized-expansion-product}
\begin{aligned}
U(\bm r_t)
&=\sum_{s_1,\ldots,s_\nu}K^{(\nu+1)}(\bm r_t,\bm r_{s_1},\ldots,\bm r_{s_\nu})
=\sum_a\prod_{k=1}^{\nu}F_{ak}(\bm r_t)\\
&=\sum_{\beta_1,\ldots,\beta_\nu}
\underbrace{\left[\sum_a\prod_{k=1}^{\nu}A_{ak,\beta_k}\right]}_{B_{\beta_1\cdots\beta_\nu}}
\prod_{k=1}^{\nu}u_{\beta_k}(\bm r_t).
\end{aligned}
\end{equation}
The basis products commute, so $U_B=U_{\mathcal S_\nu B}$, where $\mathcal S_\nu$ symmetrizes the $\beta$ indices. Distinct-source terms have body order $\nu+1$ for one-source pair factors; repeated indices contribute lower orders. The identity is exact for the represented factorization. Exact reconstruction requires the product basis, including radial factors, and product translation maps: harmonic fields are not closed under multiplication. Equation~\ref{eq:symcontract} instead projects these products into configured equivariant feature blocks.

\subsection{Equivariance}
\label{si:symcontract-equivariance}
For the coefficient-block representation $D(R)$ and a parity-compatible generalized CG map $\mathcal C_{\ell_M}$,
\[
\mathcal S_\nu D(R)^{\otimes\nu}=D(R)^{\otimes\nu}\mathcal S_\nu,
\qquad
\mathcal C_{\ell_M}\mathcal S_\nu D(R)^{\otimes\nu}
=D^{(\ell_M)}(R)\mathcal C_{\ell_M}\mathcal S_\nu.
\]
Thus symmetric contractions act equivariantly on expansion coefficients before evaluation~\citep{drautz_atomic_2019,dusson_atomic_2022,batatiaMACEHigherOrder2023}.

\section{Global Kelvin: Proofs}
\label{si:kelvin-global}
\subsection{Local-field construction and summation}
For a support of radius $R_u$ centered at $\bm c_u$, set $\bm x=\bm r_t-\bm c_u$, $r=\|\bm x\|$. Kelvin inversion~\citep{amaral_kelvin_2017} maps irregular to regular harmonics:
\[
(\mathcal K_{R_u}f)(\bm x)=\frac{R_u}{r}f\!\left(\frac{R_u^2\bm x}{r^2}\right),
\]
\begin{equation}
\label{eq:kelvin-irregular-to-regular}
\mathcal K_{R_u}\tilde{\bm I}^{(\ell)}(\bm x)
=\frac{R_u}{r}\left(\frac{R_u^2}{r}\right)^{-(\ell+1)}
\sqrt{\frac{4\pi}{2\ell+1}}\,\bm Y^{(\ell)}(\hat{\bm x})
=R_u^{-(2\ell+1)}\tilde{\bm R}^{(\ell)}(\bm x).
\end{equation}
Write the node's multipole and local states as $\bm\Phi_{M,u},\bm\Phi_{L,u}$. For $f_M^{(p)}=\sum_{\ell=0}^p\bm\Phi_{M,u}^{(\ell)}\cdot\tilde{\bm I}^{(\ell)}$,
\begin{equation}
\label{eq:kelvin-local-coefficients}
(\mathcal K_{R_u}f_M^{(p)})(\bm x)
=\sum_{\ell=0}^{p}\Delta\bm\Phi_{L,u}^{(\ell)}\cdot\tilde{\bm R}^{(\ell)}(\bm x),
\qquad \Delta\bm\Phi_{L,u}^{(\ell)}=R_u^{-(2\ell+1)}\bm\Phi_{M,u}^{(\ell)}.
\end{equation}
If the original sources lie within radius $a<R_u$, the full inverted series converges for $r<R_u^2/a$ and is regular at the center. For multiple supports $k$, field addition means evaluation at a common target, not coefficientwise addition across different centers:
\[
f_{\mathrm G}(\bm r_t)=\sum_k\sum_{\ell=0}^{p}
\Delta\bm\Phi_{L,k}^{(\ell)}\cdot\tilde{\bm R}^{(\ell)}(\bm r_t-\bm c_k).
\]
Each term must be evaluated within its convergence domain. The learned counterpart pools support states by Equation~\ref{eq:nemora-kelvin} and broadcasts them to particles:
\[
\bm a_t^{\mathrm G}=\sum_k\mathcal T^{\mathrm{L2P}}_{t\leftarrow k}(\bm\Psi_k),
\qquad \bm a_t^{\mathrm{total}}=\bm a_t^{\mathrm{hier}}+\bm a_t^{\mathrm G}.
\]
The two outputs share target irreps and channel layout. This defines an additive learned response; it does not require the feature-dependent maps to be linear or the supports to solve a common dielectric boundary problem.

\subsection{Equivariance}
\label{si:kelvin-equivariance}
For an invariant support radius and $R\in O(3)$,
\[
\frac{R_u^2R\bm x}{\|R\bm x\|^2}=R\frac{R_u^2\bm x}{r^2},
\qquad \mathcal K_{R_u}(f\circ R^{-1})=(\mathcal K_{R_u}f)\circ R^{-1}.
\]
For tensor-valued fields the component action $D(R)$ also commutes with $\mathcal K_{R_u}$. In the learned map, invariant radii, distances and gates and same-irrep channel mixing give
\[
\bm\Psi_k'{}^{(\ell)}=D^{(\ell)}(R)\bm\Psi_k^{(\ell)},
\qquad
\bm a_t'{}^{\mathrm G,(\ell)}
=\sum_k D^{(\ell)}(R)\mathcal T^{\mathrm{L2P}}_{t\leftarrow k}(\bm\Psi_k)^{(\ell)}
=D^{(\ell)}(R)\bm a_t^{\mathrm G,(\ell)}.
\]
Here support centers transform with the geometry, and support selection and pooling respect particle reindexing. The L2P identity follows from Appendix~\ref{si:learned-equivariance}.

\section{Cluster-Local Kelvin: Proofs}
\label{si:kelvin-local}
\subsection{Local-field addition}
At hierarchy node $u$, Equation~\ref{eq:kelvin-local-coefficients} supplies a regular field about the same center as the downward expansion. In a common basis, normalization and validity domain,
\[
\bm\Phi_{L,u}^{\mathrm{total}}=\bm\Phi_{L,u}+\Delta\bm\Phi_{L,u},
\qquad T(\bm\Phi_{L,u}+\Delta\bm\Phi_{L,u})=T\bm\Phi_{L,u}+T\Delta\bm\Phi_{L,u},
\quad T\in\{\mathrm{L2L,L2P}\}.
\]
With compatible normalized channels, the unmodulated learned self-update is
\begin{equation}
\tilde{\bm\Psi}_{L,u}^{(\ell)}
=\bm\Psi_{L,u}^{(\ell)}+\tilde R_u^{-(2\ell+1)}\bm\Psi_{M,u}^{(\ell)},
\qquad \tilde R_u=R_u/r_{\mathrm{cut}},
\end{equation}
For analytical tensor fields the Kelvin factor is $R_u^{-(2n+1)}$, indexed by spatial order $n$, not total coupled degree. The learned degree-diagonal map is a local-state correction, not necessarily an exact tensor-field inversion or dielectric solution.

\subsection{Equivariance}
\label{si:kelvin-local-equivariance}
Let $\Delta\bm\Psi_{L,u}^{(\ell)}$ denote the learned correction. Invariant radii and gates and same-irrep channel mixing imply
\[
\Delta\bm\Psi_{L,u}'{}^{(\ell)}=D^{(\ell)}(R)\Delta\bm\Psi_{L,u}^{(\ell)},
\qquad
\tilde{\bm\Psi}_{L,u}'{}^{(\ell)}
=D^{(\ell)}(R)\bigl(\bm\Psi_{L,u}^{(\ell)}+\Delta\bm\Psi_{L,u}^{(\ell)}\bigr).
\]
Subsequent equivariant L2L and L2P maps preserve this identity, including when feature-dependent gates make them nonlinear.

\section{Periodic Correction: Proof}
\subsection{Equivariance}
\label{si:pbc-equivariance}
With $\bm H'=R\bm H$ and consistently transformed image selection,
\[
\|R\bm H\bm n\|=\|\bm H\bm n\|,\qquad |\det(R\bm H)|=|\det\bm H|.
\]
The degree-diagonal correction in Equation~\ref{eq:pbc-tinfoil} therefore commutes with $D^{(\ell)}(R)$. This establishes equivariance, not exact periodic electrostatics.

\section{Linear-Time Complexity}
\label{si:linear-time}
Let $n_{\mathrm{node}}+n_{\mathrm{edge}}=\mathcal O(N)$, leaf occupancy and interaction-list sizes be bounded, and angular degrees, channels, radial basis size and contraction order be fixed~\citep{cheng_fast_1999}. Each node/edge operation then costs $\mathcal O(1)$, and bounded near lists give $n_{\mathrm{pair}}=\mathcal O(N)$. With $K$ Kelvin anchors,
\[
T_{\mathrm{eval}}=\mathcal O(n_{\mathrm{node}}+n_{\mathrm{edge}}+n_{\mathrm{pair}}+NK+|\mathcal N_{\mathrm{lat}}|)
=\mathcal O(N),\qquad M_{\mathrm{eval}}=\mathcal O(N),
\]
for fixed $K$ and image shell. The bound includes contractions and root reinjection, but not hierarchy construction; a separation test alone does not guarantee bounded interaction lists.

\clearpage
\section{Analytical Kernel Families}
\label{si:greens}
For $r_< < r_>$, the Coulomb, Yukawa and outgoing Helmholtz kernels admit the separated expansion~\citep{cheng_fast_1999,winkelmannModifiedHelmholtz2021,gumerovHelmholtz2005}
\[
G(\|\bm r_>-\bm r_<\|)
=\sum_{\ell=0}^{\infty} A_\ell u_\ell(r_<)v_\ell(r_>)
\sum_{m=-\ell}^{\ell}Y_{\ell m}^{\star}(\hat{\bm r}_<)Y_{\ell m}(\hat{\bm r}_>).
\]
Table~\ref{tab:greens_tier1} lists these bases and further analytical FMM families. For Yukawa, $\kappa>0$ and
$i_\ell(z)=\sqrt{\pi/(2z)}I_{\ell+1/2}(z)$,
$k_\ell(z)=\sqrt{2/(\pi z)}K_{\ell+1/2}(z)$, so $k_0(z)=e^{-z}/z$.
The $\kappa$ rescalings are a basis convention; their product supplies the required prefactor $\kappa$. For Helmholtz, $k>0$, $j_\ell$ and $h_\ell^{(1)}$ are spherical Bessel and outgoing Hankel functions.

\begin{table}[H]
\centering
\small
\caption{\textbf{Analytical free-space kernel families in three dimensions.} The first three rows use the separated form above; the remaining families require the indicated additional structure. Overall physical prefactors are omitted, and tensor kernels are stated away from their singularities.}
\label{tab:greens_tier1}
\begin{tabular}{p{2.4cm}p{3.8cm}p{8.0cm}}
\toprule
\textbf{Family} & \textbf{Kernel} & \textbf{Analytical expansion structure} \\
\midrule
Coulomb & $1/r$ &
$u_\ell=r^\ell$, $v_\ell=r^{-(\ell+1)}$, $A_\ell=4\pi/(2\ell+1)$. \\[6pt]
Yukawa & $e^{-\kappa r}/r$ &
$u_\ell=\kappa^{-\ell}i_\ell(\kappa r)$, $v_\ell=\kappa^{\ell+1}k_\ell(\kappa r)$, $A_\ell=4\pi$. \\[6pt]
Helmholtz & $e^{ikr}/r$ &
$u_\ell=j_\ell(kr)$, $v_\ell=h_\ell^{(1)}(kr)$, $A_\ell=4\pi ik$~\citep{gumerovHelmholtz2005}. \\[6pt]
Biharmonic & $r$ &
Regular branches $r^\ell,r^{\ell+2}$ and irregular branches $r^{-\ell-1},r^{1-\ell}$, multiplied by $Y_{\ell m}$; coupled branch translations~\citep{gumerovBiharmonic2006}. \\[6pt]
Inverse power & $r^{-\lambda}$, $\lambda>0$ &
Gegenbauer expansion with radial pairs $r_<^n r_>^{-(n+\lambda)}$; angular factors $C_n^{(\lambda/2)}(\cos\gamma)$. Includes dispersion powers $\lambda=6,8$~\citep{chowdhuryMultipole2005}. \\[6pt]
Stokes & $\displaystyle\frac{\bm I}{r}+\frac{\bm r\bm r^\top}{r^3}$ &
Tensor expansion obtained from harmonic and biharmonic potentials and their derivatives; not a scalar radial pair~\citep{gumerovBiharmonic2006,greengardLayerPotentials2021}. \\[6pt]
Maxwell & $\displaystyle\left(\bm I+\frac{\nabla\nabla^\top}{k^2}\right)\frac{e^{ikr}}{r}$ &
Dyadic expansion obtained by differentiating Helmholtz expansions, or using vector spherical waves~\citep{gumerovHelmholtz2005,greengardLayerPotentials2021}. \\
\bottomrule
\end{tabular}
\end{table}
Here $\cos\gamma=\hat{\bm r}_<\cdot\hat{\bm r}_>$ and the inverse-power expansion is
\[
\|\bm r_>-\bm r_<\|^{-\lambda}
=\sum_{n=0}^{\infty}\frac{r_<^n}{r_>^{n+\lambda}}C_n^{(\lambda/2)}(\cos\gamma).
\]
Each family requires its own addition coefficients and, where applicable, radial branches or tensor structure; replacing Coulomb radial factors alone is insufficient. These are unsmoothed free-space expansions, not claims of implemented NEMORA variants. Only the Coulomb-based model is benchmarked here.

\clearpage
\section{Long-Range Models}
\begin{table}[H]
\centering
\scriptsize
\caption{Long-range neural methods along six architectural axes.
\emph{Equiv.}: $\ell > 0$ tensorial long-range messages.
\emph{Learn.}: learnable propagation parameters, excluding source construction and readout.
\emph{Many-body}: pairwise (\Circle), single-mode many-body via aggregation/path (\LEFTcircle), or both atom-level and coarse-level long-range coupling (\CIRCLE).
\emph{$\ell$-mix}: angular channel mixing during long-range transport.
\emph{Hier.}: hierarchical multi-scale structure.
\emph{Scaling}: cost of the long-range component at fixed feature widths and block count; linear costs require bounded interaction lists, leaf occupancy or graph degree. Symbols describe the long-range mechanism, not the total body order of the complete potential.
\Circle\ = no, \LEFTcircle\ = partial/approximate, \CIRCLE\ = yes.}
\label{tab:methods}
\begin{tabular}{p{7.5cm}cccccc}
\toprule
Method & Equiv. & Learn. & MB & $\ell$-mix & Hier. & Scaling \\
\midrule
\multicolumn{7}{l}{\emph{Scalar long-range corrections}} \\
SpookyNet \citep{unke_spookynet_2021},
  TensorMol \citep{yaoTensorMol01ModelChemistry2018},
  AIMNet \citep{zubatyukAccurateTransferableMultitask2019,anstine_machine_2023}
  & \Circle & \Circle & \Circle & \Circle & \Circle & $\mathcal{O}(N^2)$ \\
SO3LR \citep{kabyldaMolecularSimulationsPretrained2024},
  FENNOL \citep{pleForceFieldEnhancedNeuralNetwork2023,pleFeNNolEfficientFlexible2024}
  & \Circle & \Circle & \Circle & \Circle & \Circle & $\mathcal{O}(N \cdot \langle N_{\text{LR}} \rangle)$ \\
4G--HDNNP \citep{koFourthgenerationHighdimensionalNeural2021},
  CENT/CENT2 \citep{ghasemiInteratomicPotentialsIonic2015, khajehpashaCENT2ImprovedCharge2022},
  CELLI \citep{fuchsLearningNonlocalMolecular2025},
  FeNNix-Bio1 \citep{pleFoundationModelAccurate2025a},
  \citet{marufLearningLongRangeInteractions2025a}
  & \Circle & \Circle & \Circle & \Circle & \Circle & $\mathcal{O}(N \log N)$\textsuperscript{a} \\
LES \citep{chengLatentEwaldSummation2025, kim_universal_2025},
  Neural P$^3$M \citep{wangNeuralP3MLongRange2024},
  Ewald MP \citep{kosmalaEwaldbasedLongRangeMessage2023},
  Reciprocal Space NN \citep{yu_capturing_2022}
  & \Circle & \Circle & \Circle & \Circle & \Circle & $\mathcal{O}(N \log N)$ \\
LODE \citep{grisafiIncorporatingLongrangePhysics2019, huguenin-dumittanPhysicsInspiredEquivariantDescriptors2023, fallerDensityBasedLongRangeElectrostatic2024}
  & \Circle & \Circle & \LEFTcircle & \Circle & \Circle & $\mathcal{O}(N \log N)$ \\
sGDML \citep{chmielaAccurateGlobalMachine2023}
  & \Circle & \CIRCLE & \CIRCLE & \Circle & \Circle & $\mathcal{O}(N^2)$ \\
SOG-Net \citep{jiMachineLearningInteratomicPotentials2025}
  & \Circle & \CIRCLE & \Circle & \Circle & \Circle & $\mathcal{O}(N \log N)$ \\
MLXDM \citep{tuNeuralNetworkPotential2023, tuModelingIntermolecularInteractions2024}
  & \Circle & \Circle & \Circle & \Circle & \Circle & $\mathcal{O}(N \cdot \langle N_{\text{LR}} \rangle)$ \\
\midrule
\multicolumn{7}{l}{\emph{Equivariant long-range corrections}} \\
LOREM \citep{rumiantsevLearningLongRangeRepresentations2026}
  & \CIRCLE & \Circle & \LEFTcircle & \Circle & \Circle & $\mathcal{O}(N \log N)$ \\
EquiEwald \citep{zhangSO3equivariantReciprocalspaceNeural2026}
  & \CIRCLE & \LEFTcircle & \Circle & \Circle & \Circle & $\mathcal{O}(N)$ \\
FieldMACE \citep{barrett_incorporating_2026}
  & \CIRCLE & \Circle & \LEFTcircle & \Circle & \Circle & $\mathcal{O}(N+N_{\text{env}})$\textsuperscript{b} \\
MACE-POLAR-1 \citep{batatia_mace-polar-1_2026}
  & \CIRCLE & \Circle & \LEFTcircle & \Circle & \Circle & $\mathcal{O}(N \log N)$ \\
AMP \citep{thurlemannAnisotropicMessagePassing2022}
  & \CIRCLE & \Circle & \Circle & \Circle & \Circle & $\mathcal{O}(N^2)$ \\
LSR-MP \citep{liLongShortRangeMessagePassingPhysicsInformed2024}
  & \CIRCLE & \CIRCLE & \LEFTcircle & \Circle & \Circle & $\mathcal{O}(N \!\cdot\! N_{\text{frag}})$ \\
MFN \citep{Batatia2024EquivariantNetworks}
  & \CIRCLE & \CIRCLE & \LEFTcircle & \LEFTcircle & \Circle & $\mathcal{O}(N^3)$\textsuperscript{c} \\
\midrule
\multicolumn{7}{l}{\emph{Hierarchical graph neural networks}} \\
DiffPool \citep{yingHierarchicalGraphRepresentation2019}
  & \Circle & \CIRCLE & \LEFTcircle & \Circle & \CIRCLE & $\mathcal{O}(N^3)$\textsuperscript{d} \\
HGNet \citep{rampasekHierarchicalGraphNeural2021}
  & \Circle & \CIRCLE & \LEFTcircle & \Circle & \CIRCLE & $\mathcal{O}(N)$\textsuperscript{e} \\
MGT \citep{ngoMultiresolutionGraphTransformers2023}
  & \Circle & \CIRCLE & \LEFTcircle & \Circle & \CIRCLE & $\mathcal{O}(N^2)$\textsuperscript{f} \\
EGHN \citep{hanEquivariantGraphHierarchyBased2022}
  & \CIRCLE & \CIRCLE & \LEFTcircle & \Circle & \CIRCLE & $\mathcal{O}(N)$\textsuperscript{g} \\
DualEquiNet \citep{xuDualEquiNetDualSpaceHierarchical2025}
  & \CIRCLE & \CIRCLE & \LEFTcircle & \Circle & \CIRCLE & $\mathcal{O}(N^2)$\textsuperscript{h} \\
MGVAE \citet{hyMultiresolutionEquivariantGraph2023}
  & \Circle & \CIRCLE & \LEFTcircle & \Circle & \CIRCLE & $\mathcal{O}(N^3)$\textsuperscript{i} \\
\midrule
\multicolumn{7}{l}{\emph{Global attention}} \\
RANGE \citep{carusoExtendingRangeGraph2026}
  & \Circle & \CIRCLE & \LEFTcircle & \Circle & \Circle & $\mathcal{O}(N)$ \\
EFA \citep{frankMachineLearningGlobal2026}
  & \CIRCLE & \CIRCLE & \LEFTcircle & \CIRCLE & \Circle & $\mathcal{O}(N)$ \\
AllScAIP \citep{quRecipeScalableAttentionbased2026}
  & \LEFTcircle & \CIRCLE & \LEFTcircle & \Circle & \Circle & $\mathcal{O}(N^2)$ \\
UMA \citep{woodUMAFamilyUniversal2026}
  & \LEFTcircle & \CIRCLE & \LEFTcircle & \Circle & \Circle & $\mathcal{O}(N^2)$ \\
Erwin \citep{zhdanovErwinTreebasedHierarchical2025}
  & \Circle & \CIRCLE & \LEFTcircle & \Circle & \CIRCLE & $\mathcal{O}(N)$ \\
\midrule
\multicolumn{7}{l}{\emph{FMM-inspired neural architectures}} \\
MGNO \citep{li_multipole_2020},
  FMM-Net \citep{sushnikovaFMMNetNeuralNetwork2022}
  & \Circle & \CIRCLE & \CIRCLE & \Circle & \CIRCLE & $\mathcal{O}(N)$ \\
MANO \citep{colagrandeLinearAttentionGlobal2025},
  FMA \citep{kangFastMultipoleAttention2023}
  & \Circle & \CIRCLE & \CIRCLE & \Circle & \CIRCLE & $\mathcal{O}(N)$ \\
Neural FMM \citep{fogniniLearningGreensOperators2025},
  CloudNFMM \citep{fogniniCloudNFMM2026} & \Circle & \CIRCLE & \CIRCLE & \Circle & \CIRCLE & $\mathcal{O}(N)$\textsuperscript{j} \\
\midrule
\textbf{NEMORA (ours)}
  & \CIRCLE & \CIRCLE & \CIRCLE & \CIRCLE & \CIRCLE & $\mathcal{O}(N)$ \\
\bottomrule
\end{tabular}
\\[2pt]
{\scriptsize
\textsuperscript{a}Accelerated QEq \citep{gublerAcceleratingFourthGenerationMachine2024b}; dense solves are $\mathcal{O}(N^3)$.
\textsuperscript{b}Global embedding; per-atom embedding is $\mathcal{O}(NN_{\mathrm{env}})$.
\textsuperscript{c}Dense matrix-function evaluation.
\textsuperscript{d}Dense pooling at fixed pooling ratio.
\textsuperscript{e}Sparse graphs.
\textsuperscript{f}Full attention.
\textsuperscript{g}Sparse graphs, fixed cluster count.
\textsuperscript{h}Dense neighborhoods.
\textsuperscript{i}Dense second-order variant.
\textsuperscript{j}Linear-size tree, sparse FMM evaluation, bounded leaf occupancy; fixed-tree CloudNFMM local attention is $\mathcal{O}(N^2)$.
}
\end{table}

\clearpage
\section{NEMORA Operator Details}
\subsection{Symmetric contractions}
\label{si:symcontract}
Let $\bm\Psi_t=\sum_{s\in\mathcal N_t}\bm m_{st}$ be the pooled gated M2L field, with $\mathcal N_t$ the M2L interaction list. ACE/MACE-style contractions~\citep{drautz_atomic_2019,batatiaMACEHigherOrder2023} form degree-$\nu$ products:
\begin{equation}
\label{eq:symcontract}
\mathrm{SymContract}_{\nu}(\bm\Psi_t)
\sim\Bigl[\Bigl(\sum_{s\in\mathcal N_t}\bm m_{st}\Bigr)^{\otimes\nu}\Bigr]_{\mathrm{CG}}
=\sum_{s_1,\ldots,s_\nu\in\mathcal N_t}
\bigl[\bm m_{s_1t}\otimes\cdots\otimes\bm m_{s_\nu t}\bigr]_{\mathrm{CG}},
\end{equation}
Here $[\cdot]_{\mathrm{CG}}$ couples to the retained output irreps; learned weights and channel indices are suppressed. At each level $L$, the residual update is $\tilde{\bm\Psi}^{(\ell_M,L)}=\bm\Psi^{(\ell_M,L)}+[\mathrm{SymContract}_{\nu}(\bm\Psi^{(L)})]^{(\ell_M)}$. Cross-cluster terms provide features for non-additive interactions, such as dispersion and cooperative polarization~\citep{axilrodInteractionVanWaals1943,mutoForceNonpolarMolecules1943,tkatchenkoAccurateEfficientMethod2012}. Disabling contractions leaves the gated M2L operator, whose inputs and gates may still encode many-body information. Cost is linear in node count at fixed degrees, channels and order. The factorized expansion identity and equivariance are proved in Appendix~\ref{si:symcontract-proof}.

\subsection{Kelvin transform}
\label{si:kelvin}
For a spherical support of radius $R_k$ centered at $\bm c_k$, NEMORA uses a learned response inspired by Kelvin inversion and continuum solvation~\citep{amaral_kelvin_2017,bornVolumenUndHydratationswaerme1920,onsagerElectricMomentsMolecules1936,kirkwoodDielectricPolarizationPolar1939}:
\begin{equation}
\label{eq:nemora-kelvin}
\bm{\Psi}_k^{(\ell_{M})} = \sum_{s=1}^{N} \bm{\alpha}_{sk}^{(\ell_M)} \cdot \Bigl(
\tilde{R}_k^{-(2\ell_M+1)} \cdot \bm R_\theta^{(\ell_M)}(r_{sk}) \cdot \bm{W}^V \bm{\Psi}_s^{(\ell_{M})} \Bigr)
\end{equation}
Here $\bm\alpha_{sk}^{(\ell_M)}$ is an invariant gate, $\tilde R_k=R_k/r_{\mathrm{cut}}$, and $r_{sk}=\|\bm r_s-\bm c_k\|$. The radial modulation $\bm R_\theta^{(\ell_M)}$ uses an independently parameterized polynomial encoder, and $\bm W^V$ mixes channels within the same irrep. Global supports pool and broadcast this response; cluster-local corrections enter the downward local state. Field addition and equivariance are established in Appendices~\ref{si:kelvin-global} and~\ref{si:kelvin-local}. The learned degree-diagonal response need not be an exact tensor Kelvin inversion. Cost is $\mathcal O(NK)$ for $K$ global supports and $\mathcal O(N)$ for node-local updates on a linear-size hierarchy.

\subsection{Periodic boundary correction}
\label{si:pbc-tail}
For a periodic cell of volume $V=|\det\bm H|$, the learned branch pools top-level roots into $\bm\Psi_\Box^{(\ell_M)}$ and applies a finite-shell, degree-diagonal correction motivated by periodic FMM and electrostatic boundary terms~\citep{kudinFastMultipoleMethod1998,deleeuwSimulationElectrostaticSystems1980}:
\begin{equation}
\label{eq:pbc-tinfoil}
\tilde{\bm{\Psi}}_\Box^{(\ell_M)} = \Bigl[ (1 - \delta_{\ell_M, 0})\mathcal{S}_{\mathrm{lat}}^{(\ell_M)} + \delta_{\ell_M, 1}\frac{4\pi r_{\mathrm{cut}}^3}{3V} \Bigr] \bm{\Psi}_\Box^{(\ell_M)}\,, \qquad \mathcal{S}_{\mathrm{lat}}^{(\ell_M)} = \gamma_{\mathrm{pbc}} \sum_{\bm{n} \in \mathcal{N}_{\mathrm{lat}}} \left(\frac{r_{\mathrm{cut}}}{\|\bm{R}_{\bm{n}}\|}\right)^{2\ell_M+1}
\end{equation}
Here $\mathcal N_{\mathrm{lat}}=\{\bm n\in\mathbb Z^3:\|\bm n\|_\infty\le n_{\mathrm{far}}\}\setminus\{\bm0\}$ and $\bm R_{\bm n}=\bm H\bm n$. The deltas suppress this correction's monopole and add a volume-dependent dipole term, without enforcing charge neutrality. This is a modeled root response, not an exact tinfoil lattice sum; equivariance is proved in Appendix~\ref{si:pbc-equivariance}. Cell derivatives contribute to stress. L2L reinjection is $\mathcal O(N)$ at fixed shell and truncation under Appendix~\ref{si:linear-time}'s hierarchy assumptions. The analytical branch instead uses the periodic operator below.

\subsection{Analytical NEMORA}
\label{si:latent-fmm}
The analytical variant uses scalar latent charges, fixed P2M--L2P coefficients and extensive summation, without learned transport, gates, contractions or Kelvin responses. Every pair enters either an admissible multipole interaction or direct P2P evaluation, including pairs inside the backbone cutoff. A charge--potential contraction replaces the learned long-range update and readout; only the short-range backbone and its energy and charge heads remain trainable.

Scalar heads on MACE features give $q_i=\sum_{t^\star}\mathcal{Q}_{t^\star}(\bm{h}_i^{(t^\star)})$, with energy
\begin{equation}
\label{eq:latent-fmm-energy}
E = E_{\mathrm{SR}} + \frac{1}{2}\widetilde{\bm q}^{\top}K_{\sigma}^{\mathrm{BC}}\widetilde{\bm q},
\qquad
(K_{\sigma}^{\mathrm{free}})_{ij}
= c\,\frac{\operatorname{erf}(r_{ij}/(\sqrt{2}\sigma))}{r_{ij}}
\quad (i\ne j),
\end{equation}
Here $r_{ij}=\|\bm r_i-\bm r_j\|$, $\sigma=1$~\AA, $c=90.4756/(2\pi)$ follows LES, and the free-space diagonal is zero. A fixed Cartesian Taylor FMM~\citep{dehnenHierarchicalForceCalculation2002} evaluates the Coulomb part, with a direct $\operatorname{erfc}(r/(\sqrt2\sigma))/r$ subtraction. Under PBC, images with $\|\bm n\|_\infty\le w$ enter P2P/M2L; a precomputed root operator represents the remaining lattice sum, omitting the zero reciprocal mode and matching LES's self-interaction convention. With identical sources, smoothing and boundary conventions, this is the LES energy functional~\citep{chengLatentEwaldSummation2025} evaluated by FMM. Reported periodic FMM runs use $\widetilde q_i=q_i-N^{-1}\sum_jq_j$, while LES controls use raw charges; exact comparisons must match this projection. Nonperiodic runs use $\widetilde q_i=q_i$. Forces include coordinate derivatives of the charges.

Implemented tensor-source extensions predict latent dipoles ($1\mathrm{o}$) and traceless quadrupoles ($2\mathrm{e}$) from equivariant features~\citep{stoneDistributedMultipoleAnalysis1981}, using fixed kernel derivatives throughout, including P2P and periodic terms. Zero-initialized heads recover the scalar model. Source degree $\ell_{\mathrm{source}}$ adds atomic anisotropy; box order $p$ controls expansion accuracy. Translations remain analytical, unlike NEMORA's learned transport and many-body coupling.

\clearpage

\clearpage
\section{Additional Experimental Results}
\label{si:additional-results}

\subsection{Reporting conventions and benchmark runs}
For run-level errors $x_1,\ldots,x_n$, we report the arithmetic mean and $\mathrm{SEM}=s/\sqrt n$, where $s$ is the sample standard deviation with denominator $n-1$. Energy errors are normalized per atom in each frame before aggregation; force RMSE pools Cartesian components within each run. SEMs below the displayed precision are written as upper bounds rather than zero.

Lowest means are bold and second-lowest underlined, retaining ties. Rankings compare means, not statistical significance; SEMs describe between-run variability. Comparisons are within each dataset and metric, with narrower groups identified in the captions.

All models trained here use the benchmark splits. Run counts are given in Table~\ref{tab:benchmark-seeds}; EFA Au$_2$/MgO averages two runs. Per-run energy RMSE is $10^3[\frac{1}{M}\sum_{i=1}^{M}((\widehat E_i-E_i)/N_i)^2]^{1/2}$, where $M$ is the number of evaluation frames, $N_i$ the atom count and energies are in eV. These RMSEs are averaged across runs without removing a constant energy offset.

\subsection{Charge-equilibration and electrostatic references}
The charge-supervised models ($^{\ast}$) and SpookyNet in Table~\ref{tab:charge-equilibration} differ in supervision, cost and training protocol~\citep{fuchsLearningNonlocalMolecular2025,koFourthgenerationHighdimensionalNeural2021,unke_spookynet_2021}.

\paragraph{Supervision.} CELLI and 4G--HDNNP fit electronegativities to DFT Hirshfeld charges. SpookyNet takes total charge and spin as inputs and fits partial charges to reference dipoles or charges, omitting electrostatics when neither is available. Other models in the table use only energies and forces; MACE--LES and NEMORA infer latent charges or multipoles without charge supervision. The NaCl and Au$_2$/MgO releases~\citep{koFourthgenerationHighdimensionalNeural2021} include per-atom charges, but cumulene does not. For its charge-supervised models, we computed Hirshfeld charges with ORCA 6.1.1 at $\omega$B97X-D3/def2-TZVP, matching the dataset.

\paragraph{Cost and protocol.} Both charge-equilibration methods use a direct constrained linear solve, described in their papers as cubic in atom count; iterative and particle-mesh alternatives are discussed but not benchmarked. SpookyNet uses explicit all-pairs electrostatics and standard Ewald summation for periodic cells. Settings are in Appendix~\ref{si:electrostatic-settings}. Published nonlocal charge-transfer results ($^{\dagger}$) use splits not established to match ours, so their rankings are not controlled comparisons. CELLI and Allegro values follow \citet{fuchsLearningNonlocalMolecular2025}; 2G--BPNN, 4G--HDNNP and SpookyNet follow \citet{frankMachineLearningGlobal2026}. SpookyNet cumulene RMSEs are from \citet{Batatia2024EquivariantNetworks}.

\paragraph{Results.} Charge equilibration gives lower energy errors on the monopolar charge-transfer benchmarks Na$_8$Cl$_8^+$/Na$_9$Cl$_8^+$ and Au$_2$/MgO, including with two-interaction CELLI--MACE at our backbone depth. Charge labels and the prescribed Coulomb functional may contribute, but differing protocols prevent causal attribution. NEMORA has the lowest carbon-chain energy error and the lowest or second-lowest force error on every nonlocal charge-transfer system. On all three cumulene splits, the electrostatic reference models have higher force errors than short-range MACE (Table~\ref{tab:charge-equilibration}).

\begin{table}[!htbp]
\centering\small
\setlength{\tabcolsep}{3pt}
{\def\twoline#1#2{\begin{tabular}[b]{@{}c@{}}#1\\#2\end{tabular}}%
\def\threeline#1#2#3{\begin{tabular}[b]{@{}c@{}}#1\\#2\\#3\end{tabular}}%
\caption{Charge-equilibration and explicitly electrostatic references on the cumulene and nonlocal charge-transfer benchmarks, energy and force RMSEs, with our MACE and NEMORA runs from Table~\ref{tab:cumulene_splits}. Our cumulene test and nonlocal charge-transfer validation results are means across runs. $^{\dagger}$Published nonlocal charge-transfer results, which use study-specific splits, losses and architectures; all other values are our runs, except the published SpookyNet cumulene values. $^{\ast}$Additional reference-charge supervision; SR disables charge-equilibration electrostatics. SpookyNet also takes total charge and spin as inputs. In each row, lowest means are bold and second-lowest underlined, separately for energy and force (ties retained), with published results ranked together with our runs. Sources, run counts, charge-label generation and protocol details are given in this appendix.}
\label{tab:charge-equilibration}
\begin{adjustbox}{max width=\linewidth}
\begin{tabular}{ll*{8}{r}@{\hspace{12pt}}rr}
\toprule
 & & \threeline{CELLI--}{MACE}{(2)$^{\dagger\ast}$} & Allegro$^{\dagger}$ & \twoline{CELLI--}{Allegro$^{\dagger\ast}$} & \threeline{CELLI--}{Allegro}{($J^R$)$^{\dagger\ast}$} & \threeline{CELLI--}{Allegro}{($J^R$, SR)$^{\ast}$} & \twoline{2G--}{BPNN$^{\dagger}$} & \twoline{4G--}{HDNNP$^{\dagger\ast}$} & SpookyNet$^{\dagger}$ & MACE & \twoline{\textbf{NEMORA}}{\textbf{(Ours)}} \\
\midrule
\multicolumn{12}{@{}l}{\textit{Energy RMSE (meV/atom)}} \\
\multirow{3}{*}{Cumulene} & ID & $58.791$ & --- & --- & $59.431$ & $59.588$ & $42.815$ & $41.892$ & $30.800$ & $\underline{16.898}$ & $\mathbf{3.203}$ \\
 & SOD & $17.089$ & --- & --- & $16.853$ & $16.798$ & $16.733$ & $16.768$ & $31.400$ & $\underline{16.709}$ & $\mathbf{0.744}$ \\
 & OOD & $11.202$ & --- & --- & $11.133$ & $\underline{11.116}$ & $12.070$ & $11.317$ & $23.400$ & $11.994$ & $\mathbf{3.770}$ \\
\addlinespace[3pt]
\multirow{3}{*}{Charged} & Carbon chain & $0.398$ & $0.772$ & $0.599$ & $0.609$ & --- & $1.619$ & $1.194$ & $\underline{0.364}$ & $0.425$ & $\mathbf{0.175}$ \\
 & NaCl & $\mathbf{0.097}$ & $1.612$ & $0.127$ & $\underline{0.114}$ & --- & $1.692$ & $0.481$ & $0.135$ & $1.600$ & $0.412$ \\
 & Au$_2$/MgO & $\mathbf{0.069}$ & $2.329$ & $\underline{0.077}$ & $\underline{0.077}$ & --- & $2.287$ & $0.219$ & $0.107$ & $2.385$ & $0.197$ \\
\midrule
\multicolumn{12}{@{}l}{\textit{Force RMSE (meV/\AA)}} \\
\multirow{3}{*}{Cumulene} & ID & $204.30$ & --- & --- & $208.78$ & $212.85$ & $221.06$ & $210.82$ & $162.30$ & $\underline{85.07}$ & $\mathbf{26.85}$ \\
 & SOD & $126.99$ & --- & --- & $135.93$ & $145.78$ & $174.54$ & $161.90$ & $116.20$ & $\underline{102.16}$ & $\mathbf{47.87}$ \\
 & OOD & $104.26$ & --- & --- & $115.81$ & $123.74$ & $126.15$ & $127.46$ & $87.60$ & $\underline{72.26}$ & $\mathbf{57.01}$ \\
\addlinespace[3pt]
\multirow{3}{*}{Charged} & Carbon chain & $21.45$ & $49.18$ & $31.00$ & $32.31$ & --- & $129.50$ & $78.00$ & $\underline{5.80}$ & $15.00$ & $\mathbf{1.88}$ \\
 & NaCl & $3.54$ & $47.51$ & $6.44$ & $5.15$ & --- & $57.39$ & $32.78$ & $\mathbf{1.05}$ & $43.89$ & $\underline{1.75}$ \\
 & Au$_2$/MgO & $7.95$ & $123.67$ & $12.04$ & $12.01$ & --- & $153.10$ & $66.00$ & $\underline{5.34}$ & $110.07$ & $\mathbf{5.26}$ \\
\bottomrule
\end{tabular}
\end{adjustbox}}
\end{table}

\begin{table}[!htbp]
\centering\small
\caption{Numbers of runs contributing to the benchmark means and SEMs. Counts apply to energy and force unless noted. The separate 3--64-carbon sweep uses three runs for LOREM, EFA and NEMORA and one for the other models. NR denotes that no run count or no result is reported. Additional cumulene controls have three-run means but no available SEM in the source summary; SpookyNet is a published reference with no run count given there.}
\label{tab:benchmark-seeds}
\begin{tabular}{lrrrr}
\toprule
Model & Cumulene & Carbon chain & NaCl & Au$_2$/MgO \\
\midrule
MACE & 3 & 3 & 3 & 3 \\
MACE--LES & 3 & 3 & 3 & 3 \\
MACE--RANGE & 3 & 3 & 3 & 2 \\
LOREM & 3 & 3 & 3 & 3 \\
EFA & 3 & 3 & 3 & 2 \\
MACE--MFN & 3 & 3 & 3 & NR \\
NEMORA & 3 & 3 & 3 & 3 \\
CELLI--MACE and short-range control & 3 & NR & NR & NR \\
CELLI--Allegro and short-range control & 3 & NR & NR & NR \\
4G--HDNNP and 2G--BPNN & 3 & NR & NR & NR \\
\bottomrule
\end{tabular}
\end{table}

\subsection{Additional cumulene results}

\begin{table}[!htbp]
\centering\small
\setlength{\tabcolsep}{3pt}
\caption{Additional cumulene energy and force errors. Validation and aggregate-test results are mean $\pm$ SEM. The 3--64-carbon length/angle experiment uses separately trained models; LOREM, EFA and NEMORA have three seeds on that corpus and are quoted as mean $\pm$ SEM, while MACE, MACE--LES, MACE--RANGE and MACE--MFN have one and are point estimates. Bold and underlined entries mark the lowest and second-lowest errors. LOREM has the lowest force error on that corpus; the lowest energy error is the single-run MACE--MFN result.}
\label{tab:cumulene-additional}
\begin{adjustbox}{max width=\linewidth}
\begin{tabular}{lrrr@{\hspace{12pt}}rrr}
\toprule
 & \multicolumn{3}{c}{Energy MAE (meV/atom)} & \multicolumn{3}{c}{Force RMSE (meV/\AA)} \\\cmidrule(lr){2-4}\cmidrule(lr){5-7}
Model & Validation & Test & $n_C=3$--$64$ & Validation & Test & $n_C=3$--$64$ \\
\midrule
MACE \citep{batatiaMACEHigherOrder2023} & $8.915 \pm 0.153$ & $11.833 \pm 0.432$ & $10.373$ & $109.71 \pm 0.75$ & $86.41 \pm 0.30$ & $81.19$ \\
MACE--LES \citep{chengLatentEwaldSummation2025} & $8.671 \pm 0.200$ & $11.471 \pm 0.166$ & $10.323$ & $98.16 \pm 0.50$ & $86.40 \pm 0.42$ & $80.73$ \\
MACE--RANGE \citep{carusoExtendingRangeGraph2026} & $11.445 \pm 0.822$ & $14.184 \pm 1.013$ & $12.977$ & $121.53 \pm 2.90$ & $101.66 \pm 3.18$ & $81.97$ \\
LOREM \citep{rumiantsevLearningLongRangeRepresentations2026} & $4.678 \pm 1.198$ & $16.765 \pm 1.678$ & $5.068 \pm 2.159$ & $60.60 \pm 5.87$ & $132.89 \pm 19.68$ & $\mathbf{23.03 \pm 0.97}$ \\
EFA \citep{frankMachineLearningGlobal2026} & $4.890 \pm 1.789$ & $15.912 \pm 1.267$ & $4.083 \pm 0.734$ & $\underline{56.61 \pm 8.80}$ & $113.03 \pm 10.81$ & $\underline{31.07 \pm 0.87}$ \\
MACE--MFN \citep{Batatia2024EquivariantNetworks} & $\mathbf{1.211 \pm 0.109}$ & $\mathbf{0.970 \pm 0.297}$ & $\mathbf{0.984}$ & $\mathbf{22.08 \pm 1.69}$ & $\mathbf{14.81 \pm 0.82}$ & $31.86$ \\
NEMORA & $\underline{1.377 \pm 0.069}$ & $\underline{1.400 \pm 0.100}$ & $\underline{1.478 \pm 0.137}$ & $57.05 \pm 7.48$ & $\underline{48.63 \pm 1.96}$ & $36.83 \pm 3.07$ \\
\bottomrule
\end{tabular}
\end{adjustbox}
\end{table}

\subsection{Backbone portability}
\label{si:backbone-protocol}

Table~\ref{tab:backbone-carbon-chain} compares short-range, scalar-source and degree-two NEMORA models across backbones. Interfaces are defined in Appendix~\ref{si:backbone-settings}.

\begin{table}[!htbp]
\centering\small
\setlength{\tabcolsep}{3pt}
\caption{Carbon-chain backbone and source-degree validation ablations. Each entry gives mean $\pm$ SEM over three runs. Energy and force RMSEs are in meV/atom and meV/\AA. Bold and underlining mark the lowest and second-lowest means across SR, scalar and degree-two NEMORA for each host and metric. Degree-two sources are lifted for scalar and untyped hosts. See Appendix~\ref{si:backbone-settings} for the backbone interfaces.}
\label{tab:backbone-carbon-chain}
\begin{adjustbox}{max width=\linewidth}
\begin{tabular}{lllrrr}
\toprule
Backbone & Host & Metric & SR & $\ell=0$ & $\ell=2$ \\
\midrule
MACE \citep{batatiaMACEHigherOrder2023} & Native & Energy & $\underline{0.425 \pm 0.006}$ & $0.484 \pm 0.024$ & $\mathbf{0.163 \pm 0.015}$ \\
 &  & Force & $\underline{15.15 \pm 0.51}$ & $16.70 \pm 0.46$ & $\mathbf{2.21 \pm 0.11}$ \\
 & Scalar & Energy & $\underline{0.623 \pm 0.006}$ & $0.889 \pm 0.037$ & $\mathbf{0.284 \pm 0.031}$ \\
 &  & Force & $\underline{25.91 \pm 0.51}$ & $27.39 \pm 0.48$ & $\mathbf{7.85 \pm 2.08}$ \\
NequIP \citep{batznerE3equivariantGraphNeural2022} & Native & Energy & $\underline{0.871 \pm 0.002}$ & $0.972 \pm 0.009$ & $\mathbf{0.182 \pm 0.006}$ \\
 &  & Force & $42.38 \pm 0.15$ & $\underline{32.16 \pm 0.12}$ & $\mathbf{2.85 \pm 0.18}$ \\
 & Scalar & Energy & $\underline{2.618 \pm 0.001}$ & $9.491 \pm 0.320$ & $\mathbf{0.448 \pm 0.158}$ \\
 &  & Force & $93.43 \pm 0.13$ & $\underline{47.21 \pm 1.23}$ & $\mathbf{5.97 \pm 0.92}$ \\
ViSNet \citep{Wang2024VisNet} & Native & Energy & $0.673 \pm 0.039$ & $\underline{0.670 \pm 0.009}$ & $\mathbf{0.500 \pm 0.016}$ \\
 &  & Force & $15.67 \pm 1.12$ & $\underline{11.56 \pm 0.13}$ & $\mathbf{3.49 \pm 0.12}$ \\
 & Scalar & Energy & $\underline{2.615 \pm 0.016}$ & $7.452 \pm 0.742$ & $\mathbf{0.820 \pm 0.121}$ \\
 &  & Force & $85.78 \pm 0.09$ & $\underline{48.67 \pm 0.17}$ & $\mathbf{6.35 \pm 0.35}$ \\
PET \citep{pozdnyakov_smooth_2023} & Untyped & Energy & $\underline{1.047 \pm 0.086}$ & $1.119 \pm 0.128$ & $\mathbf{0.944 \pm 0.019}$ \\
 &  & Force & $\underline{39.58 \pm 0.46}$ & $39.82 \pm 0.58$ & $\mathbf{35.06 \pm 1.07}$ \\
\bottomrule
\end{tabular}
\end{adjustbox}
\end{table}

\subsection{Component ablations}
\label{si:ablation-protocol}

Tables~\ref{tab:ablations} and~\ref{tab:ablation-splits} report retrained component ablations; Table~\ref{tab:ablation-definitions} defines each change. Each arm is compared with its corresponding FULL control.

\begin{table}[!htbp]
\centering\small
\setlength{\tabcolsep}{3pt}
\caption{Complete available component-ablation validation force RMSEs (meV/\AA), mean $\pm$ SEM over three runs. Each arm is retrained, and FULL is the corresponding retrained control. These controls differ from the benchmark configurations in Table~\ref{tab:cumulene_splits}. A dash denotes an unavailable result. Lowest means are bold and second-lowest underlined (ties retained).}
\label{tab:ablations}
\begin{adjustbox}{max width=\linewidth}
\begin{tabular}{lrrrr}
\toprule
Arm & Carbon chain & NaCl & Cumulene & Au$_2$/MgO \\
\midrule
FULL & $2.07 \pm 0.05$ & $1.89 \pm 0.05$ & $57.15 \pm 6.53$ & $\underline{8.50 \pm 0.17}$ \\
NU3 & $\mathbf{1.93 \pm 0.08}$ & --- & $50.31 \pm 1.22$ & --- \\
FMMCLASSIC & $\underline{1.95 \pm 0.03}$ & $2.78 \pm 0.23$ & $68.86 \pm 3.48$ & $\mathbf{4.86 \pm 0.10}$ \\
NOKELVIN & $2.12 \pm 0.04$ & $1.92 \pm 0.04$ & $61.21 \pm 9.06$ & $10.28 \pm 0.13$ \\
LMAX3 & $2.15 \pm 0.02$ & --- & --- & --- \\
M2LDECAY & $2.15 \pm 0.03$ & --- & --- & --- \\
NOLRHEAD & $2.18 \pm 0.06$ & $\underline{1.86 \pm 0.02}$ & $71.15 \pm 0.98$ & --- \\
M2LPIN & $2.21 \pm 0.03$ & $1.97 \pm 0.07$ & $67.19 \pm 3.43$ & $8.86 \pm 0.03$ \\
ELL3 & $2.23 \pm 0.13$ & --- & --- & --- \\
NU0 & $2.26 \pm 0.06$ & $2.33 \pm 0.02$ & $62.66 \pm 4.04$ & $8.95 \pm 0.09$ \\
MAXLEV1 & $2.64 \pm 0.16$ & $2.61 \pm 0.18$ & $\underline{49.18 \pm 5.24}$ & --- \\
MAXLEV3 & $2.75 \pm 0.24$ & --- & --- & --- \\
FIXEDTREE & $3.21 \pm 0.09$ & $3.56 \pm 0.22$ & $88.37 \pm 12.24$ & $8.58 \pm 0.06$ \\
LMAX1 & $3.52 \pm 0.33$ & $3.36 \pm 0.40$ & $106.03 \pm 3.31$ & $8.71 \pm 0.27$ \\
MAXL1 & $4.27 \pm 0.54$ & $2.51 \pm 0.09$ & $107.21 \pm 2.56$ & --- \\
NOFUSE & $14.94 \pm 0.52$ & $1.89 \pm 0.09$ & $58.03 \pm 8.18$ & --- \\
NOHIER & $15.01 \pm 0.25$ & $44.17 \pm 0.21$ & $89.41 \pm 7.53$ & $112.97 \pm 0.37$ \\
LAD2 & $16.86 \pm 0.50$ & --- & --- & --- \\
LMAX0 & $17.00 \pm 0.55$ & $3.50 \pm 0.22$ & $107.74 \pm 1.02$ & --- \\
LAD3 & $17.30 \pm 0.59$ & --- & --- & --- \\
MAXL0 & $17.53 \pm 0.95$ & --- & --- & --- \\
NOKESL & $2.32 \pm 0.07$ & $2.40 \pm 0.02$ & $64.76 \pm 5.17$ & --- \\
NOGATE & $2.39 \pm 0.18$ & $1.88 \pm 0.04$ & $\mathbf{47.67 \pm 9.72}$ & --- \\
NOVSQRT & --- & $\mathbf{1.83 \pm 0.05}$ & --- & --- \\
TRES1 & --- & $1.99 \pm 0.06$ & --- & --- \\
NU2 & --- & $2.00 \pm 0.07$ & --- & --- \\
NOHIERPBC & --- & --- & --- & $8.65 \pm 0.13$ \\
NOKELVPBC & --- & --- & --- & $8.77 \pm 0.10$ \\
SLABMASK & --- & --- & --- & $8.83 \pm 0.36$ \\
NSHELLS2 & --- & --- & --- & $8.87 \pm 0.14$ \\
MICSPLIT & --- & --- & --- & $8.92 \pm 0.23$ \\
NOTINFOIL & --- & --- & --- & $8.92 \pm 0.25$ \\
MACE & $15.00 \pm 0.32$ & $43.89 \pm 0.09$ & $105.43 \pm 2.19$ & $111.56 \pm 1.65$ \\
\bottomrule
\end{tabular}
\end{adjustbox}
\end{table}
\begin{table}[!htbp]
\centering\small
\setlength{\tabcolsep}{3pt}
\caption{Cumulene component-ablation test force RMSEs (meV/\AA), mean $\pm$ SEM from the per-run split evaluations. FULL is the three-run reference configuration. Lowest means are bold and second-lowest underlined (ties retained).}
\label{tab:ablation-splits}
\begin{adjustbox}{max width=\linewidth}
\begin{tabular}{lrrr}
\toprule
Arm & Test-ID & Test-SOD & Test-OOD \\
\midrule
FIXEDTREE & $67.52 \pm 20.60$ & $127.97 \pm 29.75$ & $104.35 \pm 8.19$ \\
FMMCLASSIC & $39.04 \pm 2.03$ & $151.67 \pm 15.65$ & $106.64 \pm 2.53$ \\
FULL & $\underline{28.53 \pm 1.77}$ & $58.39 \pm 9.79$ & $64.12 \pm 8.91$ \\
LMAX0 & $89.19 \pm 1.73$ & $104.11 \pm 0.66$ & $75.79 \pm 0.46$ \\
LMAX1 & $88.39 \pm 1.64$ & $130.28 \pm 7.88$ & $80.10 \pm 2.19$ \\
M2LPIN & $33.27 \pm 1.95$ & $96.72 \pm 36.88$ & $77.94 \pm 14.54$ \\
MACE & $86.00 \pm 0.40$ & $102.07 \pm 0.05$ & $71.91 \pm 0.08$ \\
MAXL1 & $89.75 \pm 2.48$ & $107.57 \pm 1.98$ & $81.72 \pm 1.62$ \\
MAXLEV1 & $278.23 \pm 48.46$ & $183.77 \pm 25.81$ & $116.91 \pm 10.55$ \\
NOFUSE & $30.92 \pm 2.85$ & $72.28 \pm 6.19$ & $101.14 \pm 40.98$ \\
NOGATE & $\mathbf{26.45 \pm 6.12}$ & $\underline{55.77 \pm 3.62}$ & $\underline{62.83 \pm 1.40}$ \\
NOHIER & $77.23 \pm 6.64$ & $151.20 \pm 28.10$ & $121.50 \pm 29.02$ \\
NOKELVIN & $29.45 \pm 2.30$ & $64.11 \pm 15.40$ & $70.20 \pm 9.44$ \\
NOKESL & $33.18 \pm 0.38$ & $\mathbf{44.96 \pm 3.06}$ & $\mathbf{54.27 \pm 2.60}$ \\
NOLRHEAD & $29.23 \pm 2.67$ & $87.29 \pm 35.32$ & $85.92 \pm 13.58$ \\
NU0 & $33.08 \pm 4.32$ & $77.38 \pm 22.92$ & $64.77 \pm 2.23$ \\
NU3 & $30.10 \pm 2.10$ & $62.60 \pm 9.87$ & $70.03 \pm 1.32$ \\
\bottomrule
\end{tabular}
\end{adjustbox}
\end{table}

Removing the hierarchy returns carbon-chain, NaCl and Au$_2$/MgO force errors close to short-range MACE, accounting for most of NEMORA's improvement. On cumulene, the Kelvin branch alone retains some improvement, but the hierarchy is needed for the lowest errors.

Consistent with the backbone experiments, restricting NEMORA to scalar features and operators increases carbon-chain force error from 2.07 to 17.00~meV/\AA. On NaCl, the scalar model still reaches 3.50~meV/\AA, below the short-range error of 43.89~meV/\AA. Cumulene is especially sensitive to feature degree: degree-zero and degree-one variants approach the short-range error. These joint feature/operator ablations also remove a quadrupole-dependent Kelvin head (Table~\ref{tab:ablation-definitions}).

Other effects are system-dependent. Removing the standalone Kelvin branch has little effect on the two small cluster systems but increases Au$_2$/MgO force error from 8.50 to 10.28~meV/\AA. Removing per-box Kelvin contributions increases NaCl error from 1.89 to 2.40~meV/\AA. FMMCLASSIC pins radial functions and removes many-body contraction and alignment gates while retaining learned source and energy readouts. It improves Au$_2$/MgO force error to 4.86~meV/\AA\ but worsens cumulene test errors, suggesting that simpler transport suffices for some predominantly electrostatic systems while cumulene benefits from more flexible operators.

Several smaller effects are comparable to between-run variation, particularly on cumulene. MAXLEV1 also illustrates differing validation and test behavior: its validation error is lower, but its ID test error is substantially higher.

\subsection{Ionic systems}
\label{si:analytical-ions}
We use equimolar binary systems with truncated Lennard-Jones interactions and fixed reference charges $\pm0.6\,e$: spherical clusters labeled by direct Coulomb summation and periodic cells labeled by Ewald/PPPM~\citep{ewaldBerechnungOptischerUnd1921,hockneyComputerSimulationUsing1988}. These are separate benchmarks from the small NaCl clusters of Table~\ref{tab:ko}. Training and validation contain $N=256,512,1024$; $N=2048,4096$ test size extrapolation. Model settings are given in Appendix~\ref{si:settings-ions}.

\begin{table}[!htbp]
\centering\small
\setlength{\tabcolsep}{3pt}
\caption{Periodic and non-periodic ionic test errors, E RMSE (meV/atom)\,/\,F RMSE (meV/\AA). Training includes systems with $N=256,512,1024$, while OOD denotes larger systems not included in training. $(q_{\rm Cl},q_{\rm Na})$ are the mean latent source charges. Within each system size and error metric, the lower mean is bold and the higher underlined.}
\label{tab:latent-fmm-ionic}
\label{tab:latent-fmm-ionic-energy}
\begin{adjustbox}{max width=\linewidth}
\begin{tabular}{@{}lrr@{}}
\toprule
System & MACE--LES & NEMORA ($\ell_Q=0$) \\
\midrule
\multicolumn{3}{@{}l}{\textit{Non-periodic clusters}} \\
\quad $N=256$ & $\mathbf{5.042}$\,/\,$\underline{16.81}$ & $\underline{5.266}$\,/\,$\mathbf{16.60}$ \\
\quad $N=512$ & $\mathbf{0.496}$\,/\,$\underline{16.93}$ & $\underline{0.532}$\,/\,$\mathbf{16.71}$ \\
\quad $N=1024$ & $\mathbf{4.538}$\,/\,$\underline{16.82}$ & $\underline{4.836}$\,/\,$\mathbf{16.60}$ \\
\quad $N=2048$ (OOD) & $\mathbf{8.000}$\,/\,$\underline{16.76}$ & $\underline{8.477}$\,/\,$\mathbf{16.56}$ \\
\quad $N=4096$ (OOD) & $\mathbf{10.792}$\,/\,$\underline{16.86}$ & $\underline{11.414}$\,/\,$\mathbf{16.83}$ \\
\quad $(q_{\rm Cl},q_{\rm Na})$ & $(-0.5999,+0.5999)$ & $(-0.5999,+0.5999)$ \\
\addlinespace[3pt]
\multicolumn{3}{@{}l}{\textit{Periodic cells}} \\
\quad $N=256$ & $\underline{0.308}$\,/\,$\underline{9.80}$ & $\mathbf{0.284}$\,/\,$\mathbf{9.64}$ \\
\quad $N=512$ & $\mathbf{0.203}$\,/\,$\underline{10.01}$ & $\underline{0.211}$\,/\,$\mathbf{9.83}$ \\
\quad $N=1024$ & $\underline{0.155}$\,/\,$\underline{9.88}$ & $\mathbf{0.133}$\,/\,$\mathbf{9.69}$ \\
\quad $N=2048$ (OOD) & $\underline{0.142}$\,/\,$\underline{9.95}$ & $\mathbf{0.109}$\,/\,$\mathbf{9.78}$ \\
\quad $N=4096$ (OOD) & $\underline{0.120}$\,/\,$\underline{9.95}$ & $\mathbf{0.090}$\,/\,$\mathbf{9.79}$ \\
\quad $(q_{\rm Cl},q_{\rm Na})$ & $(-0.6002,+0.6003)$ & $(-0.6000,+0.6000)$ \\
\bottomrule
\end{tabular}
\end{adjustbox}
\end{table}

Scalar-source analytical NEMORA matches LES closely in energy and force accuracy and extrapolates to larger systems under both boundary conditions (Table~\ref{tab:latent-fmm-ionic}). Force errors remain nearly constant across sizes; for non-periodic clusters, NEMORA has 0.2--1.3\% lower force RMSE at every size. Independently learned LES and NEMORA charges also agree with the reference charges, $\pm0.6$ in the benchmark's reduced units. Together, these results support recovery and size transfer of the known long-range functional and consistent learned scalar charges.

\subsection{Periodic water}
\label{si:analytical-water}
The revPBE-D3 corpus contains 1,366 training and 547 validation configurations with 384 or 768 atoms. Model settings are given in Appendix~\ref{si:settings-water}.

\begin{table}[!htbp]
\centering\small
\setlength{\tabcolsep}{3pt}
\caption{Periodic revPBE-D3 water test errors. Systems with $N=384$ and $768$ atoms are included in the training and testing sets. Energy error is in meV/atom and force error in meV/\AA. Appendix~\ref{si:settings-water} gives the numerical settings. $\langle q_{\rm O}\rangle$ and $\langle q_{\rm H}\rangle$ are test-set means of the latent scalar sources. Within the same system size and error metric, lowest means are bold and second-lowest underlined (ties retained).}
\label{tab:latent-fmm-water}
\begin{adjustbox}{max width=\linewidth}
\begin{tabular}{lrrrr}
\toprule
Model & Energy RMSE & Force RMSE & $\langle q_{\rm O}\rangle$ & $\langle q_{\rm H}\rangle$ \\
\midrule
\multicolumn{5}{l}{$N=384$} \\
MACE & $\underline{0.340}$ & $21.86$ & --- & --- \\
MACE--LES & $0.351$ & $\underline{15.75}$ & $-0.1919$ & $0.0959$ \\
NEMORA ($\ell_{\rm source}=0$) & $0.345$ & $15.88$ & $-0.2054$ & $0.1027$ \\
NEMORA ($\ell_{\rm source}=1$) & $\mathbf{0.259}$ & $\mathbf{13.07}$ & $-0.4766$ & $0.2383$ \\
\midrule
\multicolumn{5}{l}{$N=768$} \\
MACE & $0.499$ & $20.09$ & --- & --- \\
MACE--LES & $\underline{0.459}$ & $\underline{13.31}$ & $-0.1878$ & $0.0939$ \\
NEMORA ($\ell_{\rm source}=0$) & $0.474$ & $13.37$ & $-0.2011$ & $0.1006$ \\
NEMORA ($\ell_{\rm source}=1$) & $\mathbf{0.358}$ & $\mathbf{10.95}$ & $-0.4755$ & $0.2377$ \\
\bottomrule
\end{tabular}
\end{adjustbox}
\end{table}

LES and scalar-source NEMORA both improve force accuracy over bare MACE (Table~\ref{tab:latent-fmm-water}). Adding atomic dipoles ($\ell_\text{source}=1$) further improves energy and force errors in this single-seed experiment, at additional computational cost, and changes the mean scalar charges on O and H reported in the table. No quadrupolar accuracy comparison is reported.

With identical source, smoothing and boundary conventions, scalar analytical NEMORA and LES evaluate the same long-range energy functional by different numerical methods; the reported periodic runs differ in neutrality projection (Appendix~\ref{si:latent-fmm}). Particle-mesh LES could scale as $\mathcal{O}(N\log N)$, whereas the FMM translation chain scales as $\mathcal{O}(N)$ under the conditions below, motivating larger-system applications beyond direct summation.

\subsection{Numerical convergence}
\label{si:analytical-convergence}
We hold charges fixed to separate solver error from fitting error. For an admissible pair, $\rho=(R_s+R_t)/\|\bm c_t-\bm c_s\|<\theta<1$. Classical multipole estimates predict geometric decay of truncation error with $p$ at fixed separation; a smaller $\theta$ tightens admissibility at the expense of additional direct work~\citep{greengard_fast_1987,cheng_fast_1999}. Table~\ref{tab:latent-fmm-convergence} tests both controls in float64 on 96 uniformly sampled points in $[-9,9]^3$ with centered Gaussian charges, exercising all five translation stages. Increasing $p$ reduces energy and force errors, while tightening $\theta$ reduces force error. At $\theta=0.4$, every pair is routed to P2P, recovering direct summation to roundoff. The partition audit finds no missing or multiply counted pairs.

\begin{table}[!htbp]
\centering
\small
\caption{Fixed-charge numerical errors. In the free-space Coulomb test, $\epsilon_E=|E-E_{\mathrm{ref}}|/|E_{\mathrm{ref}}|$ and $\epsilon_{F,\infty}=\|\bm F-\bm F_{\mathrm{ref}}\|_\infty/\|\bm F_{\mathrm{ref}}\|_\infty$ use direct summation as reference. The periodic test uses one 256-ion configuration with charges $\pm0.6\,e$ and a converged Ewald reference (reciprocal-grid parameter $d_l=0.6$~\AA); its last column is the absolute force component RMS difference in meV/\AA. Here $w$ is the explicit image-shell radius. All entries use float64 and contain no prediction error against training labels.}
\label{tab:latent-fmm-convergence}
\begin{tabular}{rrrrr}
\toprule
\multicolumn{5}{l}{\emph{Free-space Coulomb, $N=96$}} \\
$p$ & $\theta$ & M2L rows & $\epsilon_E$ & $\epsilon_{F,\infty}$ \\
\midrule
2 & 0.5 & 4  & $3.01\times10^{-4}$ & $1.89\times10^{-4}$ \\
4 & 0.5 & 4  & $1.28\times10^{-5}$ & $1.01\times10^{-5}$ \\
6 & 0.5 & 4  & $5.66\times10^{-7}$ & $1.09\times10^{-6}$ \\
4 & 0.6 & 10 & $8.95\times10^{-6}$ & $2.38\times10^{-5}$ \\
4 & 0.4 & 0  & $2.56\times10^{-16}$ & $3.87\times10^{-16}$ \\
\midrule
\multicolumn{5}{l}{\emph{Periodic smoothed Coulomb, $N=256$}} \\
$p$ & $\theta$ & $w$ & $\epsilon_E$ & Force RMS difference \\
\midrule
3 & 0.50 & 1 & $8.77\times10^{-4}$ & 2.427 \\
3 & 0.50 & 2 & $4.22\times10^{-5}$ & 2.369 \\
3 & 0.35 & 1 & $1.09\times10^{-3}$ & 1.016 \\
3 & 0.35 & 2 & $1.60\times10^{-4}$ & 0.531 \\
4 & 0.35 & 2 & $8.04\times10^{-5}$ & 0.101 \\
5 & 0.35 & 2 & $3.86\times10^{-5}$ & 0.031 \\
\bottomrule
\end{tabular}
\end{table}

For periodic ions at $p=3,w=2$, tightening $\theta$ from 0.5 to 0.35 reduces force error by a factor of 4.46, while direct interaction rows increase from 312,989 to 915,502. Individual energy errors need not decrease monotonically because signed contributions can cancel. The observed trends agree with classical FMM convergence; they do not establish error bounds for NEMORA's learned operators. Tree geometry also affects finite-order error: on a 384-atom water configuration with fixed charges $q_{\mathrm H}=0.2$, $q_{\mathrm O}=-0.4$, the $p=3,\theta=0.35,w=2$ evaluator with maximum leaf occupancy 32 has relative energy error $3.03\times10^{-6}$ and force error 0.057~meV/\AA\ against converged Ewald. Occupancy 8 and the fixed-depth training tree give 0.345 and 0.168~meV/\AA, respectively. High-order force-loss derivatives require float64 in the current implementation. The reference smoothing correction enumerates all within-graph pairs, so these accuracy tests do not establish linear runtime for the complete analytical backend.

At fixed expansion order and bounded interaction lists, the analytical FMM translation chain has $\mathcal{O}(N)$ cost. Linear scaling, higher-degree tensor sources and learnable extensions motivate its use on larger systems.

\subsection{Water NVE energy conservation}
\label{si:water-nve}
\begin{figure}[!htbp]
    \centering
    \includegraphics[width=0.5\linewidth]{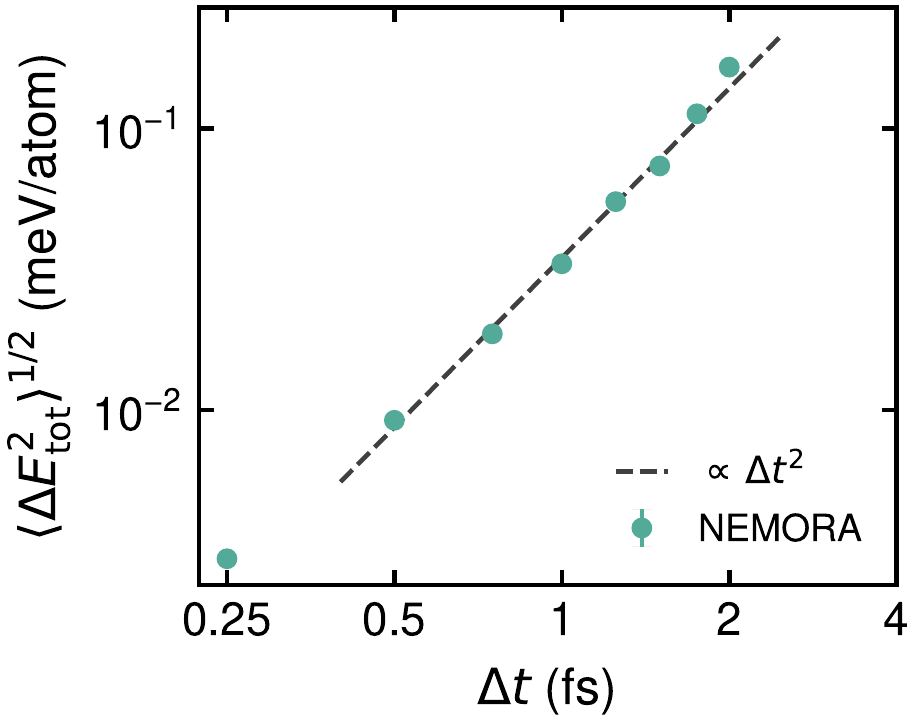}
    \caption{Root-mean-square total-energy fluctuation in 10~ps NVE trajectories of 128 water molecules with scalar NEMORA ($\ell_Q=0$) as a function of the velocity-Verlet timestep $\Delta t$. The dashed line is proportional to $\Delta t^2$.}
    \label{fig:nve-dt}
\end{figure}

The NVE tests use the scalar NEMORA model ($\ell_Q=0$) of Table~\ref{tab:periodic-ions}, velocity-Verlet integration and deuterium masses for hydrogen. The 200~ps, 512-molecule trace in Figure~\ref{fig:water}(c) follows 10~ps of NVT equilibration at 300~K and uses the same trained checkpoint at expansion order five. Relative to the NVT simulations in that figure, changing the expansion order changes only the numerical approximation, without retraining. The order-three timestep scan starts 128 water molecules from the same equilibrated state and runs for 10~ps at each timestep from 0.25 to 2~fs, rebuilding the hierarchy at every step. The RMS total-energy fluctuation about a linear fit (Figure~\ref{fig:nve-dt}) scales approximately as $\Delta t^2$ between 0.5 and 2~fs, with log--log slope 2.06, consistent with a second-order symplectic integrator. At 0.25~fs it approaches a fitted floor of about $2\times10^{-3}$~meV/atom, consistent with single-precision round-off.

\subsection{Elongated-water molecular dynamics and dipole correlations}
\label{si:water-elongated}
The MACE, MACE--LES and NEMORA simulations in Figure~\ref{fig:water} use the models of Table~\ref{tab:periodic-ions} and the shared backbone in Appendix~\ref{si:settings-water}. Each starts from the same equilibrated configuration of 2048 water molecules (6144 atoms) in a $24.84\times24.84\times99.35$~\AA\ orthorhombic periodic cell. We use NVT Langevin dynamics at 300~K, a 0.5~fs timestep, deuterium masses for hydrogen and a 50~fs frame-saving interval.

Molecular dipoles $\bm\mu_i$ are computed from Wannier-centroid positions predicted by the Deep Wannier model of the DPLR framework, trained on maximally localized Wannier centers of revPBE-D3 water~\citep{zhang_deep_2022}, with point charges $+6$ on O, $+1$ on H and $-8$ on the centroid. For wavevectors along the long axis, $\bm k=k\hat{\bm z}$ with $k=2\pi n/L_z$ and $n=1,2,\ldots$ up to $k=10$~\AA$^{-1}$, we compute
\begin{equation}
g_L(k)=\frac{\langle|\hat m_z(k)|^2\rangle}{N\langle\mu^2\rangle},\qquad \hat m_z(k)=\sum_{i=1}^{N}\mu_{i,z}\,e^{ikz_i},
\end{equation}
where $z_i$ is the oxygen coordinate and $N$ the number of molecules. The elongated cell gives a smallest wavevector of $0.063$~\AA$^{-1}$. We discard the first 5~ps and average over frames separated by 100~fs, giving more than 2000 frames per model. As $k\rightarrow0$, $g_L(k)$ approaches a finite value set by the static dielectric constant~\citep{chengLatentEwaldSummation2025}; a model without long-range electrostatics instead shows a spurious rise.

\subsection{Scaling protocol and complete measurements}
\label{si:scaling}
We benchmark fixed-density randomized particle clusters and cubic periodic boxes using the settings in Appendix~\ref{si:scaling-settings}, on one H200 with untrained models in full fp32 (TF32 disabled for every model). Timing is the median wall time of one energy evaluation with coordinate differentiation for forces, taken over 60 interleaved evaluations in a fresh process for each size. Memory is the peak device memory of one such evaluation with a single structure resident: the XLA buffer assignment of the compiled program (arguments, temporaries and outputs) for the JAX models, and the allocator peak reset before the call for the PyTorch models. Dataset preparation and compilation are excluded from timing and memory calculations.

Long-range exponents are log--log fits over the largest four sizes after subtracting the matched short-range cost: the same program with the long-range module disabled for the JAX models, the virtual-node blocks bypassed for RANGE, and a plain MACE of the same backbone for LES. Time uses the device kernel-busy difference and memory the difference of peaks. This reduces framework-specific fixed overhead in the fitted slopes but does not make absolute timings or memory statistics directly comparable across frameworks.

\begin{table}[!htbp]
\centering\small
\setlength{\tabcolsep}{3pt}
\caption{Energy-and-force wall time (ms) for non-periodic clusters on one H200. $\alpha_{\mathrm{LR}}$ is the fitted scaling exponent of the long-range cost. A dash marks a size that exceeded the GPU memory.}
\label{tab:scaling-1}
\begin{adjustbox}{max width=\linewidth}
\begin{tabular}{rrrrrrrr}
\toprule
$N$ & MACE & LES & RANGE & EFA & LOREM & MFN & NEMORA \\
\midrule
64 & $\mathbf{0.7}$ & 16.4 & 21.8 & $\underline{1.0}$ & 2.2 & 124.7 & 4.3 \\
128 & $\mathbf{0.8}$ & 18.3 & 24.4 & $\underline{1.5}$ & 2.5 & 635.4 & 4.7 \\
256 & $\mathbf{1.0}$ & 20.5 & 27.4 & $\underline{2.2}$ & 2.8 & 4601.6 & 6.2 \\
512 & $\mathbf{1.3}$ & 20.8 & 27.8 & $\underline{3.9}$ & 4.0 & 35133.5 & 6.9 \\
1024 & $\mathbf{1.8}$ & 21.0 & 28.3 & 6.7 & $\underline{5.8}$ & -- & 8.7 \\
2048 & $\mathbf{2.8}$ & 23.5 & 31.6 & 12.7 & 13.2 & -- & $\underline{11.9}$ \\
4096 & $\mathbf{5.6}$ & 42.3 & 46.0 & 31.4 & 44.0 & -- & $\underline{19.3}$ \\
8192 & $\mathbf{10.8}$ & 98.8 & 78.0 & 68.2 & 162.1 & -- & $\underline{37.5}$ \\
16384 & $\mathbf{21.4}$ & 277.9 & 143.7 & 138.2 & 702.4 & -- & $\underline{75.5}$ \\
32768 & $\mathbf{43.1}$ & 891.5 & 277.9 & 281.7 & 2971.6 & -- & $\underline{162.6}$ \\
65536 & $\mathbf{85.1}$ & -- & 547.2 & -- & 13533.2 & -- & $\underline{334.5}$ \\
131072 & $\mathbf{172.6}$ & -- & 1099.6 & -- & -- & -- & $\underline{606.9}$ \\
262144 & 345.8 & -- & -- & -- & -- & -- & 1238.6 \\
\midrule
$\alpha_{\mathrm{LR}}$ & -- & 1.99 & 0.99 & 1.02 & 2.21 & 2.77 & 0.96 \\
\bottomrule
\end{tabular}
\end{adjustbox}
\end{table}
\begin{table}[!htbp]
\centering\small
\setlength{\tabcolsep}{3pt}
\caption{Energy-and-force wall time (ms) for periodic boundaries on one H200. $\alpha_{\mathrm{LR}}$ is the fitted scaling exponent of the long-range cost. A dash marks a size that exceeded the GPU memory.}
\label{tab:scaling-2}
\begin{adjustbox}{max width=\linewidth}
\begin{tabular}{rrrrrrrr}
\toprule
$N$ & MACE & LES & RANGE & EFA & LOREM & MFN & NEMORA \\
\midrule
64 & $\underline{0.8}$ & 17.2 & 26.0 & $\mathbf{0.7}$ & 2.2 & 120.7 & 5.3 \\
128 & $\mathbf{0.9}$ & 19.1 & 27.9 & $\underline{1.0}$ & 2.7 & 636.4 & 5.6 \\
256 & $\mathbf{1.0}$ & 21.3 & 29.9 & $\underline{1.3}$ & 3.4 & 4615.0 & 7.3 \\
512 & $\mathbf{1.4}$ & 21.6 & 29.3 & $\underline{1.9}$ & 6.1 & 35150.2 & 8.3 \\
1024 & $\mathbf{1.9}$ & 21.9 & 28.7 & $\underline{2.6}$ & 11.1 & -- & 9.3 \\
2048 & $\mathbf{3.1}$ & 23.7 & 31.9 & $\underline{4.6}$ & 31.1 & -- & 13.4 \\
4096 & $\mathbf{6.0}$ & 42.5 & 47.6 & $\underline{8.8}$ & 110.0 & -- & 20.5 \\
8192 & $\mathbf{11.5}$ & 89.5 & 80.1 & $\underline{20.1}$ & 363.4 & -- & 37.6 \\
16384 & $\mathbf{22.2}$ & 233.9 & 146.1 & $\underline{40.9}$ & -- & -- & 80.3 \\
32768 & $\mathbf{44.8}$ & -- & 281.5 & $\underline{88.7}$ & -- & -- & 167.2 \\
65536 & $\mathbf{89.0}$ & -- & 554.3 & $\underline{191.4}$ & -- & -- & 340.8 \\
131072 & $\mathbf{177.0}$ & -- & 1106.4 & $\underline{399.4}$ & -- & -- & 662.1 \\
262144 & $\mathbf{352.2}$ & -- & -- & $\underline{816.5}$ & -- & -- & 1329.8 \\
\midrule
$\alpha_{\mathrm{LR}}$ & -- & 1.88 & 0.98 & 0.98 & 1.92 & 2.78 & 1.00 \\
\bottomrule
\end{tabular}
\end{adjustbox}
\end{table}
\begin{table}[!htbp]
\centering\small
\setlength{\tabcolsep}{3pt}
\caption{Peak GPU memory (MiB) for non-periodic clusters on one H200. $\alpha_{\mathrm{LR}}$ is the fitted scaling exponent of the long-range cost. A dash marks a size that exceeded the GPU memory.}
\label{tab:scaling-3}
\begin{adjustbox}{max width=\linewidth}
\begin{tabular}{rrrrrrrr}
\toprule
$N$ & MACE & LES & RANGE & EFA & LOREM & MFN & NEMORA \\
\midrule
64 & $\mathbf{43.6}$ & 99.3 & 110 & 62.7 & 57.1 & 1780 & $\underline{54.5}$ \\
128 & $\mathbf{61.2}$ & 134 & 154 & 87.7 & 98.9 & 6810 & $\underline{80.2}$ \\
256 & $\mathbf{85.9}$ & 181 & 215 & $\underline{123}$ & 171 & 26900 & 127 \\
512 & $\mathbf{153}$ & 303 & 372 & 221 & 328 & 107000 & $\underline{212}$ \\
1024 & $\mathbf{273}$ & 508 & 643 & 398 & 629 & -- & $\underline{369}$ \\
2048 & $\mathbf{511}$ & 1130 & 1220 & $\underline{602}$ & 1300 & -- & 738 \\
4096 & $\mathbf{1130}$ & 2990 & 2390 & 7000 & 2760 & -- & $\underline{1540}$ \\
8192 & $\mathbf{2360}$ & 9060 & 4740 & 14000 & 5720 & -- & $\underline{3470}$ \\
16384 & $\mathbf{4620}$ & 30600 & 9480 & 27700 & 12400 & -- & $\underline{6440}$ \\
32768 & $\mathbf{9150}$ & 111000 & 19100 & 55500 & 27800 & -- & $\underline{13000}$ \\
65536 & $\mathbf{18200}$ & -- & 38100 & -- & 64600 & -- & $\underline{25400}$ \\
131072 & $\mathbf{38700}$ & -- & 77300 & -- & -- & -- & $\underline{52100}$ \\
262144 & 78000 & -- & -- & -- & -- & -- & 106000 \\
\midrule
$\alpha_{\mathrm{LR}}$ & -- & 2.12 & 0.99 & 1.00 & 2.01 & 2.01 & 0.94 \\
\bottomrule
\end{tabular}
\end{adjustbox}
\end{table}
\begin{table}[!htbp]
\centering\small
\setlength{\tabcolsep}{3pt}
\caption{Peak GPU memory (MiB) for periodic boundaries on one H200. $\alpha_{\mathrm{LR}}$ is the fitted scaling exponent of the long-range cost. A dash marks a size that exceeded the GPU memory.}
\label{tab:scaling-4}
\begin{adjustbox}{max width=\linewidth}
\begin{tabular}{rrrrrrrr}
\toprule
$N$ & MACE & LES & RANGE & EFA & LOREM & MFN & NEMORA \\
\midrule
64 & $\mathbf{52}$ & 101 & 115 & $\underline{52.4}$ & 88.8 & 1780 & 64.4 \\
128 & $\underline{74.7}$ & 138 & 161 & $\mathbf{73.6}$ & 139 & 6820 & 90.8 \\
256 & $\underline{107}$ & 187 & 225 & $\mathbf{103}$ & 217 & 26900 & 154 \\
512 & $\underline{186}$ & 359 & 388 & $\mathbf{168}$ & 506 & 107000 & 271 \\
1024 & $\underline{322}$ & 688 & 669 & $\mathbf{273}$ & 1180 & -- & 440 \\
2048 & $\underline{591}$ & 1820 & 1270 & $\mathbf{471}$ & 4600 & -- & 858 \\
4096 & $\underline{1260}$ & 5670 & 2460 & $\mathbf{752}$ & 17500 & -- & 1730 \\
8192 & $\underline{2550}$ & 19700 & 4910 & $\mathbf{1470}$ & 66600 & -- & 3580 \\
16384 & $\underline{4700}$ & 72800 & 9750 & $\mathbf{2970}$ & -- & -- & 6830 \\
32768 & $\underline{9610}$ & -- & 19600 & $\mathbf{6000}$ & -- & -- & 14000 \\
65536 & $\underline{19500}$ & -- & 39200 & $\mathbf{12000}$ & -- & -- & 27200 \\
131072 & $\underline{38300}$ & -- & 78300 & $\mathbf{24000}$ & -- & -- & 51800 \\
262144 & $\underline{79100}$ & -- & -- & $\mathbf{47900}$ & -- & -- & 107000 \\
\midrule
$\alpha_{\mathrm{LR}}$ & -- & 2.10 & 1.00 & 0.87 & 2.39 & 2.01 & 0.89 \\
\bottomrule
\end{tabular}
\end{adjustbox}
\end{table}
\begin{table}[!htbp]
\centering\small
\setlength{\tabcolsep}{6pt}
\caption{Largest number of atoms $N$ that fits on one H200 with fp32 precision for each model, in non-periodic clusters and periodic boxes. Note that LOREM and EFA use a different short-range architecture and therefore are not directly comparable.}
\label{tab:scaling-5}
\begin{tabular}{lrr}
\toprule
Model & Non-periodic & Periodic \\
\midrule
MACE & 508{,}552 & 496{,}256 \\
LES & 37{,}096 & 23{,}128 \\
RANGE & 233{,}400 & 233{,}552 \\
EFA & 43{,}960 & 618{,}112 \\
LOREM & $\approx$115{,}000 & 11{,}520 \\
MFN & 512 & 512 \\
NEMORA & 347{,}512 & 330{,}096 \\
\bottomrule
\end{tabular}
\end{table}
\clearpage

\clearpage
\section{Datasets}
\label{si:datasets}

\subsection{Cumulene}

The guaranteed nonlocal cumulene benchmark~\citep{Batatia2024EquivariantNetworks} tests the dependence of a carbon chain's energy on the relative angle of its terminal CH$_2$ groups. For Table~\ref{tab:cumulene_splits}, 200 training and 50 validation configurations use chain lengths $n_C\in\{3,\ldots,10,13,14\}$, with Gaussian perturbations of optimized geometries and rotations of the terminal groups. The 170 test configurations comprise 50 in-distribution (ID) frames at the training lengths, 60 semi-out-of-distribution (SOD) frames at $n_C\in\{11,12\}$ and 60 out-of-distribution (OOD) frames at $n_C\in\{15,16\}$. Reference energies and forces were computed in ORCA at the $\omega$B97X-D3/def2-TZVP level.

The length and angle sweeps in Figure~\ref{fig:cumulene}(a,b) and the $n_C\in3$--$64$ columns of Table~\ref{tab:cumulene-additional} use separately trained models. Their dataset combines all 420 frames above with 145 additional configurations: length sweeps at $\varphi=5^\circ$ and $85^\circ$, and an angle sweep at $n_C=12$. These 565 frames are randomly split into 509 training and 56 validation frames. This expanded-data protocol is separate from the held-out ID/SOD/OOD comparison.

\subsection{Charged Carbon Chain (C$_{10}$H$_2$/C$_{10}$H$_3^+$)}
The non-periodic C$_{10}$H$_2$/C$_{10}$H$_3^+$ benchmark~\citep{koFourthgenerationHighdimensionalNeural2021} contains neutral decapentayne and its protonated counterpart, testing the nonlocal response to protonation. We retain the original splits of 9,017 training and 1,002 validation configurations, with roughly equal numbers of neutral and charged structures. Table~\ref{tab:ko} reports validation errors.

\subsection{NaCl Clusters (Na$_8$Cl$_8^+$, Na$_9$Cl$_8^+$)}
The NaCl benchmark~\citep{koFourthgenerationHighdimensionalNeural2021} contains non-periodic Na$_8$Cl$_8^+$ and Na$_9$Cl$_8^+$ clusters, both with net charge $+1$. Adding a neutral sodium atom changes the binding energy of a sodium ion on the opposite side of the cluster, approximately 21~\AA\ away. We use a stratified split of 4,500 training and 500 validation configurations. Consistent with prior work~\citep{koFourthgenerationHighdimensionalNeural2021,rumiantsevLearningLongRangeRepresentations2026}, Table~\ref{tab:ko} reports validation errors.

\subsection{Au$_2$/MgO}
The Au$_2$/MgO benchmark~\citep{koFourthgenerationHighdimensionalNeural2021} consists of an Au$_2$ dimer adsorbed on a periodic MgO(001) slab, with or without Al dopant atoms placed deep inside the slab. Each configuration contains 110 atoms in a $9.05\times9.05\times26.46$~\AA\ cell. Although the dopants lie far from the surface, they change the charge distribution of the slab and, with it, the preferred adsorption geometry of the dimer. We use 4,500 training and 500 validation configurations and, as for NaCl, report validation errors in Table~\ref{tab:ko}.

\subsection{MD22}
MD22~\citep{chmielaAccurateGlobalMachine2023} contains non-periodic \textit{ab initio} molecular dynamics trajectories of seven biomolecular and supramolecular systems, labeled at the PBE+MBD level. Table~\ref{tab:md22-data-protocol} gives the atom counts, split sizes and training batch sizes; model settings are in Appendix~\ref{si:md22}.

\begin{table}[H]
\centering\small
\caption{MD22 split sizes (configurations) and batch sizes for MACE, MACE--LES and NEMORA. DHA denotes docosahexaenoic acid.}
\label{tab:md22-data-protocol}
\setlength{\tabcolsep}{5pt}
\begin{adjustbox}{max width=\linewidth}
\begin{tabular}{lrrrrr}
\toprule
System & Atoms & Training & Validation & Test & Batch size \\
\midrule
Ac--Ala3--NHMe & 42 & 6000 & 750 & 6000 & 8 \\
DHA & 56 & 8000 & 1000 & 8000 & 6 \\
Stachyose & 87 & 8000 & 1000 & 8000 & 4 \\
AT--AT & 60 & 3000 & 375 & 3000 & 6 \\
AT--AT--CG--CG & 118 & 2000 & 250 & 2000 & 2 \\
Buckyball catcher & 148 & 600 & 75 & 600 & 2 \\
Double-walled nanotube & 370 & 800 & 100 & 800 & 2 \\
\bottomrule
\end{tabular}
\end{adjustbox}
\end{table}

\subsection{Periodic Ionic Benchmark}
The ionic benchmark contains equal numbers of cations and anions with charges $\pm0.6\,e$. Reference energies and forces combine a Lennard-Jones potential ($\epsilon=0.1$~eV, $\sigma=3.0$~\AA, cutoff 6~\AA) with periodic Coulomb interactions evaluated by PPPM~\citep{hockneyComputerSimulationUsing1988} in LAMMPS~\citep{thompson_lammps_2022}. Geometries are sampled every 400~fs from charge-blind Lennard-Jones dynamics in cubic cells at fixed density, with edge lengths from 20.5~\AA\ ($N=256$) to 51.7~\AA\ ($N=4096$). Each trajectory receives an independent random, charge-neutral assignment, held fixed across its frames. Because charges are assigned after geometry sampling, the corpus does not develop equilibrium ionic screening correlations. At each training size ($N=256,512,1024$), independent trajectories provide 500 training (50 trajectories), 200 validation (20 trajectories) and 50 test configurations (10 trajectories); $N=2048$ and $4096$ contribute 50 test configurations each. A non-periodic counterpart uses spherical clusters of the same sizes and interactions, labeled by direct Coulomb summation, with 400 training, 100 validation and 50 test configurations per training size.

\subsection{Periodic Water}
\label{si:water}
The periodic water dataset was generated in this work. It contains liquid-water configurations in cubic periodic cells with 128 and 256 molecules (384 and 768 atoms), labeled with CP2K~\citep{kuhne_cp2k_2020} using revPBE~\citep{zhang_comment_1998} with D3 dispersion~\citep{grimme_consistent_2010}. The Gaussian and plane-wave calculations use TZV2P MOLOPT basis sets~\citep{vandevondele_gaussian_2007}, GTH pseudopotentials~\citep{goedecker_separable_1996} and a 1200~Ry plane-wave cutoff. Table~\ref{tab:water-splits} gives the split sizes.

\begin{table}[H]
\centering\small
\caption{Periodic water split sizes, in configurations.}
\label{tab:water-splits}
\begin{tabular}{lrrrr}
\toprule
Molecules & Atoms & Training & Validation & Test \\
\midrule
128 & 384 & 788 & 328 & 474 \\
256 & 768 & 578 & 219 & 346 \\
\midrule
Total & --- & 1366 & 547 & 820 \\
\bottomrule
\end{tabular}
\end{table}

\clearpage
\section{Model Settings}
\label{si:model-settings}

Shared backbone and optimization settings are given first, followed by model-specific settings and per-system overrides. MACE-based comparisons use a common short-range architecture where specified. For LOREM and EFA, we prioritize comparable receptive fields ($T r_{\mathrm{cut}}$), angular resolution and maximum irrep degree, and adjust short-range channel counts to balance parameter counts; the explicit settings below record the remaining architectural differences. LOREM, EFA and MACE--RANGE retain their own training implementations.

\subsection{MACE}

The MACE-based models use the backbone in Table~\ref{tab:sr-mace-params}, with per-system overrides in Appendix~\ref{si:system-settings}. MACE, MACE--LES and NEMORA follow Table~\ref{tab:optimizer-settings}; MACE--RANGE uses its separately listed optimizer and schedule. The ionic and water benchmarks use smaller backbones, specified below. Reference MACE options follow v0.3.15; explicitly listed model and system settings take precedence.

\begin{table}[H]
    \centering
    \caption{Shared parameters for short-range MACE backbone upon which all MACE-based long-range models are built.}
    \small
    \begin{tabular}{c|c}
    \toprule
   \textbf{ Hyperparameter} & \textbf{Value} \\
    \midrule
    Interactions & 2 \\[5pt]
    $\ell_\textrm{max}$ & 3 \\[5pt]
    Hidden irreps & 128x0e + 128x1o + 128x2e \\[5pt]
    $\nu$ & 3 \\[5pt]
    Polynomial cutoff exponent & 5 \\[5pt]
    Radial basis functions & 8 \\[5pt]
    Precision & float32 (the Au$_2$/MgO NEMORA run: float64) \\[5pt]
    \bottomrule
    \end{tabular}
    
    \label{tab:sr-mace-params}
\end{table}
\begin{table}[H]
    \centering
    \caption{Default optimizer settings for MACE, MACE--LES and NEMORA; per-system overrides are listed below.}
    \small
    \begin{tabular}{c|c}
    \toprule
   \textbf{ Hyperparameter} & \textbf{Value} \\
    \midrule
    Optimizer & Adam \\[5pt]
    Learning rate & 0.01 \\[5pt]
    Energy:Force weight & 10:100 \\[5pt]
    SWA Energy:Force weight & 1000:1000 \\[5pt]
    EMA decay & 0.99 \\[5pt]
    
    \bottomrule
    \end{tabular}
    \label{tab:optimizer-settings}
\end{table}

MACE, MACE--LES and NEMORA minimize the weighted energy--force loss
\begin{equation}
\label{eq:loss}
\mathcal{L} = \frac{w_E}{B}\sum_{n=1}^{B}\left(\frac{E_n^{\mathrm{pred}}-E_n^{\mathrm{ref}}}{N_n}\right)^2 + \frac{w_F}{3A_B}\sum_{n=1}^{B}\sum_{i=1}^{N_n}\sum_{\alpha=1}^{3}\bigl(F_{n,i,\alpha}^{\mathrm{pred}}-F_{n,i,\alpha}^{\mathrm{ref}}\bigr)^2,
\end{equation}
where $B$ is the number of configurations in a batch, $N_n$ is the atom count of configuration $n$, and $A_B=\sum_n N_n$. Energy errors are normalized per atom before averaging over configurations; force errors are averaged over all Cartesian components in the batch. We use the loss weights in Table~\ref{tab:optimizer-settings}, with overrides in Appendix~\ref{si:system-settings}. These models use no stress, virial or dipole supervision. Per-system cutoffs, batch sizes and schedules are given below.

\subsection{MACE--LES}
MACE--LES~\citep{chengLatentEwaldSummation2025} predicts one scalar latent charge per atom using a separate readout at each MACE layer and sums these readouts before evaluating the long-range energy. The Gaussian smoothing width is $\sigma=1$~\AA, self-interactions are removed, and periodic calculations use $d_l=2$~\AA\ with reciprocal cutoff $k_{\max}=2\pi/d_l$ and the zero mode omitted. Non-periodic calculations use the corresponding real-space pair sum. The implementation's energy normalization factor is 90.4756; the long-range energy is added after backbone scaling, with the short-range atomic interaction shift fixed to zero. Backbone and training settings follow the corresponding benchmark below.

\subsection{LOREM}

\begin{table}[H]
\centering
\caption{LOREM hyperparameters for the cumulene and nonlocal charge-transfer benchmark systems.}
\small
\resizebox{\linewidth}{!}{
\begin{tabular}{lcccc}
\toprule
Hyperparameter & Cumulene & Cumulene (3-64) & NaCl & Carbon \\
\midrule
Cutoff radius (\AA) & 3.0 & 3.0 & 3.0 & 3.0 \\
$\ell_{\textrm{max,lr}}$ & 2 & 2 & 2 & 2 \\
$\ell_{\textrm{max}}$ & 6 & 6 & 6 & 6 \\
Feature channels & 128 & 128 & 128 & 128 \\
Message-passing layers & 1 & 1 & 1 & 1 \\
Radial basis functions & 8 & 8 & 8 & 8 \\
Spherical features & 8 & 8 & 8 & 8 \\
Optimizer & Adam & Adam & Adam & Adam \\
Learning rate & $10^{-3}$ & $10^{-3}$ & $10^{-3}$ & $10^{-3}$ \\
Schedule & linear decay & linear decay & linear decay & linear decay \\
Minimum learning rate & $10^{-6}$ & $10^{-6}$ & $10^{-6}$ & $10^{-6}$ \\
Max epochs & 4000 & 2000 & 1200 & 1200 \\
\bottomrule
\end{tabular}
}
\label{tab:lorem-hyperparameters}
\end{table}

\subsection{EFA}

\begin{table}[H]
\centering
\caption{EFA hyperparameters for the cumulene and nonlocal charge-transfer benchmark systems. The EPE maximum frequency is $\pi$ for non-periodic systems and $\pi/4$ for periodic systems.}
\small
\resizebox{\linewidth}{!}{
\begin{tabular}{lcccc}
\toprule
Hyperparameter & Cumulene & Cumulene (3-64) & NaCl & Carbon \\
\midrule
Cutoff radius (\AA) & 2.0 & 2.0 & 3.0 & 2.0 \\
Feature channels & 144 & 144 & 256 & 144 \\
Message-passing layers & 3 & 3 & 2 & 3 \\
MP maximum degree & 3 & 3 & 3 & 3 \\
MP radial basis functions & 32 & 32 & 32 & 32 \\
ERA maximum degree & 2 & 2 & 2 & 2 \\
ERA maximum length (\AA) & 19.0 & 86.777 & 25.0 & 19.0 \\
ERA query/key channels & 32 & 32 & 32 & 32 \\
ERA value channels & 64 & 64 & 64 & 64 \\
Lebedev points & 50 & 50 & 50 & 50 \\
Optimizer & Adam & Adam & Adam & Adam \\
Learning rate & $10^{-3}$ & $10^{-3}$ & $10^{-3}$ & $10^{-3}$ \\
Schedule & exp decay & exp decay & exp decay & exp decay \\
Final learning rate & $10^{-5}$ & $10^{-5}$ & $10^{-5}$ & $10^{-5}$ \\
Loss weights (E/F) & 0.01 / 0.99 & 0.01 / 0.99 & 0.01 / 0.99 & 0.01 / 0.99 \\
Max epochs & 200000 & 50000 & 10000 & 200000 \\
\bottomrule
\end{tabular}
}
\label{tab:efa-hyperparameters}
\end{table}

\subsection{MACE--RANGE}
MACE--RANGE uses the short-range backbone of Table~\ref{tab:sr-mace-params} with a 3~\AA\ cutoff. It is trained with a separate implementation, so its optimizer and schedule differ from the other MACE-based models (Table~\ref{tab:mace-range-hyperparameters}). Instead of SWA, the loss weights are switched once during training.

\begin{table}[H]
\centering
\caption{MACE--RANGE module and training hyperparameters for the cumulene and nonlocal charge-transfer benchmark systems. Backbone settings follow Table~\ref{tab:sr-mace-params}.}
\small
\begin{adjustbox}{max width=\linewidth}
\begin{tabular}{lcccc}
\toprule
Hyperparameter & Cumulene & Cumulene (3-64) & NaCl & Carbon \\
\midrule
Virtual nodes & 4 & 3 & 3 & 3 \\
Virtual hidden channels & 432 & 432 & 432 & 432 \\
Virtual basis dimension & 10 & 10 & 10 & 10 \\
Virtual heads & 8 & 8 & 8 & 8 \\
Optimizer & AdamW & AdamW & AdamW & AdamW \\
Learning rate & $10^{-3}$ & $10^{-3}$ & $10^{-3}$ & $10^{-3}$ \\
Weight decay & $5\times10^{-7}$ & $5\times10^{-7}$ & $5\times10^{-7}$ & $5\times10^{-7}$ \\
Scheduler & CustomStepLR & CustomStepLR & CustomStepLR & CustomStepLR \\
Scheduler steps & 300, 200, 100 & 300, 200, 100 & 608, 192, 160 & 608, 192, 160 \\
Scheduler gammas & 0.8, 0.8, 0.8 & 0.8, 0.8, 0.8 & 0.8, 0.8, 0.8 & 0.8, 0.8, 0.8 \\
Max epochs & 2000 & 2000 & 600 & 600 \\
Loss-weight switch epoch & 400 & 500 & 300 & 300 \\
Batch size & 64 & 16 & 16 & 32 \\
\bottomrule
\end{tabular}
\end{adjustbox}
\label{tab:mace-range-hyperparameters}
\end{table}

\subsection{NEMORA}
\label{si:nemora-settings}
All learnable NEMORA models attach after the first MACE product block (index $-2$) and use operators up to degree $\ell=2$. They use a single Kelvin anchor and apply one Highway-gated Neumann iteration with cumulant modulation $\gamma_{\mathrm{cum}}=0.5$ unless stated otherwise. The tree is built with the adaptive octree of Section~\ref{sec:hierarchy}. The backbone and NEMORA parameters form separate optimizer groups; the NEMORA group uses a linear warmup of 2000 steps and a tensor-product weight decay of $10^{-4}$. Per-system irreps, contraction orders, learning rates and tree settings are given in Appendix~\ref{si:system-settings}.

\subsection{Per-System Settings}
\label{si:system-settings}
Within each benchmark, MACE, MACE--LES and NEMORA share the backbone, data splits and training schedule unless stated otherwise. Defaults follow Tables~\ref{tab:sr-mace-params} and~\ref{tab:optimizer-settings} and Appendix~\ref{si:nemora-settings}; the settings below take precedence. Loss weights are written as energy:force. Ionic-system and water settings are given in Appendices~\ref{si:settings-ions} and~\ref{si:settings-water}.

\subsubsection{Cumulene}
\label{si:settings-cumulene}
\paragraph{Backbone and training.} The backbone has two interaction layers, hidden irreps $128\times0\mathrm e+128\times1\mathrm o+128\times2\mathrm e$, $\ell_{\max}=3$, correlation order $\nu=3$ and a 3~\AA\ cutoff. Training uses Adam (AMSGrad) with an initial learning rate of $10^{-2}$, a batch size of 16 and loss weights of $1{:}100$; the SWA stage uses a learning rate of $10^{-3}$ and weights of $100{:}10$. The main benchmark trains for 2000 epochs with SWA from epoch 1000, and the 3--64 sweep model for 1000 epochs with SWA from epoch 500. The MACE and MACE--LES baselines of the 3--64 sweep instead use weights of $10{:}100$ and $1000{:}1000$.

\paragraph{NEMORA.} Both cumulene models use the same configuration, with hidden irreps $32\times0\mathrm e+32\times1\mathrm o+32\times2\mathrm e$, maximum degree $\ell=2$ and $\nu=3$. NEMORA parameters are trained with a learning rate of $10^{-3}$. The tree targets three children per node and has two levels, with $\theta_{\mathrm{MAC}}=0.5$. Aggregation is extensive throughout the hierarchy, and the energy readout adds a bilinear long-range energy head.

\subsubsection{Charged Carbon Chain}
\label{si:settings-carbon}
\paragraph{Backbone and training.} The backbone matches cumulene, with a 3~\AA\ cutoff. Models train for 600 epochs with a batch size of 32, loss weights of $10{:}100$, and SWA from epoch 300 with weights of $1000{:}1000$.

\paragraph{NEMORA.} NEMORA uses hidden irreps $32\times0\mathrm e+32\times1\mathrm o+32\times2\mathrm e$, $\ell=2$, $\nu=2$ and the trainer's default NEMORA learning rate of $10^{-3}$. The tree targets four children per node and has two levels, with $\theta_{\mathrm{MAC}}=0.5$. Upward moments are extensive and the fusion uses a density mean.

\subsubsection{NaCl Clusters}
\label{si:settings-nacl}
\paragraph{Backbone and training.} The backbone matches cumulene, with a 3~\AA\ cutoff. Models train for 600 epochs with a batch size of 16, loss weights of $10{:}100$, and SWA from epoch 300 with weights of $1000{:}1000$.

\paragraph{NEMORA.} NEMORA uses hidden irreps $40\times0\mathrm e+40\times1\mathrm o+40\times2\mathrm e$, $\ell=2$, $\nu=3$ and a learning rate of $10^{-3}$. The tree targets eight children per node, has two levels and at most four atoms per leaf, with $\theta_{\mathrm{MAC}}=0.5$. The downward pass uses two resolvent iterations.

\subsubsection{Au$_2$/MgO}
\label{si:settings-aumgo}
\paragraph{Backbone and training.} The backbone matches cumulene, with a 3~\AA\ cutoff. Models train for 600 epochs with a batch size of 2, loss weights of $10{:}100$, and SWA from epoch 300 with weights of $1000{:}1000$. The MACE SWA stage uses weights of $100{:}100$.

\paragraph{NEMORA.} NEMORA is trained in float64, with hidden irreps $32\times0\mathrm e+32\times1\mathrm o+32\times2\mathrm e$, $\ell=2$, $\nu=2$ and a learning rate of $10^{-3}$. The tree targets eight children per node, has three levels and at most four atoms per leaf, with $\theta_{\mathrm{MAC}}=0.6$. The periodic slab uses Madelung lattice sums in both the hierarchy, with two image shells, and the Kelvin branch.

\subsubsection{MD22}
\label{si:md22}
We train MACE, MACE--LES and NEMORA on all seven non-periodic MD22 systems~\citep{chmielaAccurateGlobalMachine2023}. Within each system, the three models use the same data splits, MACE backbone, batch size and training schedule (Table~\ref{tab:md22-data-protocol}). Each result averages three independently trained models, evaluated on the held-out test split at the checkpoint with the lowest validation force RMSE within the SWA stage. Seeds are 42, 43 and 44, except for AT--AT--CG--CG: MACE--LES uses 42, 44 and 45, and NEMORA uses 43, 44 and 45.

\paragraph{Additional baselines and evaluation.}
MACE--RANGE, LOREM and EFA are evaluated on the same held-out splits with seeds 42, 43 and 44, using their original implementations, trained on the 400-epoch protocol of Table~\ref{tab:md22-data-protocol}. LOREM uses the checkpoint with the lowest validation force RMSE, MACE--RANGE uses the final checkpoint after 400 epochs, and EFA uses its retained validation-selected checkpoint. The LOREM nanotube seed-43 recovery uses element-wise gradient clipping at 1.0 and is scored at its best-validation checkpoint. Energy RMSE equally weights per-frame errors after division by atom count; force RMSE equally weights Cartesian components. Atomic reference energies are restored without fitting a test-set offset.

EFA evaluation passes checkpoint parameters as dynamic JIT arguments, as in native training and validation, while retaining the same energy model, automatic-differentiation force calculation and NPZ batching. Inputs and parameters remain float32; matrix products use highest FP32 precision and GPU operations are deterministic. Each full test split is evaluated in two independent GPU processes, with identical predictions in all 21 runs. Before scoring, CPU/GPU parity passed on all 750 Ala3 validation frames, and independent-process validation repeats passed for Ala3 and the nanotube. The acceptance tolerances are $10^{-4}$~eV/\AA\ for forces and eight float32 ULPs for energies. CPU/GPU parity was not separately checked for every test checkpoint, and the compiler-internal cause of the earlier closed-parameter discrepancy remains unverified.

\paragraph{Shared backbone and training.}
All three models use a MACE backbone with two interaction layers, hidden irreps $128\times0\mathrm e+128\times1\mathrm o+128\times2\mathrm e$, edge spherical harmonics up to $\ell_{\max}=3$, correlation order $\nu=3$ and a 4~\AA\ cutoff. Models are trained for 400 epochs with Adam (AMSGrad) at an initial learning rate of $10^{-2}$ and energy:force loss weights of $10{:}100$. Validation is evaluated every two epochs; before SWA, the plateau scheduler is stepped each epoch using the latest validation loss, with reduction factor 0.8 and patience 50. EMA decay is 0.99 and backbone weight decay is $5\times10^{-7}$. SWA starts at epoch 200 with a learning rate of $10^{-3}$ and loss weights of $1000{:}1000$. Atomic reference energies, force-RMS scaling and average neighbor counts are estimated from training data; the LES interaction shift is fixed to zero.

\paragraph{NEMORA and the adaptive tree.}
NEMORA is attached after the first MACE product block, with hidden irreps $32\times0\mathrm e+32\times1\mathrm o+32\times2\mathrm e$, maximum operator degree $\ell=2$ and contraction order $\nu=2$. Its parameters are trained with a learning rate of $10^{-3}$. The radial network uses eight basis functions and one hidden layer of width 64 followed by its output layer; the invariant gate also has width 64. The tree is built adaptively in the PCA frame, targeting four children per node, with at most 16 atoms per leaf and up to eight children per node; its depth adapts to each geometry rather than being fixed. Far-field interactions use $\theta_{\mathrm{MAC}}=0.5$, and a single global Kelvin anchor is used. Aggregation exponents are $\alpha_{\mathrm{up}}=0$ and $\alpha_{\mathrm{down}}=1$.

\subsubsection{Ionic systems}
\label{si:settings-ions}
Within each corpus, analytical NEMORA and LES share the backbone, data splits, optimizer, and training schedule. The ionic backbone has two interactions, a 4.5~\AA\ cutoff, and $64\times0\mathrm{e}+64\times1\mathrm{o}+64\times2\mathrm{e}$ hidden irreps. Training runs in float32 for 240 epochs with a batch size of 4, an initial learning rate of $10^{-2}$ and energy:force loss weights of $10{:}100$, with stochastic weight averaging (SWA) starting at epoch 120 with weights of $1000{:}1000$. The periodic dipole arm instead uses a batch size of 2 and float64 for the long-range calculation; its backbone remains float32. The analytical controls are single-seed runs, while the periodic MACE baseline is a three-seed mean, and checkpoints are selected by validation force error. The analytical expansion order is $p=3$, with $\theta=0.35$ for clusters and $\theta=0.35$, $w=2$ for periodic interaction lists. The periodic LES controls use reciprocal-grid parameter $d_l=2.0$~\AA. The tree is occupancy-adaptive, with at most 16 atoms per leaf and up to six levels.

\subsubsection{Periodic water}
\label{si:settings-water}
MACE, MACE--LES and each analytical model share one backbone: one MACE interaction, a 6~\AA\ cutoff and $128\times0\mathrm{e}+128\times1\mathrm{o}$ hidden irreps, trained for 800 epochs with an initial learning rate of $10^{-2}$ and loss weights of $10{:}100$, and SWA starting at epoch 400 with weights of $1000{:}1000$. Both source-degree arms use per-size batches of four 384-atom or two 768-atom frames. The analytical settings are $p=3$, $\theta=0.35$ for periodic lists and image-shell radius $w=2$, with the long-range term evaluated in float64. The tree has a fixed depth of two levels with a target of eight children per node.

\subsection{Backbone interfaces}
\label{si:backbone-settings}
For a native equivariant host, NEMORA receives the host's typed scalar and tensor features. For scalar-only hosts, a geometry-based lifting map combines scalar features with relative-position angular information to construct tensor sources. PET has no native irreducible-representation decomposition, so its attachment uses an explicit lift as well. The lift preserves equivariance when its scalar inputs are invariant; it does not impose exact equivariance on the entire PET backbone.

\subsection{Charge-equilibration and electrostatic reference settings}
\label{si:electrostatic-settings}
CELLI's charge-loss weight is 50 on the carbon chain, 10 on NaCl and 200 on Au$_2$/MgO. SpookyNet uses six message-passing layers against our two, SpookyNet's carbon-chain and Au$_2$/MgO results use a larger cutoff than its original.

\subsection{Scaling benchmark settings}
\label{si:scaling-settings}
The MACE-based scaling models use two MACE interactions, correlation order three, a 4~\AA\ cutoff and $128\times0\mathrm{e}+128\times1\mathrm{o}$ hidden irreps. Models are built with untrained weights. EFA and LOREM use their own short-range architectures (two message-passing rounds and a 4~\AA\ cutoff); their long-range blocks are kept as in the accuracy benchmarks.

\clearpage
\subsection{Component ablation settings}
\label{si:ablation-settings}
\begin{table}[H]
\centering\small
\caption{Definitions of the component-ablation arms. All arms are retrained. Joint changes are identified explicitly.}
\label{tab:ablation-definitions}
\begin{tabular}{lp{0.70\linewidth}}
\toprule
Arm & Change relative to FULL \\
\midrule
NOHIER & Remove the hierarchy, retaining the standalone Kelvin branch. Periodic root injection is also removed. \\
NOKELVIN & Remove the standalone Kelvin branch. \\
NOKESL & Remove the per-box Kelvin contribution to the downward pass. \\
FIXEDTREE & Replace adaptive coarsening by a fixed top-down octree. \\
MAXLEV1/3 & Set maximum hierarchy depth to one or three. \\
LMAX0/1/3 & Set source-feature and operator degree together. On cumulene, degrees below two also disable the quadrupole-dependent Kelvin head. \\
MAXL0/1 & Reduce only the source-feature degree, retaining degree-two operators. \\
ELL3 & Increase only the operator degree to three. \\
NU0/2/3 & Change the many-body contraction order; zero removes the contraction. \\
M2LPIN & Pin the radial shape to $(1+u^2)^{-(\ell+1)/2}$, retaining a learned gain. \\
M2LDECAY & Learn the radial decay exponent. \\
NOVSQRT & Remove V-list square-root normalization. \\
TRES1 & Reduce the resolvent truncation order from two to one. \\
NOFUSE & Remove the additive long-range feature update. \\
NOLRHEAD & Remove the separate long-range energy head. \\
NOGATE & Remove the multipole alignment gate. \\
LAD2/3 & Remove many-body contraction and use scalar degree; LAD3 also removes the Kelvin branch. \\
FMMCLASSIC & Remove contraction, Kelvin branch, alignment gate and three normalization modes, and pin the radial shape. Learned source/readout maps remain. \\
NOHIERPBC & Remove periodic root injection. \\
NOKELVPBC & Remove the Kelvin periodic image sum. \\
NOTINFOIL & Remove the surface-dipole correction. \\
NSHELLS2 & Increase the periodic image-shell radius from one to two. \\
SLABMASK & Mask replica images along the slab vacuum direction. \\
MICSPLIT & Use a minimum-image near/far split with two image shells. \\
\bottomrule
\end{tabular}
\end{table}

\end{document}